%% file: GeoKmeans.tex
\documentclass[sn-basic, referee]{sn-jnl}

\usepackage{mathrsfs}%
\usepackage[title]{appendix}%
\usepackage{textcomp}%
\usepackage{manyfoot}%
\usepackage{algorithmicx, algpseudocode}%
\usepackage{listings}%
\usepackage{wrapfig}

\usepackage[ruled, lined, linesnumbered, longend]{algorithm2e}

\usepackage{amsmath, amsfonts, amssymb, mathrsfs, bbm, mathtools, amsthm, bm}
\usepackage{color, colortbl}
\usepackage{enumerate}
\usepackage{textcomp}
\usepackage{stfloats}
\usepackage{url}
\usepackage{verbatim}
\usepackage{graphicx, subcaption}
\usepackage{booktabs, makecell, float, longtable, adjustbox, multirow, adjustbox}
\usepackage{varwidth}
\usepackage{tabularx}
\usepackage{cite}
\usepackage[usenames,dvipsnames,svgnames]{xcolor}

\usepackage[most]{tcolorbox}
\newtcolorbox{myquote}[1][]{%
    colback=black!5,
    colframe=black!5,
    notitle,
    sharp corners,
    borderline west={2pt}{0pt}{red!80!black},
    enhanced,
    breakable,
}

\definecolor{mygreen}{HTML}{66CC99}
\definecolor{myred}{HTML}{CC6666}

\def\km{$k$-means}
\def\kg{$\mathsf{G}k$-means}
\def\kb{Ball-$k$means}
\def\ham{Hamerly}
\def\ann{Annulus}
\def\exp{Exponion}
\def\kd{Dualtree}
\def\blklst{Blacklist}
\def\elk{Elkan}
\newtheorem{theorem}{Theorem}[section]
\newtheorem{definition}{Definition}[theorem]
\newtheorem{lemma}{Lemma}[theorem]

\begin{document}

\title[Geometric-$k$-means]{Geometric-$k$-means: A Bound Free Approach to Fast and Eco-Friendly $k$-means}

\author*[1]{\fnm{Parichit} \sur{Sharma}}\email{parishar@iu.edu}

\author[1]{\fnm{Marcin} \sur{Stanislaw}}\email{mamalec@iu.edu}

\author[1,2]{\fnm{Hasan} \sur{Kurban}}\email{hkurban@hbku.edu.qa}

\author[3]{\fnm{Oguzhan} \sur{Kulekci}}\email{kulekcmo@MiamiOH.edu}

\author[1]{\fnm{Mehmet} \sur{Dalkilic}}\email{dalkilic@iu.edu}

\affil*[1]{\orgdiv{Department of Computer Science}, \orgname{Luddy School of Informatics, Computing \& Engineering}, \orgaddress{\city{Bloomington}, \postcode{47408}, \state{Indiana}, \country{USA}}}

\affil[2]{\orgdiv{College of Science \& Engineering}, \orgname{Hamad Bin Khalifa University}, \orgaddress{\city{Doha}, \postcode{23874}, \country{Qatar}}}

\affil[3]{\orgdiv{Department of Computer Science and Software Engineering}, \orgname{Miami University}, \orgaddress{\city{Oxford}, \postcode{45056}, \state{Ohio}, \country{USA}}}

\abstract{This paper introduces Geometric-$k$-means (or \kg{} for short), a novel approach that significantly enhances the efficiency and energy economy of the widely utilized $k$-means algorithm, which, despite its inception over five decades ago, remains a cornerstone in machine learning applications. The essence of \kg{} lies in its active utilization of geometric principles, specifically scalar projection, to significantly accelerate the algorithm without sacrificing solution quality. This geometric strategy enables a more discerning focus on data points that are most likely to influence cluster updates, which we call as high expressive data (HE). In contrast, low expressive data (LE), does not impact clustering outcome, is effectively bypassed, leading to considerable reductions in computational overhead. Experiments spanning synthetic, real-world and high-dimensional datasets, demonstrate \kg{} is significantly better than traditional and state of the art (SOTA) \km{} variants in runtime and distance computations (DC). Moreover, \kg{} exhibits better resource efficiency, as evidenced by its reduced energy footprint, placing it as more sustainable alternative.}

\keywords{Fast $k$-means, Data-Centric AI, Single-cell RNASeq, Big Data, AI \& Sustainability}

\maketitle

\input{Introduction}

\input{Literature}

\input{Methods}

\input{Experiments}
\input{ConclusionFutureWork}

\backmatter

\bmhead{Acknowledgements}
Authors thank the technical support extended by the system administration team of Luddy School of Informatics, Computing \& Engineering for their help in deploying the system infrastructure for this study.

\bmhead{Author Contributions}
PS: Conceptualization, Methodology, Development \& Design, Experiment, Investigation, Writing – original draft, Writing – review \& editing, Visualization. MS: Development \& Design, Validation, Investigation, Writing – original draft, Writing – review \& editing. HK: Investigation, Validation, Visualization, Supervision. OK: Methodology, Validation, Writing – review \& editing, MD: Writing – original draft, Writing – review \& editing, Resources, Supervision, Project - administration.

\bmhead{Funding}
Not applicable.


\bmhead{Data availability} All datasets used in this work are openly available. Relevant sources are cited in the manuscript.

\section*{Declarations}

\begin{itemize}
    \item \textbf{Conflict of interest} The authors declare that they have \textbf{no} competing financial or non financial interests.
    \item \textbf{Ethics approval} Not applicable.
    \item \textbf{Consent to participate} Not applicable.
    \item \textbf{Consent for publication} All authors give their consent for the publication.
\end{itemize}

\bibliography{References}

\newpage

\begin{appendices}

\input{Appendix}

\end{appendices}

\end{document}

%% file: Introduction.tex
\section{Introduction}\label{sec:introduction}

The \km{} clustering algorithm remains a foundational pillar within machine learning for its straightforward application, efficiency, and application across domains \citep{jain2010data, ahmed2020k, ikotun2023k}. Recent enhancements to the traditional \km{} framework have embraced a strategy focused on the data itself, aiming to partition it more efficiently by predicting changes in data point membership between iterations. More particularly, since $k$-means is a sequence of cross-products of $k$ centroids with all data as distance computations, reducing the effective data size reduces the run-time. Two general approaches currently exist: (1) bounded methods, which utilize distance bounds to determine when distance calculations are necessary, while still guaranteeing identical results to standard \km{}, and (2) unbounded methods, which achieve acceleration without requiring the maintenance of such distance bounds. \kb{} \citep{XiaBallKmeans2021} is the state of the art for accelerating \km{} without distance bounds and has become popular due to its real-world performance.

\begin{figure}[!htbp]
\centering
\includegraphics[scale=.80]{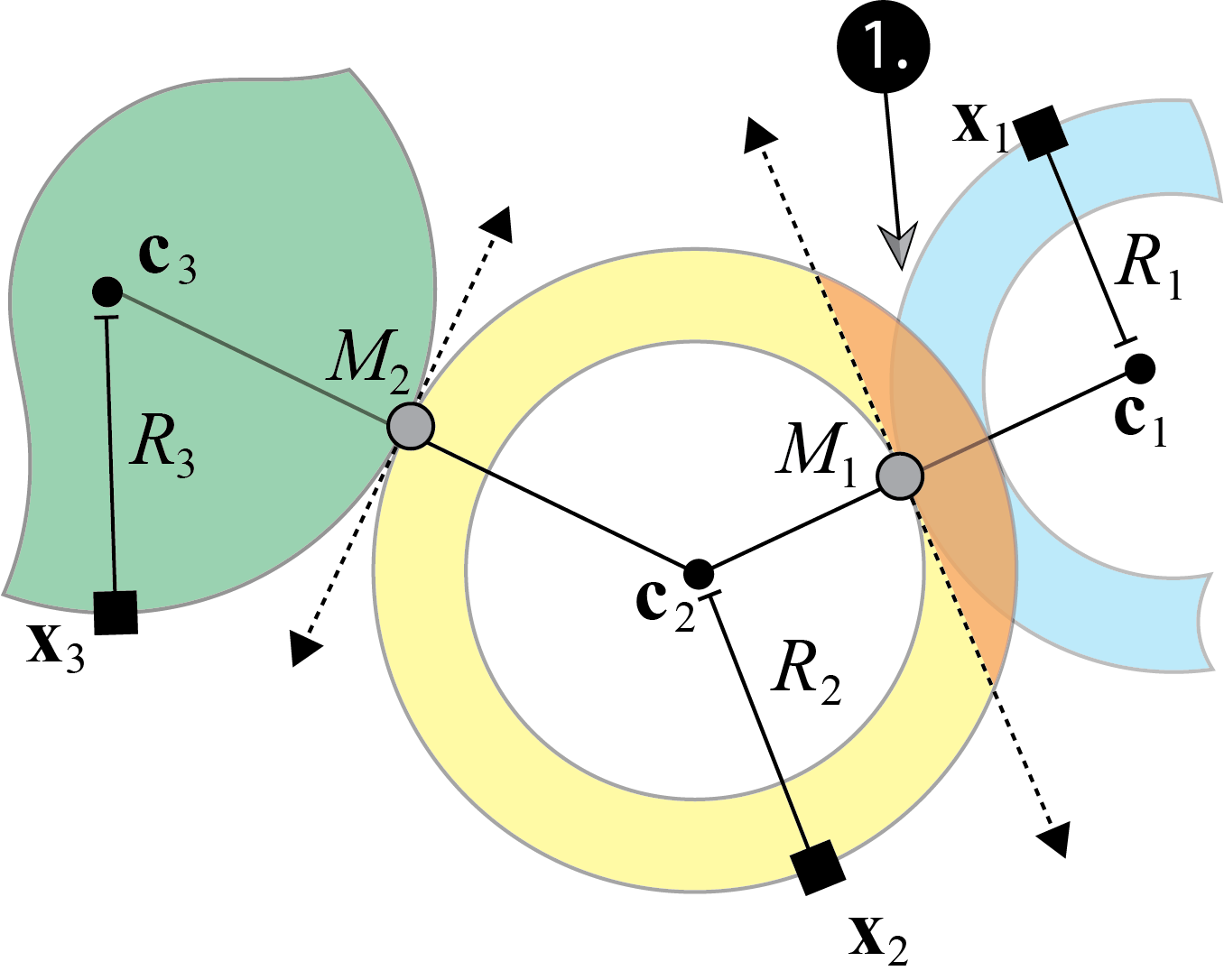}
\caption{\textbf{Illustrating neighbors and HE}
$M_1, M_2$ indicate midpoints between the centroids. Centroids $\mathbf{c}_1, \mathbf{c}_2$ are considered as neighbors because they overlap and their radii (determined from furthest members) establish a region of overlap (arrow 1.); data residing in this region is considered high expressive since these can switch membership. The data outside the overlapping region is considered as LE/LHE, and may not change membership.} 
\label{fig:figs}\label{fig:Figure1}
\end{figure}

In this work, we propose a second, unbounded method motivated by the spatial orientation of data. Fig.\,\ref{fig:Figure1} highlights the general concept of neighborhood and cluster overlap as used in this new method. Data in the overlapping region is used in the subsequent iteration. To make discussion simpler, we reuse existing definitions and call this \textbf{high expressive} (HE) data, due to their potential for altering cluster memberships and influencing the centroids. The complement of the overlapping region respective to the centroid is \textbf{low expressive data} (LE) characterizing unchanging membership. Convergence occurs similarly in both bounded and unbounded \km{}, though rates depend on the amount of data and so forth, as discussed in the next section. We point out that, from an expressiveness perspective, convergence grows faster commensurate with a growing amount of LE. Presented here is a novel approach to determining HE and LE data, both more effectively and rapidly by using principles of geometry. We call this new approach Geometric-$k$-means.

This work presents three significant contributions to the field: (1) a novel algorithm that not only consumes fewer resources but also centers around data processing efficiency, surpassing both traditional and cutting-edge algorithms in performance without compromising solution quality. (2) we offer a proof of convergence and demonstrate that our algorithm, \kg{}, achieves the same results as \km{} when both are initialized in the same way, despite \kg{} operating at a notably faster pace. (3) We initiate the analysis on the energy consumption of various \km{} methodologies, revealing that \kg{} stands out for its resource efficiency, specifically through lower energy usage. Furthermore, we delve into the geometric principles underlying the \kg{} algorithm to shed light on its operational foundations. The rest of the paper is structured as follows: Sec.\,\ref{sec:literature} reviews the relevant literature; Sec.\,\ref{sec:methods} details the \kg{} algorithm, the theoretical underpinnings that contribute to its efficiency, and the proof of its results' symmetry with those of \km{}; Sec.\,\ref{sec:Experiments} outlines the experimental setup and discuss the results. Lastly, Sec.\,\ref{sec:summary} concludes with the summary of this work and outlines avenues for future research.
  

%% file: Literature.tex
\section{Related work}\label{sec:literature}
In \km{} clustering algorithm, the lion's share of compute time is the repeated distance calculations (DC) between data and centroids. This process is repeated until either the centroids converge \textit{i.e.} membership does not change or some user-defined threshold on the number of iterations is reached. There exist different ways to implement \km: \citep{LloydKmeans, MacQueenKmeans, Forgy1965ClusterAO}; however, none of these methods address reducing DC. Diverse techniques have emerged to expedite the conventional \km{} algorithm. The majority can be broadly categorized as follows: 1) methods that result in the same solution as \km{} \citep{Pelleg1999, Moore2000TheAH, steven2002kmeans, Kanungo02anefficient, Elkan2003, hamerly1, petrandhamerly2016, Newling2016, xia2020fast}, 2) algorithms engineered for swiftness, albeit yielding approximate solutions divergent from \km{} \citep{Philbin2007, Davidkmeans, philbin2010scalable, Wang2012, Hu2017FastKF, Deng2018FastKB, FRANTI201995}, and 3) methods that expedite convergence by preprocessing steps, most notably centroid initialization \citep{Bradley98refininginitial, Arthur2007, Bachem_Lucic_Hassani_Krause_2016}. Good initialization may lead to faster convergence but does not ameliorate the problem of redundant DC. On the other hand, approximate algorithms diverge from the quality of solution. In contrast, our focus in this work is designing an exact and faster \km{} clustering. 

Geometric-$k$-means capitalizes on the phenomenon that, as the \km{} algorithm continues, an increasingly smaller fraction of data changes its membership; as a corollary, an increasingly large fraction of data remains assigned to the same cluster. From a data-centric perspective, the pool of HE data increasingly becomes smaller as LE becomes larger--both eventually fixed meaning stability; this characterizes the increase in the number of ignored DC. Conceptually, the idea of selecting data for distance calculations has been implemented in different ways for example, \citep{Pelleg1999} make use of $k$-d tree for partitioning and storing the data where distance calculations are only performed with the closest neighbor nodes of a data point; however, $k$-d tree methods only perform well on data with 8-10 dimensions while performance decreases rapidly as the number of dimensions increases. 

\citep{Moore2000TheAH} developed a metric tree-based structure to organize the data into anchor hierarchies, and this new structure performed well in high dimensions. \citep{Elkan2003} proposed to store the bounds on the distance of a data point to centroids and only perform the distance calculation when the bounds are violated. This idea was further refined in \citep{hamerly1} by storing fewer bounds per data point and using them in a more effective pruning strategy that allows skipping the innermost loop more frequently. \citep{drake2013faster} propose \ann{}, which makes use of annulur region centered on the origin and triangle inequality to shorten number of centroids during DC. The ideas proposed by \elk{} and \ham{} were refined by \citep{Newling2016}, by defining a spherical region (called Exponion) centered on the cluster centroids, and further tightening the bounds, which helps to reduce DC. In a parallel work, \citep{petrandhamerly2016} propose the idea of limiting distance calculations to neighbor centroids. \citep{pmlr-v84-forster18a, FlorianandForster} propose to accelerate mixture models and \km{} clustering by using truncated variational EM, which enables each iteration to scale sublinearly with the number of clusters. The approach optimizes a variational free energy bound instead of the likelihood directly, maintaining comparable clustering quality while drastically reducing computational demands. A bound-free method was recently introduced in \citep{XiaBallKmeans2021} (referred to as \kb{}), where the distance computations are done for data in specific annular regions referred to as the active area. \kb{} has been shown to perform better than bounded \km{} methods. With respect to data expressiveness, it has been used to actively reduce computations for soft-clustering, i.e. EM (expectation-maximization) algorithm, \citep{kurban2017using, KurbanExp2021, SHARMA2022100944}. To utilize the expressiveness of data, authors embed and update an ordered structure during each iteration of EM, and subequently consider data in leaf nodes as HE. Here, we adapt data expressiveness for \km{}, and propose a novel algorithm, which is exact, demonstrably faster, and is independent of distance bounds.

%% file: Methods.tex
\section{Methods}\label{sec:methods}
We use the following notation throughout the paper. A datum $\mathbf{x} \in \mathbb{R}^d$ is a $d$ dimensional vector over the reals. The data $D = \{\mathbf{x}_1, \mathbf{x}_2, \ldots, \mathbf{x}_m\}$ is a set of $m$ vectors. We write $x_i$ for the $i^{th}$ element of $\mathbf{x}$. We write the scalar product of vectors $\mathbf{x}$ and $\mathbf{y}$ as $\mathbf{x}\cdot\mathbf{y} = \sum_{i=1}^d x_iy_i$. The length (or norm) of vector $\mathbf{x}$ is $||\mathbf{x}|| = \sqrt{\mathbf{x} \cdot \mathbf{x}}$. The distance between $\textbf{x}$ and $\mathbf{y}$ is written $d(\mathbf{x},\mathbf{y}) = \|\mathbf{x}-\mathbf{y}\|$. A vector from $\mathbf{x}$ to $\mathbf{y}$ is denoted as $\overrightarrow{\mathbf{xy}}$. The midpoint coordinates are determined: $\mathsf{m}(\mathbf{x},\mathbf{y}) = \frac{1}{2}(\mathbf{x} + \mathbf{y})$. The number of centroids is $k \in \mathbb{N}$. \\

The set of centroids (mean) at the $j^{th}$ iteration is denoted by $\mathbf{C}^{(j)} = \{\mathbf{c}_1^{(j)},\ldots, \mathbf{c}_k^{(j)}\}$. Assignment of data to the clusters at $j^{th}$ iteration is denoted by $\mathbf{A}^{(j)} = \{\mathbf{a}_1^{(j)},\ldots, \mathbf{a}_k^{(j)}\}$ $s.t.$ $\mathbf{a}_{1}$ contain data points assigned to the centroid $\mathbf{c}_1$. Index function $\mathcal{I}(\mathbf{x}_i) = j$, indicate that a data point $\mathbf{x}_i$ is assigned to centroid $\mathbf{c}_j$. Given a subset $D' \subseteq D$ and a centroid $\mathbf{c}_j$, sum of squared errors ($SSE$) can be represented as: $\sum_{i=1}^{|D'|} ||\mathbf{x}_i - \mathbf{c}_j||^2$. Similarly, $SSE$ for all clusters can be written as: $SSE(D, \mathbf{C}) = \sum_{i = 1}^{m} ||\mathbf{x}_i - \mathbf{c}_{\mathcal{I}(\mathbf{x_i})}||^2$. We reuse the notation $s(\mathbf{c})$ (introduced by Elkan) to denote half the distance between a centroid $\mathbf{c}$ and its closest other centroid. The notation \textsf{LE}, \textsf{LHE} and \textsf{HE} are used to refer to the low expressive, likely high expressive and high expressive data, respectively.\\


\begin{definition}[Radius]\label{def:radius}
The radius of a cluster is the maximum of the distances between the centroid and the data assigned to the cluster: 
\begin{equation}
\label{eq:radius}
r(\mathbf{c}_j) \;=\; \mathrm{max} \; \{d(\mathbf{x}, \mathbf{c}_j) \;|\; \forall\ \mathbf{x} \in \mathbf{a}_j \}
\end{equation}
\hfill
$\blacksquare$
\end{definition}


\subsection{\textbf{Finding the Neighbor Clusters}}\label{sec:finding_neighbor} Lloyd's \km{} requires the distance be computed between a data point and all centroids, but remains agnostic towards the proximity of the data point to specific centroids. For instance: if a data point $\mathbf{x}$ is assigned to $\mathbf{c}_i$, then $\mathbf{x}$ is already closer to $\mathbf{c}_i$ (and its neighbors) than $\mathbf{c}_j$ when $\mathbf{c}_j$ is not a neighbor of $\mathbf{c}_i$. DC between $\mathbf{x}$ and $\mathbf{c}_j$ are, therefore, not required. Using neighbors to reduce the distance computations is not new and has been reported previously by \citep{steven2002kmeans} and was subsequently combined with the idea of distance bounds \citep{petrandhamerly2016}. \kb{} define neighbors by comparing the inter-centroid distances with cluster radius and propose conditions satisfying which the neighborhood calculations can be skipped. This approach can significantly reduce the time for neighbor finding, particularly for higher number of clusters, or when centroids are relatively stable. In worst case, however, DC remains $O(k^2)$, especially in the initial phase of \kb{} when centroids could shift abruptly. Moreover, centroids are sorted as an additional step to help filter data points in later stages of the algorithm. However, sorting can become a significant overhead, especially if the cluster neighborhood undergoes frequent changes. In designing \kg{}, we found an approach that does not require sorting, instead only keeping track of the nearest neighbor of each centroid. The pseudo code of this approach is provided in Alg. ~\ref{algo:neighbors_finding}.\\

\begin{definition}[Neighboring centroids]\label{def:neighbors}
Given two distinct centroids $\mathbf{c}$ and $\mathbf{c}'$ belonging to set of centroids $\mathbf{C}$, we say $\mathbf{c}'$ is a neighbor of $\mathbf{c}$ if:

\begin{equation}\label{eq:neighbor}
\frac{1}{2}d(\mathbf{c}, \mathbf{c}') \leq r(\mathbf{c}) + s(\mathbf{c}) \; \forall\ \mathbf{c}, \mathbf{c}' \in \mathbf{C} \ \wedge \mathbf{c} \neq \mathbf{c}'
\end{equation}
\hfill
$\blacksquare$
\end{definition}

\begin{lemma}\label{lemma:neighborhood_lemma_1}
A centroid $\mathbf{c}'$ can \textbf{not} be a neighbor of $\mathbf{c}$, if: 
\begin{equation}\label{eq:neighbor_eq1}
\frac{1}{2}(d(\mathbf{c}, \mathbf{c}')) > r(\mathbf{c}) + s(\mathbf{c})
\end{equation}

\noindent \textbf{\textsc{Proof}} Let $\mathbf{c}^n$ be the nearest centroid to $\mathbf{c}$. If $\frac{1}{2}(d(\mathbf{c}, \mathbf{c}')) > r(\mathbf{c}) + s(\mathbf{c})$, then $\mathbf{c}'$ cannot be a neighbor of $\mathbf{c}$, because $\forall \mathbf{x} \in \mathbf{c}$, $d(\mathbf{x}, \mathbf{c}) \: < \: d(\mathbf{x}, \mathbf{c}')$, and $d(\mathbf{x}, \mathbf{c}^n) \: < \: d(\mathbf{x}, \mathbf{c}')$.

\vspace{-0.5cm}

\begin{eqnarray}
d(\mathbf{c}, \mathbf{c}') &\leq& d(\mathbf{x}, \mathbf{c}) + d(\mathbf{x}, \mathbf{c}') \; (\text{from} \: \triangle \: \text{inequality}) \\
d(\mathbf{x}, \mathbf{c}^n) &\leq& d(\mathbf{x}, \mathbf{c}) + d(\mathbf{c}, \mathbf{c}^n) \; (\text{from} \: \triangle \: \text{inequality})\\
\because \; \frac{1}{2}d(\mathbf{c}, \mathbf{c^n}) &=& s(\mathbf{c}) \\ 
d(\mathbf{c}, \mathbf{c}^n) &=& 2s(\mathbf{c})\\
\therefore \; d(\mathbf{x}, \mathbf{c}^n) &\leq& d(\mathbf{x}, \mathbf{c}) + 2s(\mathbf{c}) \; (Using \: Eqs. \: 5 \: and \: 7)
\end{eqnarray}

\noindent \textbf{Case 1:} $d(\mathbf{x}, \mathbf{c}) < d(\mathbf{x}, \mathbf{c}')$\\
Assume $\frac{1}{2}(d(\mathbf{c}, \mathbf{c}')) > r(\mathbf{c}) + s(\mathbf{c})$,\\
$\because \; r(\mathbf{c}) \geq d(\mathbf{x}, \mathbf{c})$ (definition of radius), we can rewrite $\frac{1}{2}(d(\mathbf{c}, \mathbf{c}')) > r(\mathbf{c}) + s(\mathbf{c})$ as:
\begin{eqnarray}
    \frac{1}{2}d(\mathbf{c}, \mathbf{c}') &>& d(\mathbf{x}, \mathbf{c}) + s(\mathbf{c}) \\
    d(\mathbf{x}, \mathbf{c}) &<& \frac{1}{2}d(\mathbf{c}, \mathbf{c}') - s(\mathbf{c}) \label{eq:neighbor_eq0}
\end{eqnarray}
$\because$ $d(\mathbf{x}, \mathbf{c}) < \frac{1}{2}d(\mathbf{c}, \mathbf{c}') - s(\mathbf{c})$, for a positive constant $\epsilon > 0$, we can rewrite Eq. \ref{eq:neighbor_eq0} as $d(\mathbf{x}, \mathbf{c}) = (\frac{1}{2}d(\mathbf{c}, \mathbf{c}') - s(\mathbf{c})) - \epsilon$, if we consider $\epsilon = 0$, then $d(\mathbf{x}, \mathbf{c}) = (\frac{1}{2}d(\mathbf{c}, \mathbf{c}') - s(\mathbf{c}))$, if $\epsilon >0$, then it follows: $d(\mathbf{x}, \mathbf{c}) < \frac{1}{2}d(\mathbf{c}, \mathbf{c}') - s(\mathbf{c})$, 
\begin{eqnarray}
    d(\mathbf{x}, \mathbf{c}) &=& (\frac{1}{2}d(\mathbf{c}, \mathbf{c}') - s(\mathbf{c})) - \epsilon  \; (\epsilon > 0) \label{eq:neighbor_condition1}\\
    \frac{1}{2}d(\mathbf{c}, \mathbf{c}') &\leq& \frac{1}{2}d(\mathbf{x}, \mathbf{c}) + \frac{1}{2}d(\mathbf{x}, \mathbf{c}') \; (From \: Eq. \: 4)\label{eq:tr_ineq}
\end{eqnarray}
Substitute $d(\mathbf{x}, \mathbf{c})$ from Eq. \ref{eq:neighbor_condition1} in Eq. \ref{eq:tr_ineq}
\begin{eqnarray}
    \frac{1}{2}d(\mathbf{c}, \mathbf{c}') &\leq& \frac{1}{2} \left( \left( \frac{1}{2}d(\mathbf{c}, \mathbf{c}') - s(\mathbf{c}) -\epsilon \right) + \frac{1}{2}d(\mathbf{x}, \mathbf{c}') \right)\\ 
    d(\mathbf{x}, \mathbf{c}') &\geq& d(\mathbf{c}, \mathbf{c}') + s(\mathbf{c}) + \epsilon \label{eq:neighbor_condition_2}\\
    d(\mathbf{x}, \mathbf{c}') &>& d(\mathbf{x}, \mathbf{c}) \; (Using \: Eqs. \: \ref{eq:neighbor_condition1} \: \& \: \ref{eq:neighbor_condition_2})
\end{eqnarray}

If $\frac{1}{2}(d(\mathbf{c}, \mathbf{c}')) > r(\mathbf{c}) + s(\mathbf{c})$, then $\mathbf{x}$ is closer to $\mathbf{c}$ than $\mathbf{c}'$.

\noindent \textbf{Case 2:} $d(\mathbf{x}, \mathbf{c}^n) \: < \: d(\mathbf{x}, \mathbf{c}')$
\begin{eqnarray}
d(\mathbf{c}, \mathbf{c}') &\leq& d(\mathbf{x}, \mathbf{c}) + d(\mathbf{x}, \mathbf{c}') \; (\text{From Eq.} \: 4)\label{eq:neighbor_condition3}\\
2d(\mathbf{x}, \mathbf{c}) + 2s(\mathbf{c}) < d(\mathbf{c}, \mathbf{c}') &\leq& d(\mathbf{x}, \mathbf{c}) + d(\mathbf{x}, \mathbf{c}') \; (Using \: Eqs.\ \: \ref{eq:neighbor_eq0} \: \& \: \ref{eq:neighbor_condition3})\\
2d(\mathbf{x}, \mathbf{c}) + 2s(\mathbf{c}) &<& d(\mathbf{x}, \mathbf{c}) + d(\mathbf{x}, \mathbf{c}')\label{eq:neighbor_cond56}\\
d(\mathbf{x}, \mathbf{c}') &>& d(\mathbf{x}, \mathbf{c}) + 2s(\mathbf{c})\label{eq:neighbor_condition4}\\
d(\mathbf{x}, \mathbf{c}^n) &\leq& d(\mathbf{x}, \mathbf{c}) + 2s(\mathbf{c}) \; (\text{From Eq.} \: 8) \label{eq:neighbor_condition5}\\
\therefore d(\mathbf{x}, \mathbf{c}') &>& d(\mathbf{x}, \mathbf{c}^n) \; (\text{Using Eqs.} \: \ref{eq:neighbor_condition4} \: \& \: \ref{eq:neighbor_condition5})
\end{eqnarray}

If $\frac{1}{2}d(\mathbf{c}, \mathbf{c}') > r(\mathbf{c}) + s(\mathbf{c})$, then $\mathbf{c}'$ can not be the neighbor of $\mathbf{c}$, because $d(\mathbf{x}, \mathbf{c}) \: < \: d(\mathbf{x}, \mathbf{c}')$, and $d(\mathbf{x}, \mathbf{c}^n) \: < \: d(\mathbf{x}, \mathbf{c}')$.
\hfill $\blacksquare$
\end{lemma}

As corollary to Lemma \ref{lemma:neighborhood_lemma_1}, a centroid $\mathbf{c}'$ is a neighbor of $\mathbf{c}$ if and only if $\frac{1}{2}(d(\mathbf{c}, \mathbf{c}')) \leq r(\mathbf{c}) + s(\mathbf{c})$. This follows directly from the definition of neighboring clusters (Definition 3.0.2) and Lemma~\ref{lemma:neighborhood_lemma_1}.

\input{Algorithm1}

\subsection{Effectively Determining HE}

Previous work \citep{KurbanKmeans2017}, \citep{KurbanExp2021}, \citep{SHARMA2022100944} leverages probabilities to arrange data within heap nodes, effectively placing HE data in the leaves to expedite retrieval during program execution. While using the ordered data structure helps to segregate data into HE and LE, it does not facilitate the dynamic treatment to managing the size of HE data, opting instead for a fixed threshold. \kg{} is distinct in following ways: 1) utilize scalar projection to discern HE data rather than using sorted data, 2) instead of fixed threshold, determination of HE data occurs dynamically at runtime, 3) independent of any data structure’s implicit constraints (unlike other methods), and 4) \kg{} does not rely on distance bounds, relieving it from the need of maintaining and validating the bounds. In the following sections, we elucidate fundamental characteristics of \kg{} that connect DC, scalar projection, and the \textsf{HE} together. 

\subsection{Low expressive data}
\begin{figure}[t]
\centering\
\includegraphics[scale=0.45]{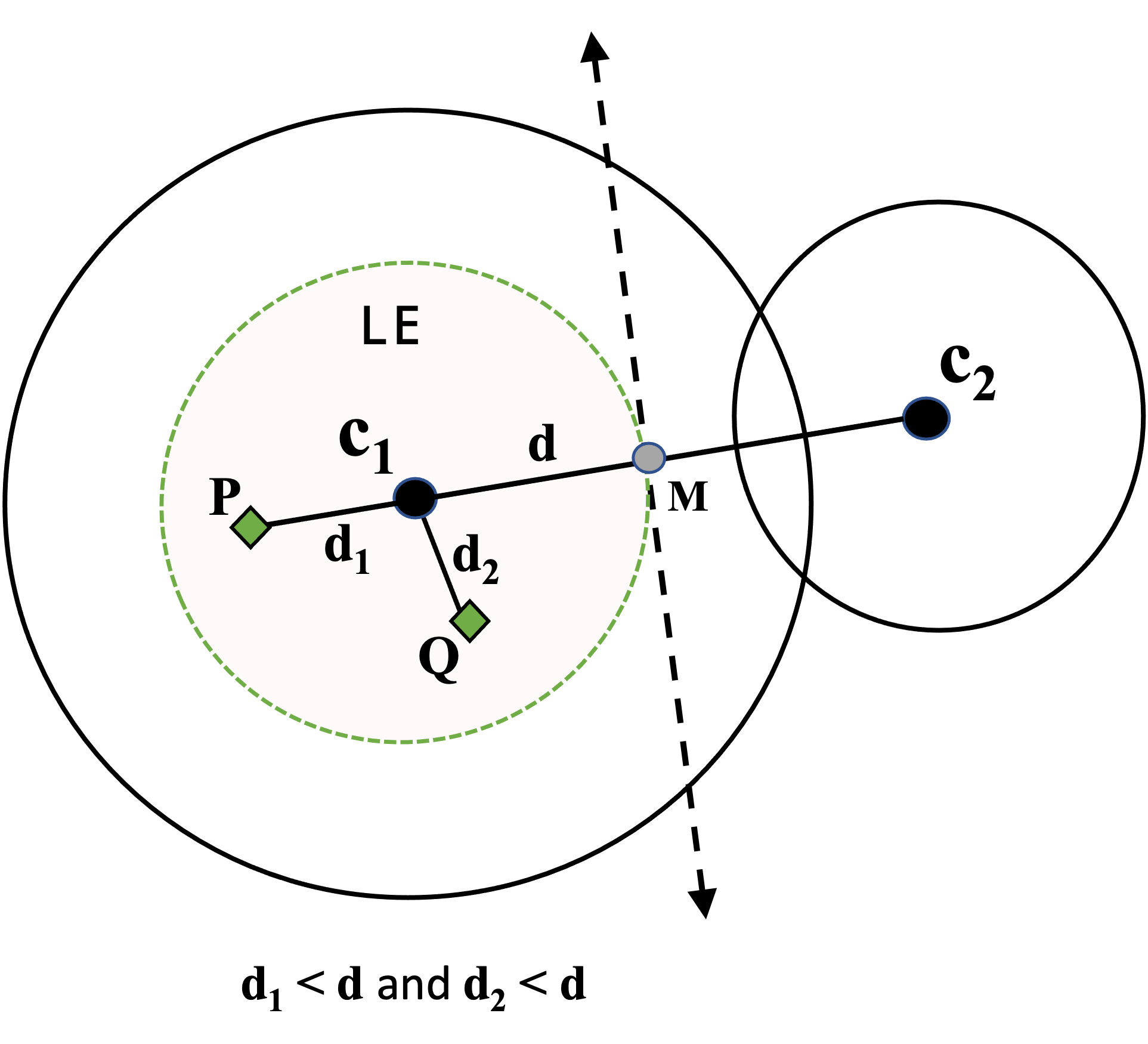}
\caption{\textbf{Illustrating \textsf{LE}}: $\mathbf{c}_2$ is the closest neighbor of $\mathbf{c}_1$, \textbf{M} denote the center of line segment $\overline{\mathbf{c}_1\mathbf{c}_2}$, and $\mathbf{d} = \frac{1}{2}d(\mathbf{c_1}\mathbf{c}_2)$. Data points $\mathbf{P}$ and $\mathbf{Q}$ are \textsf{LE} because their distance to $\mathbf{c}_1$ is less than half the distance between $\mathbf{c}_1$ and $\mathbf{c}_2$.} \label{fig:low_expressive}
\end{figure}

We show that distance computations concerning low expressive data can be disregarded, because LE data points maintain their assigned membership without alteration. Hence, it is prudent to solely compute distances pertinent to potential HE data points. An illustration of LE points is shown in Fig.~\ref{fig:low_expressive}.


\begin{lemma}[distance and \textsf{LE}]\label{lemma1}
A data point $\mathbf{x}$ assigned to $\mathbf{c}_i$ is defined as $\mathsf{LE}$, $\exists \mathbf{c}_j \in n(\mathbf{c}_i)$, if $d(\mathbf{x}, \mathbf{c}_i) < \frac{1}{2} d(\mathbf{c}_i, \mathbf{c}_j)$. Such a $\mathsf{LE}$ data point $\mathbf{x}$ cannot change its membership and will remain assigned to $\mathbf{c}_i$; consequently, a distance computation is not required. Notice that among all neighbors of $\mathbf{c}_i$, the closest one is the natural candidate to check whether there exists a $\mathbf{c}_j$ confirming our definition of \textsf{LE}. Therefore, it is enough to check the LE condition on the closest neighboring centroid $\mathbf{c}_j$ of $\mathbf{c}_i$.
\end{lemma}

\noindent \textbf{\textsc{Proof}} The lemma has been previously established by Elkan. We share our version of the proof in the appendix \ref{sec:app_le_proof}.\\

\begin{lemma}[distance and \textsf{LHE}]\label{lemma2}
A data point $\mathbf{x}$ assigned to $\mathbf{c}_i$ is likely high expressive or $\mathsf{LHE}$, if $d(\mathbf{x}, \mathbf{c}_i) > \frac{1}{2} d(\mathbf{c}_i, \mathbf{c}_j) \: \forall \mathbf{c}_j \in n(\mathbf{c}_i)$. $\mathsf{LHE}$ points may or may not change membership and require a distance computation. Unlike $\mathsf{LE}$, checking for $\mathsf{LHE}$ is done with every neighbor of $\mathbf{c}_i$. This is discussed in \ref{sec:proof_he_case}.
\hfill $\blacksquare$
\end{lemma}

    


\begin{figure}[t]
\centering\
    \includegraphics[scale=0.45]{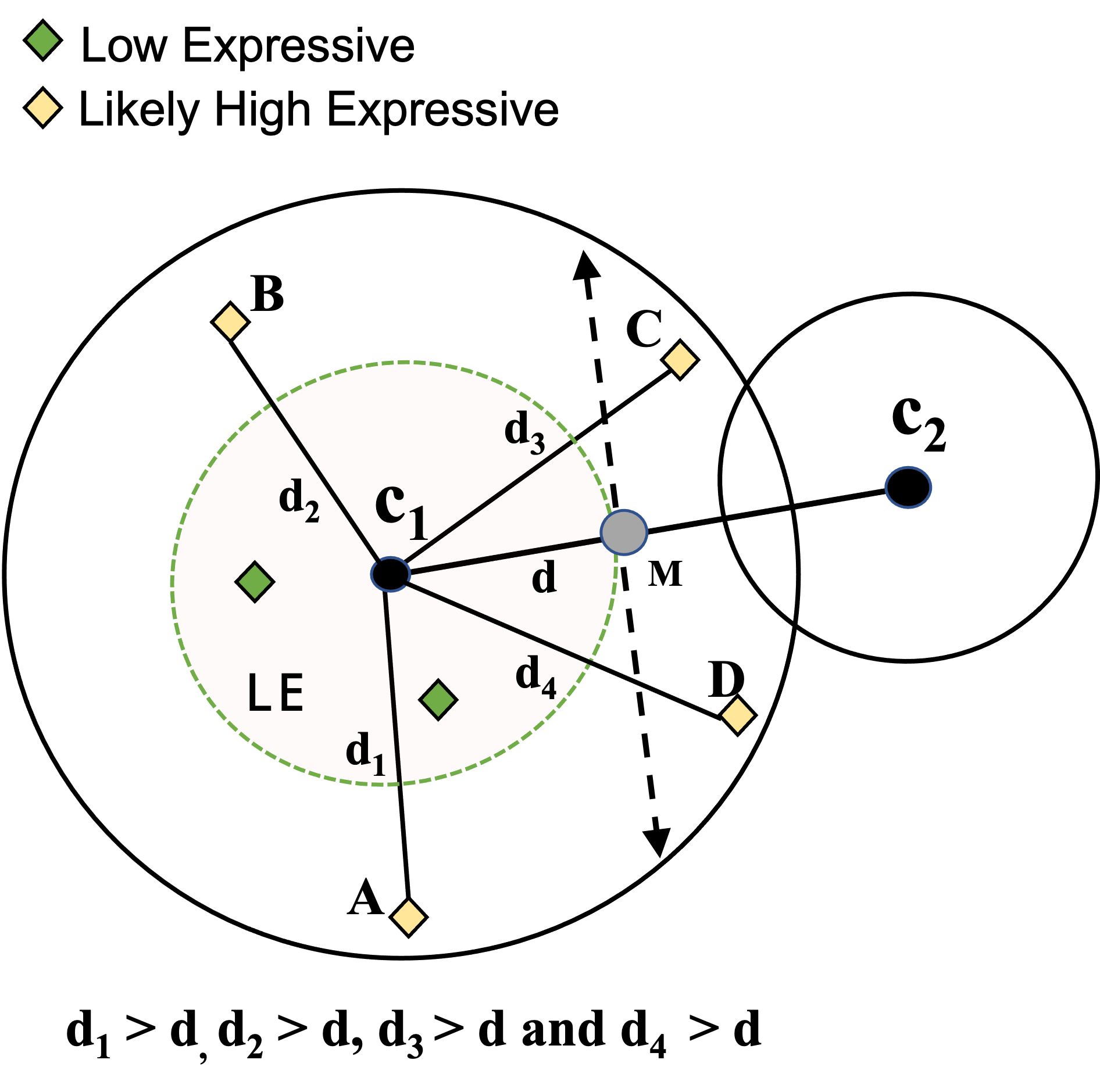}
    
    \caption{\textbf{Illustrating \textsf{LE} and \textsf{LHE}}: $\mathbf{c}_2$ is the closest neighbor of $\mathbf{c}_1$, \textbf{M} denote the center of $\overline{\mathbf{c}_1\mathbf{c}_2}$. $\frac{1}{2}d(\mathbf{c}_1\mathbf{c}_2) = \mathbf{d}$. Distance between $\mathbf{c}_1$ and points $\mathbf{A, B, C, D}$ is $\mathbf{d}_1, \mathbf{d}_2, \mathbf{d}_3, \mathbf{d}_4$, respectively. (a) Characterization of \textsf{LHE}: Per Lemma \ref{lemma2}, $A$, $B$, $C$ and $D$ are \textsf{LHE}.}
    \label{fig:twoneighbors_le_lhe}
\end{figure}


Lemmas\,\ref{lemma1} and \ref{lemma2} identify data as LE or LHE based solely on distance, but remain unaware of the orientation of data. More specifically, not all LHE points may change their membership. To explain this further, consider the example shown in Fig.\,~\ref{fig:twoneighbors_le_lhe}. In the shown spatial configuration, data points $A, B, C$, and $D$ are assigned to $\mathbf{c}_1$. By Lemma \ref{lemma2}, all four points will be marked as LHE, and will force distance computation. However, its apparent that $A$ and $B$ are already closer to $\mathbf{c}_1$, and can be safely removed from the distance calculation (as they won't change their membership). As a result, only $C$ and $D$ require distance computation. The important question is: how to differentiate the points $C$ and $D$ from $A$ and $B$, without resorting to distance calculations? To this end, we propose a novel methodology employing vector affine product to discern $C$ and $D$ (HE data) and integrate knowledge of the data points' orientation. This concept is formalized below. 

\subsection{High expressive data}
We underscore the significance of \kg{} by showing that the orientation, or direction of data, serves as a reliable indicator for predicting changes in membership status, obviating the need for redundant distance computations.

\begin{lemma}[scalar projection and \textsf{HE}]\label{lemma3}
For a pair of centroids $\mathbf{c}_i, \mathbf{c}_j$ where $\mathbf{c}_j \in n(\mathbf{c}_i)$.  
A data point $\mathbf{x}$ assigned to $\mathbf{c}_i$ is $\mathsf{HE}$ when:
 \begin{eqnarray}
\overrightarrow{\mathsf{m}(\mathbf{c}_i, \mathbf{c}_j)\mathbf{x}}\cdot \overrightarrow{\mathsf{m}(\mathbf{c}_i, \mathbf{c}_j)\mathbf{c}_j} &>& 0 \label{eq:scalar}
  \end{eqnarray}
\noindent \textbf{\textsc{Proof}}\\ For angle $\theta$ and vectors $\mathbf{a},\mathbf{b}$ 
\begin{eqnarray}
    \cos(\theta) &=& \frac{\mathbf{a}\cdot\mathbf{b}}{||\mathbf{a}||\, ||\mathbf{b}||} 
\end{eqnarray}
We then define scalar $s$ as:
\begin{eqnarray}
    s &=& ||\mathbf{a}||\cos(\theta)  
    = \frac{\mathbf{a}\cdot \mathbf{b}}{||\mathbf{b}||}\label{eq:first}\\
    \mathsf{sign}(s) &=& \left\{\begin{array}{ll}0^o \leq \theta \leq 90^o &  > 0\\ 
    90^o < \theta \leq 180^o & \leq 0 \end{array}\right.\label{eq:zero}
\end{eqnarray}
from the definition of $\cos$. \\
Using assumption \ref{eq:scalar} along with Eq.\,\ref{eq:zero}, we have $cos(\theta) > 0$, so $0^o \leq \theta \leq 90^o$. Hence, $\overrightarrow{\mathsf{m}(\mathbf{c}_i, \mathbf{c}_j)\mathbf{x}}$ and $\overrightarrow{\mathsf{m}(\mathbf{c}_i, \mathbf{c}_j)\mathbf{c}_j}$ are oriented in the same direction; $\mathbf{x}$ lies in the quadrants nearest to the neighboring centroid which makes the projection positive. Therefore, $\mathbf{x}$ is $\mathsf{HE}$. \hfill $\blacksquare$
\end{lemma}

The simplicity of Lemma\,\ref{lemma3} belies its importance: the vector affine product is practically much faster to compute, compared to distance calculations, despite both having the same asymptotic complexity (comparative analysis in \ref{sec:app_cost_dist_comp_vs_scal_proj}). In essence, \kg{} applies successive filtering to reduce distance computations by, 1) confining distance calculations exclusively to neighboring centroids as opposed to considering all centroids, 2) bypassing low expressive data, and 3) discerning HE points from LHE, and conducting distance computations only for HE. The result is significant enhancement in efficiency while preserving quality of result. The pseudo-code for \kg{} is given in Alg. ~\ref{algo:Geokmeans}.

\begin{figure}[t]
    \centering
    \includegraphics[scale=0.50]{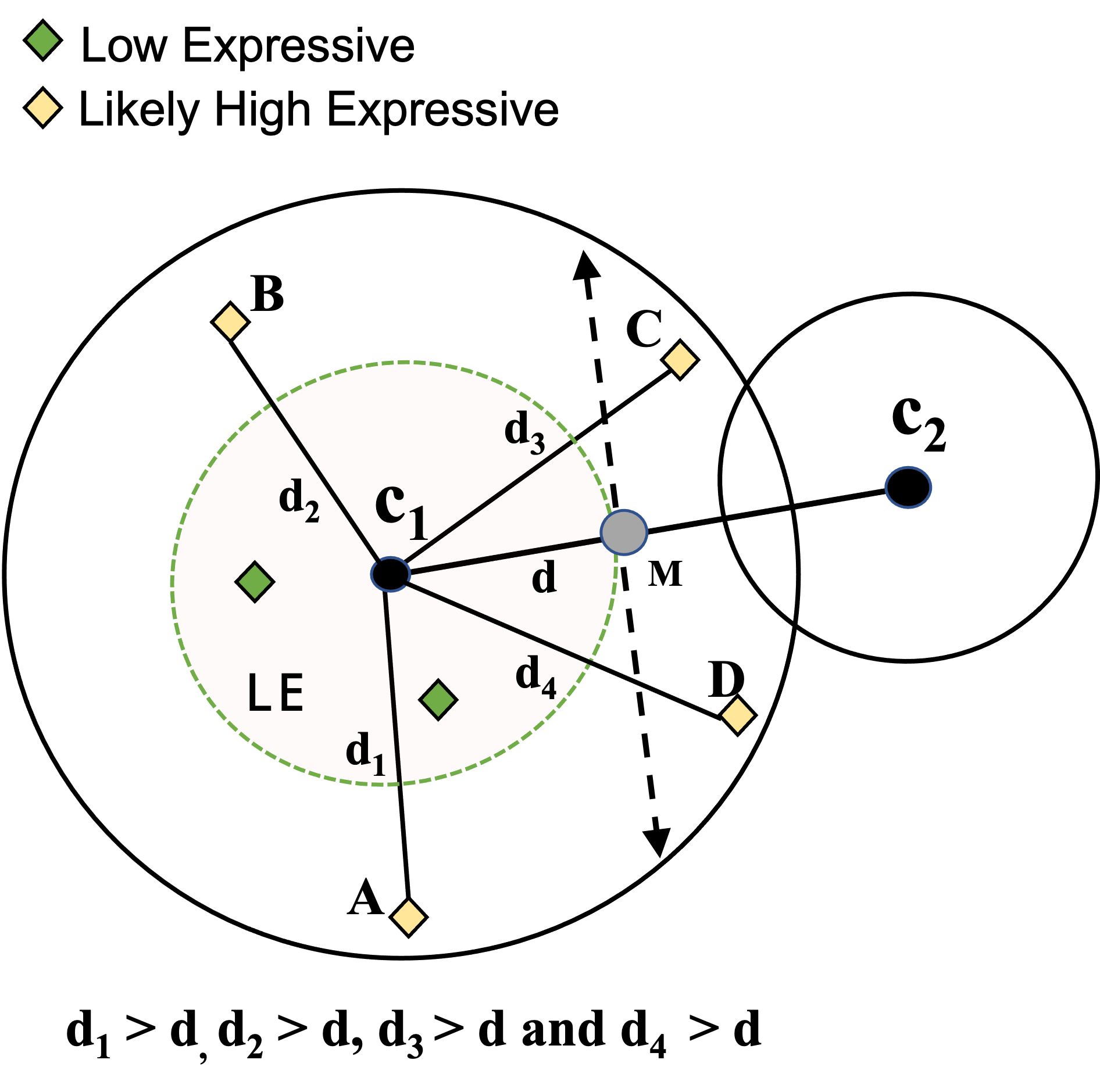}
    
    \caption{\textbf{Illustrating \textsf{LHE} and \textsf{HE}}: $\mathbf{c}_2$ is the closest neighbor of $\mathbf{c}_1$, \textbf{M} denote the center of $\overline{\mathbf{c}_1\mathbf{c}_2}$. $\frac{1}{2}d(\mathbf{c}_1\mathbf{c}_2) = \mathbf{d}$. Distance between $\mathbf{c}_1$ and points $\mathbf{A, B, C, D}$ is $\mathbf{d}_1, \mathbf{d}_2, \mathbf{d}_3, \mathbf{d}_4$, respectively. Not all \textsf{LHE} points qualify as \textsf{HE}. Only $C$ and $D$ are valid \textsf{HE}, because the angle subtended from $\protect \overrightarrow{\rm M\mathbf{c}_2}$ to $\protect \overrightarrow{\rm MC}$ and $\protect \overrightarrow{\rm MD}$ is $< 90^{\circ}$. Conversely, \textsf{LHE} points $A, B$ can be regarded as \textsf{LE} because angle between vectors $\protect \overrightarrow{\rm M\mathbf{c}_2}$ and $\protect \overrightarrow{\rm M\mathbf{B}}$,  $\protect \overrightarrow{\rm M\mathbf{A}}$ is $> 90^{\circ}$.}
    \label{fig:twoneighbors_lhe_he}
\end{figure}

We refer to Fig.\,\ref{fig:twoneighbors_lhe_he} and revisit the question of differentiating between LHE and HE without resorting to distance calculations. In this instance, the differentiation can be discerned through the orientation of data within the native space; specifically, LHE points $A$ and $B$ are oriented towards $\mathbf{c}_1$ and hence, are as good as LE, whereas $C$ and $D$ are oriented towards $\mathbf{c}_2$, thus are HE, since they are poised to change their membership, subsequently impacting centroid computations. We posit that this insight represents a significant stride toward devising more efficient variants of \km{}. While existing methods employing distance bounds are effective, they make use of distances, therefore, are susceptible to performing more computations owing to the presence of LHE points. Experimental evaluation demonstrate that, compared to other methods, \kg{} reduce more distance computations. In essence, the objective is to actively divert the majority of calculations for HE data, while bypassing LE/LHE data (impact of the algorithmic components of \kg{} on DC is shared in ~\ref{app:profiling_optimization}).

\input{Algorithm2}

\subsection{Proof of Symmetry \& Convergence}
We demonstrate that \kg{} is symmetrical with \km{} i.e. when initialized with the same centroids, \kg{} and \km{} will assign data points to identical clusters and converge to the same solution, despite \kg{} performing fewer distance computations. This property ensures that \kg{} preserves the clustering quality of \km{} while improving computational efficiency. Note that \kg{} is similar to \km{} except that the distance computations are only done for HE data. The proof consists of two parts: first, we show that \kg{} will converge in a finite number of steps. In the second part, we describe steps that both algorithms share, and prove that assignment to clusters is symmetric. While the first part provides important assurance about the algorithm's behavior during execution, the second part confirms that \kg{} reach the same solution as \km{}.


\input{proof_of_convergence}

\input{proof_of_symmetry}

It follows from the proof of convergence and symmetry that, a) \kg{} will converge, and b) it converges to the same local optimum as \km{}.


\input{complexity_analysis}

%% file: Algorithm1.tex
\begin{algorithm}[!htb]
\SetKwInOut{Input}{Input}
\SetKwInOut{Output}{Output}

\Input{centroids $\mathbf{C}$, vector $s^k$, matrix $M^{k \times k}$, radii $R = \{ r(\mathbf{c}_1), \ldots, r(\mathbf{c}_k)\}$}
\tcp{$s_i$ = half the distance between $\mathbf{c}_i$ and the nearest centroid}
\tcp{$r_i = r(\mathbf{c}_i)$}
\tcp{$M^{k \times k}$ stores inter-centroid distances}
\BlankLine

\Output{Neighbors $n = \{n(\mathbf{c}_i),\ldots, n(\mathbf{c}_k)\}$, $M$, $MID^{k \times k}$, $A^{k \times k}$}
\BlankLine
\tcp{$n(\mathbf{c}_i)$: neighbors of $\mathbf{c}_i$}
\tcp{$MID^{k \times k}$: matrix to store midpoint coordinates, $A^{k \times k}$: matrix to store affine vectors}

\tcp{Initialize all values in $s$ to $\infty$}
\BlankLine

\For{$i \leftarrow 1$ \KwTo $k$} {

    $M_{ii} = \infty$
    \BlankLine
    
    \For{$j \leftarrow i+1$ \KwTo $k$} {
        
        \BlankLine
        $dist \gets \frac{1}{2} \; d(\mathbf{c}_i, \mathbf{c}_j)$\\
        $M_{ij} \gets dist$\\
        $M_{ji} \gets M_{ij}$\\
        
        // Find nearest centroid\\
        \If{$dist < s_i$}{
            $s_i \gets dist$
        }
        \If{$dist < s_j$}{
            $s_j \gets dist$
        } 
    }
}
    
\For{$i \leftarrow 1$ \KwTo $k$} {

    \For{$j \leftarrow 1$ \KwTo $k$} {
        
        \BlankLine
        \If{$i \neq j \; \& \; M_{ij} \leq r_i + s_i$}{
            Add $\mathbf{c}_j$ to $\mathbf{n}(\mathbf{c}_i)$

            $MID_{ij} \gets \mathsf{m}(\mathbf{c}_i, \mathbf{c}_j)$
            \BlankLine
            $MID_{ji} \gets MID_{ij}$
            \BlankLine
            $A_{ij} \gets \overrightarrow{MID_{ij}\mathbf{c}_j}$
        }
    }
}

\Return{$n, M, MID, A$}
\caption{Neighbors}
\label{algo:neighbors_finding}
\end{algorithm}

%% file: Algorithm2.tex
\begin{algorithm}[!htbp]
\scriptsize
\SetKwFunction{UpdateCentroids}{UpdateCentroids}
\SetKwFunction{CheckConvergence}{CheckConvergence}
\SetKwFunction{findClosest}{findClosest}
\SetKwFunction{distance}{distance}
\SetKwFunction{length}{length}
\SetKwInOut{Input}{Input}
\SetKwInOut{Output}{Output}

\Input{$D$: Data, $k$: clusters, $T$: max iterations, $\epsilon$: convergence threshold, $s$}
\Output{$\mathbf{A}$: Clusters, $\mathbf{C}$: Centroids}

$t \gets 0$

Randomly assign $k$ data points as initial centroids in $\mathbf{C}$

For data $\mathbf{x}_{i \in {1,\ldots,m}}$, find their closest centroid $\mathbf{c}_k$; assign $\mathbf{x}_i$ to $\mathbf{c}_k$; update $\mathbf{a}_k$

\BlankLine
Initialize radius as shown in Def. \ref{def:radius}

\BlankLine
\While{$t < T$}{
    
    \BlankLine
    \tcp{update centroids}
    \For{$i \leftarrow 1$ \KwTo $k$}{
        Calculate $\mathbf{c}_i^{new}$ as the centroid of all data points assigned to cluster $\mathbf{a}_i$
    }
    
    \BlankLine
    If $d(\mathbf{C}, \; \mathbf{C}^{new}) \leq \epsilon$ ; break

    \BlankLine
    \tcp{Find neighbors}
    $n, M, MID, A \gets \mbox{Neighbors}(\mathbf{C}^{new}, s, M, R)$

    \BlankLine
    \tcp{reset the radius}
    $r(\mathbf{c}_{i} = 0) \;|\; i \in \{1,\ldots,k\}$
    
    \BlankLine
    \For{$i \leftarrow 1$ \KwTo $m$}{

        \BlankLine
        $j = \mathcal{I}(\mathbf{x}_i)$


        \BlankLine
        \tcp{if $\mathbf{x}$ is LE; continue (Lemma \ref{lemma1})}
        \If{$d(\mathbf{x}_i, \mathbf{c}_j) \leq s(\mathbf{c}_j)$}{
            continue
        }
            
        \BlankLine
        \tcp{Loop through neighbors of $\mathbf{c}_j$}
        \For{$\mathbf{c}_l \in n(\mathbf{c}_j)$}{

            \BlankLine
            \tcp{Check if $\mathbf{x}$ is LHE; Lemma \ref{lemma2}}
            \If{$d(\mathbf{x}_i, \mathbf{c}_j) > M_{il}$}{

            \BlankLine
             \tcp{Check if $\mathbf{x}$ is HE; Lemma \ref{lemma3}}
             \If{$\overrightarrow{MID_{il}\mathbf{x}_i} \cdot \overrightarrow{MID_{il}\mathbf{c}_l} > 0$}
            {
                Compute $d(\mathbf{x}_i, \mathbf{c}_l)$; if $\mathbf{x}_i$ is closer to $\mathbf{c}_l$; assign $\mathbf{x}_i$ to $\mathbf{c}_l$\\
                update $\mathbf{a}_l$;
                If $d(\mathbf{x}_i, \mathbf{c}_l) > r(\mathbf{c}_l)$; set $r(\mathbf{c}_l) = d(\mathbf{x}_i, \mathbf{c}_l)$
            }  
                
            }
        }  
    }
    
    \BlankLine
    $\mathbf{C} = \mathbf{C}_{new}$
    \BlankLine
    t $\gets$ t + 1
}
\Return{$\mathbf{A}$, $\bm{C}$}
\caption{\kg{}}
\label{algo:Geokmeans}
\end{algorithm}

%% file: proof_of_convergence.tex
\subsubsection{Convergence Properties}
To establish that \kg{} will necessarily converge, we first show that $SSE$ is minimized, when the center associated with each cluster is the actual mean (centroid) of data assigned to the cluster. We use this result in the next part of this proof.

\input{convergence_proof_1}

%% file: convergence_proof_1.tex
\begin{lemma}\label{lemma:convergence_proof_part1}
Given a set of data points $D = \{ \mathbf{x}_{1}, \ldots, \mathbf{x}_m \}$ and $m > 1$. The actual centroid of points in $D$ can be found as: $\mathbf{c} = \frac{1}{m} \sum_{i=1}^m \mathbf{x}_i$, let $\mathbf{x}_p$ be any arbitrary point in $D$. We show that $SSE(D, \mathbf{c}) \leq SSE(D, \mathbf{x}_p)$.

\noindent \textbf{\textsc{Proof}} \kg{} inherits the convergence properties of \km{}, which are well studied in the literature. For brevity, we defer the proof sketch to Appendix \ref{app:convergence_proof}.

\end{lemma}

%% file: proof_of_symmetry.tex
\subsubsection{Symmetry}
While convergence has been established, we now need to show that this convergence leads to the same solution as \km{}. For \kg{}, we denote the centroids and cluster assignments using the new notation, \textit{e.g.}, $\underline{\mathbf{C}}$ and $\underline{\mathbf{A}}$, to distinguish them from the corresponding centroids and cluster assignments of \km{}.\\

\begin{enumerate}    
    \item \textbf{\textit{Initialize:}} Initialize $k$ centroids for both algorithms as follows:  $\mathbf{c}_{1}^{(1)}, \mathbf{c}_2^{(1)}, ..., \mathbf{c}_{k}^{(1)}$ and $\underline{\mathbf{c}_{1}}^{(1)}, \underline{\mathbf{c}_{2}}^{(1)} , ..., 
    \underline{\mathbf{c}_{k}}^{(1)}$, where $\underline{\mathbf{c}_{i}}^{(1)} \gets \mathbf{c}_i^{(1)}$ for $1 \leq i \leq k$.
    
    \item \textbf{\textit{Assign data:}} Calculate pairwise distances between data and centroids-assign data to its nearest centroid. This step is the same in both \km{} and \kg{}, so we dispense with details here.
    
    \item \textbf{\textit{Update centroids:}} The update is based on the assignments in step 2); thus $\underline{\mathbf{c}_{i}}^{(2)} = \mathbf{c}_i^{(2)}$ for $1 \leq i \leq k$.
    
    \item \textbf{\textit{Re-assign data:}} At this step, the algorithms begin to diverge, particularly \km{} will recalculate distances for all data, but \kg{} will only calculate distances for HE data and assign them to the nearest centroid. Consider a cluster represented by the centroid $\mathbf{c}_i$. Notice that $\mathbf{a}_i = \underline{\mathbf{a}_i}$ for $1 \leq i \leq k$, due to step-3. After distance computations, the results can differ only if \kg{} assigns data differently than \km{}. At this point, observe that by Theorem \ref{proof:symmetry}, this is not possible, and therefore, both \km{} and \kg{} will produce exactly the same assignments at each step, thus resulting in identical results.\\
    
\begin{theorem}[Symmetry of assignment of data to centroids]\label{proof:symmetry}
There can be no data point that is assigned to different clusters by \km{} and the \kg{}. 

\noindent \textbf{\textsc{Proof}}\\ 
A data point $\mathbf{x}$ assigned to $\mathbf{c}_i$ can be labeled as either HE, LHE or LE by \kg{}, and we examine each case below.
    
\noindent \textbf{Case 1:} $\mathsf{HE}(\mathbf{x})$ 
Since $\mathsf{HE}(\mathbf{x})$, both \km{} and \kg{} will compute distance identically. The assignment will be updated as per step 2 above.

\noindent \textbf{Case 2:} $\mathsf{LHE}(\mathbf{x})$ 
From the construction of algorithm, if $\mathsf{LHE}(\mathbf{x})$, then its checked if such a point will change its membership (HE) or not (LE). If $\mathbf{x}$ is found to be \textsf{HE}, then by \textbf{Case 1}, assignment is identical to \km{}. If $\mathbf{x}$ is \textsf{LE}, then assignment will be identical as explained in the following \textbf{Case 3}.
    
\noindent \textbf{Case 3:} $\mathsf{LE}(\mathbf{x})$ \\ \kg{} will not recompute distance for $\mathbf{x}$, membership is unchanged. $\mathbf{x}$ cannot be assigned to a different cluster $\mathbf{c}_z \neq \mathbf{c}_i$ by \km{}. 
    
\noindent{\textsc{Proof by Contradiction}\\ 
From triangle inequality
\begin{eqnarray}
    d(\mathbf{x}, \mathbf{c}_i) +\; d(\mathbf{x}, \mathbf{c}_z) &\geq& d(\mathbf{c}_i, \mathbf{c}_z)
\label{triangle}
\end{eqnarray}

Assume that \km{} assign $\mathbf{x}$ differently than \kg{} i.e. let's assume that $\mathbf{x}$ is assigned to $\mathbf{c}_z$, then:

\begin{equation}
    d(\mathbf{x}, \mathbf{c}_z) < d(\mathbf{x}, \mathbf{c}_i)\label{eq:step1}
\end{equation}

Since, $\mathbf{x}$ is $\mathsf{LE}$, hence by Lemma \ref{lemma1}:
\begin{equation}
    d(\mathbf{x} ,\mathbf{c}_i) < \frac{1}{2} d(\mathbf{c}_i, \mathbf{c}_z) \label{eq:step2}
\end{equation}

From Eqs.\ \ref{eq:step1} and \ref{eq:step2}   
\begin{equation}
    d(\mathbf{x}, \mathbf{c}_z) < \frac{1}{2} d(\mathbf{c}_i, \mathbf{c}_z)\label{eq:step3}
\end{equation}

Add Eqs.\ \ref{eq:step2} and \ref{eq:step3}   
\begin{equation}
    d(\mathbf{x}, \mathbf{c}_i) + d(\mathbf{x}, \mathbf{c}_z) < \; d(\mathbf{c}_i, \mathbf{c}_z) \label{result}
\end{equation}
        
Eq. \ref{result} contradicts the law of triangle inequality as stated in Eq.\ \ref{triangle}, more specifically Eqs. \ref{triangle}-\ref{result} forms a contradiction; thus, membership is the same and \km{} won't assign $\mathsf{LE}$ data differently than \kg{}. \hfill $\blacksquare$
}\end{theorem}
    
\item \textbf{\textit{Continue with the next iterations:}} After the second iteration, both algorithms will assign data points in exactly the same way; as a result, $\underline{\mathbf{c}_{i}}^{(3)} = \mathbf{c}_{i}^{(3)}$ for $1 \leq i \leq k$. Iterations continue, symmetry will hold until convergence.
\end{enumerate}

%% file: complexity_analysis.tex
\subsection{Complexity}\label{sec:complexity_analysis}

To partition $m$ points in $k$ clusters, the per iteration complexity of \km{} is $\mathcal{O}(mkd)$. Alg.~\ref{algo:Geokmeans} extends \km{} by including:

\begin{enumerate}
    \item Computing neighbors for each centroid (line 10)
    \item Identifying low expressive ($\textsc{LE}$) data points (line 14)
    \item Identifying high expressive ($\textsc{HE}$) data points using scalar projection (line 19)
    \item Performing distance computation only between $\textsc{HE}$ data and their neighbor clusters (line 20)
\end{enumerate}

Our implementation of neighbor finding (Alg. ~\ref{algo:neighbors_finding}) consists of two nested loops over $k$, thus leading to per iteration complexity of $\mathcal{O}(k^2d)$. If we denote the average number of neighbor centroids by $n$ and the average number of $\textsc{HE}$ data points by $m'$, distance computations between $\textsc{HE}$ data and neighbor centroids will cost $\mathcal{O}(m'nd)$. Additionally, to determine if a data point is $\textsc{LE}$, we compute the distance to its assigned centroid, this costs $\mathcal{O}(md)$. We can add the cost of finding neighbors, determining $\textsc{LE}$, and performing DC between $\textsc{HE}$ and neighbor centroids to obtain: $\mathcal{O}(k^2d + m'nd + md)$. As \kg{} proceeds, the centroids become stable and an increasingly large fraction of data become $\textsc{LE}$; by complement $\textsc{HE}$ data decreases, hence $m' \lll m$; empirical evaluations confirms this belief. In comparison to unbounded \kb{}, if we denote the average number of neighbor clusters by $n$ and average number of data points for which \kb{} perform distance computations by $m''$ then time complexity of \kb{} is $\mathcal{O}(k^2d +  + nm''d + md)$ (ignoring the cost of sorting due to lower asymptotic growth). Experimental evaluations show that $m'' \gg m'$, and in some cases, $m''$ is one to two orders of magnitude more than $m'$. As a result, depending on the data, complexity of \kb{} could be higher than \kg{}. Per iteration runtime and space complexities of various algorithms is listed in Table \ref{tab:complexity_costs}. Time complexity of \km{} and \ham{} is higher than \kg{}, and while \ann{} and \exp{} offer more pragmatic speed-up, their time complexity is higher than other methods. \kd{} offers more competitive approach, however, its bound depends on expansion constant $c_k$ (measure of intrinsic data dimensionality) and tree imbalance. Performance is limited when the tree become imbalanced, or specifically on high dimensional data \citep{curtin2017dual}. This is also observed in experimental results. 

\kg{} requires $\mathcal{O}(k^2d)$ space for storing midpoint coordinates and affine vectors. Cost of storing data is $\mathcal{O}(md)$ and $\mathcal{O}(kd)$ space is needed to store the centroids. The space complexity of \kg{} is $\mathcal{O}(md + k^{2}d)$ (this subsumes the cost of storing centroids). In practice, however, the effective space cost is typically lower than the theoretical upper bound of $O(md + k^2d)$. This is because centroids rarely form a fully connected neighborhood graph - most centroids are neighbors to only a small subset of other centroids. If we denote $n$ as the maximum number of neighbors per centroid (where typically $n < k$), the amortized space complexity of \kg{} becomes $O(md + knd)$, which is competitive with other methods. We empirically validated this by comparing the memory usage of \kg{} with distance based algorithms. Our findings (Table~\ref{tab:table_space_usage}, appendix), show that the memory consumption of \kg{} is similar to its counterparts.

\input{Table1}

%% file: Table1.tex
\begin{table}[!htb]
    \centering
    \caption{Runtime and Space Complexity (per iteration)}
    \label{tab:complexity_costs}

    \begin{tabular}{lrr}
    \toprule
    \textsf{Algorithm} & \textsf{Runtime} & \textsf{Space}\\
    \midrule
        
        \km{} & $\mathcal{O}(mkd)$ & $\mathcal{O}(md + kd)$ \\

        \ham{} & $\mathcal{O}(mkd + k^{2}d)$ & $\mathcal{O}(md + kd)$ \\

        \ann{} & $\mathcal{O}(mkd + k^{2}d)$ & $\mathcal{O}(md + kd)$ \\

        \kd{} & $\mathcal{O}(klog(k) + m)$ & $\mathcal{O}(md + kd)$ \\

        \exp{} & $\mathcal{O}(mkd + k^{2}d)$ & $\mathcal{O}(md + kd + k^{2})$ \\
        
        \kb{} &  $\mathcal{O}(md + nm''d + k^{2}d)$ & $\mathcal{O}(md + kd + k^{2})$  \\
        
        \kg{} &  $\mathcal{O}(md + nm'd + k^{2}d)$ & $\mathcal{O}(md + k^{2}d)$ \\

    \bottomrule\\
    \end{tabular}
\end{table}

%% file: Experiments.tex
\input{Table2}


\section{Experiments \& Analysis}\label{sec:Experiments}
Experiments are performed with varying $m$: size, $d$: dimension, and $k$: cluster number. Specifically, 20 data sets are used (10 real-world and 10 synthetically generated large data). Table~\ref{tab:data_details} provide details. In the following sections, \kg{} is compared with six well-known and exact \km{} algorithms. Lloyd's \km{} is used as baseline. We have selected the best known unbounded \km{} i.e. \kb{}, fastest bounded version-\exp{}, fast tree-based \km{} i.e. \kd{}, and two additional fast \km{} variants namely, \ann{} and \ham{}. \blklst{} is another fast and structure based \km{}, however \kd{} improves upon the design of \blklst{} and is shown to be empirically better \citep{curtin2017dual}, hence we have included \kd{} in the experiments. Also, Hamerly's algorithm is an extension of \citep{Elkan2003}, performing better in most cases \citep{Newling2016}, so we consider \ham{} here. Except \kb{} and \kd{}, all algorithms are implemented from scratch in C++. For \kb{} and \kd{}, we use the code published by their authors. As a novel exploration-impact of using faster \km{} alternatives on downstream energy consumption is also studied in Sec.\,\ref{sec:results_energy_usage}. Further details on the computing platform and data are found in \ref{sec:app_experimental_system} and \ref{sec:app_data_generation_process} (appendix).

\subsubsection{Implementation note}\label{sec:implementation_note}
All algorithms are initialized with the same centroids (randomly sampled from the data) in a given trial. Ten trials are done, and average statistics are reported. On real-world and scRNASeq data, maximum iterations are set to 500. On synthetic data, max iterations are set to 100. Execution stops when either maximum iterations are reached or centroids does not change, whichever is earlier.

\subsection{Experiments on Real Data}\label{sec:exp_real_data} 
Experiments are done by varying $k$ on real-world datasets. Since \kg{} result in same solution as \km{}, performance is evaluated by comparing the average DC and runtime. However, for completeness, appendix \ref{app:exp_real_data_accuracy} provide empirical results to show that \kg{} produces the same clustering assignments as \km{}. While our main experiments use random initialization for comparative consistency, we also evaluated all algorithms using $k$-means++ initialization, with complete results provided in ~\ref{app:kpp_comparison}. These results confirm that \kg{} maintains its performance advantages with $k$-means++ initialization, consistently achieving superior distance computation savings and competitive runtime performance across most datasets.

\subsubsection{Distance Computations}\label{sec:exp_real_data_distances} Table\,\ref{tab:table_random_distances} show the number of average distance computations and savings over baseline \km{} on different values of $k$. Not surprisingly, faster variants perform better than the baseline \km{} algorithm. However, for most datasets, \kg{} consistently outperformed other algorithms, and the performance gap widens as $m, k$ increases. On the larger Kddcup and Twitter data: \kg{} achieves a drastic reduction in distance computations up to 2-3 orders of magnitude less than \km{}. Both \kb{} and \exp{} also performed well and obtained a noticeable reduction in DC up to 1-2 orders of magnitude less than \km{}. However, the DC savings of \kg{} is $98-99\%$ more than \km{}, and it is better than \kb{} and \exp{}. On Sensor and Kegg data: \kg{} does the fewest DC, up to 1-2 orders of magnitude less than \km{}. DC done by \kb{} and \exp{} are also in a similar range, but \kg{} achieves much better speed-up.

The performance gap shrinks on Breastcancer and CreditRisk data because of the smaller data size. On Breastcancer, the DC of all algorithms (except \km{} and \ham{}) are in a similar range, but \kg{} attains the highest DC savings over baseline \km{}. On CreditRisk data, \kd{} performed the least number of distance computations and also saves the most DC. This is expected as tree-based approaches save more DC in low dimensions \citep{Pelleg1999}.

The variance in algorithmic performance stems from disparities in their underlying designs. When contrasted with unbounded methods such as \kb{} and \kg{}, bounded alternatives (\ham{}, \ann{} and \exp{}) perform more distance computations. Particularly, if the bounds are violated, \ham{} calculates the distance of a point with all $k$ centroids, whereas \ann{} does fewer computations by pruning the set of centroids based on annulus region centered on the origin, \exp{} reduces the number of candidate centroids by using a spherical region centered on the currently assigned centroid \citep{Newling2016}. \kb{} generally does less DC than its bounded counterparts, but akin to bounded algorithms, \kb{} is also agnostic to the orientation of a data point. Consequently, \kb{} does more work than \kg{}, which uses data orientation to reduce the DC further.

\input{Table3}

\input{Table4}


\subsubsection{Runtime}\label{sec:exp_real_data_runtime} 

Results showing the average runtime and speed-up of different algorithms over baseline \km{} are given in Table\,\ref{tab:table_random_runtime}. It is observed that on most datasets and $k$, \kg{} is faster than other algorithms, and runtime speed-up of \kg{} generally improves as $m, k$ increases. On Twitter: \kg{} has the best runtime speedup, and for $k=500$, the performance of \kg{} and \exp{} are very similar. On Kddcup data: both \kg{} and \exp{} perform similarly, \kg{} is slightly better for $k=300$ while \exp{} is better on $k=100, 500$. To determine the data orientation, \kg{} calculate the projection of a data point with neighboring centroids, which adds a minor overhead to the total computation time. In specific cases where the performance of \kg{} is similar to \exp{}, we note that both are similar in runtime per iteration (appendix, Table\,\ref{tab:table_runtime_per_iter}). For example, on Twitter and $k=500$, the runtime of a single iteration of \kg{} is $\approx 2.30$ seconds while that of \exp{} is $\approx 2.27$ seconds. On Kddcup, for $k=100$, an iteration of \kg{} and \exp{} took $0.447$ and $0.436$ seconds, respectively. For $k=500$, the runtime per iteration is $0.962$, $0.924$ seconds for \kg{} and \exp{}, respectively. In these rare instances, even though performance is similar, we note that \kg{} is more efficient because it does less DC.

On Sensor and Kegg data: \kg{} is faster and has better speedup over other methods. On Breastcancer and CreditRisk data, as noted previously, performance gap reduce owing to smaller data size, and runtime of all algorithms is similar to baseline \km{}. However, on Breastcancer, \kg{} is marginally better for $k=20, 30$. On CreditRisk, \exp{} is slightly faster for $k=20$, \kg{} is better on $k=30$, and \kb{} does better on $k=50$. Each algorithm employs different infrastructure for reducing DC, affecting the runtime differently. Particularly, on small data overhead due to the added infrastructure of faster \km{} variants reduces the performance gap.

\subsection{Experiments on Synthetic Data}\label{sec:exp_synthetic_data}
\begin{figure}[!htb]
\centering
\includegraphics[width=.48\textwidth]{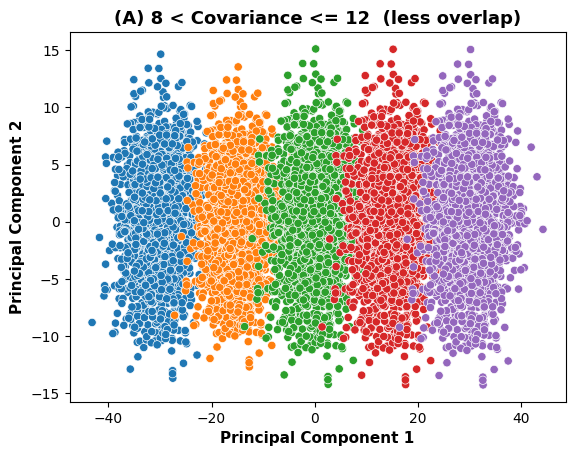}\hfill
\includegraphics[width=.48\textwidth]{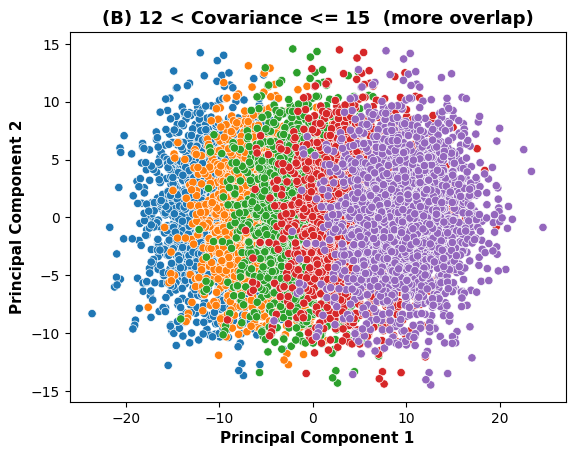}
\caption{PCA visualization of synthetic data. (A) data with moderate overlap between clusters. (B) data with higher overlap (higher variance) between clusters.}
\label{fig:example_syn_data} 
\end{figure}

Experiments are designed with two objectives: (1) to observe how the algorithms would scale with problem size, as $m \; \text{and} \; k$ are gradually increased; (2) to study the impact of varying a single property while holding others constant. The expectation is that under controlled conditions, most well-conducted algorithms should be able to scale well with the problem size. To test this, large synthetic data with balanced clusters (same number of samples in each cluster) are generated by sampling from multivariate Gaussian distribution. Here, we study the impact of size and clusters while performance on high dimensional data is discussed in next section. Additionally, we observe how well the algorithms perform when clustering data with varying degree of overlap between classes. Separation between clusters is controlled by modulating the variance (low variance leads to lower spread of data and vice versa). In clustering experiments, size ($m$) and dimensions ($d$) of data are fixed, $k$ is increased and separation between clusters is gradually decreased, thus transitioning from more separated to less separated clusters. In Scalability experiments, cluster ($k$) and dimensions ($d$) are held constant, while size ($m$) is increased. This controlled progression allows us to evaluate algorithm performance as clustering difficulty increases, whether by increasing $k$ or increasing dataset size $m$. Fig.\,\ref{fig:example_syn_data} show representative subset of the data used in the experiments. Appendix~\ref{sec:app_data_generation_process} provides more information about data generation. To complete experiments in reasonable time, we remove \km{} due to its very long runtime. We found that on several trials, \kd{} terminated early without completion, so we omit it from comparison.

\subsubsection{Results \& Analysis}\label{sec:results_synthetic_data}
\begin{figure}[!htb]
\centering
\includegraphics[scale=0.65]{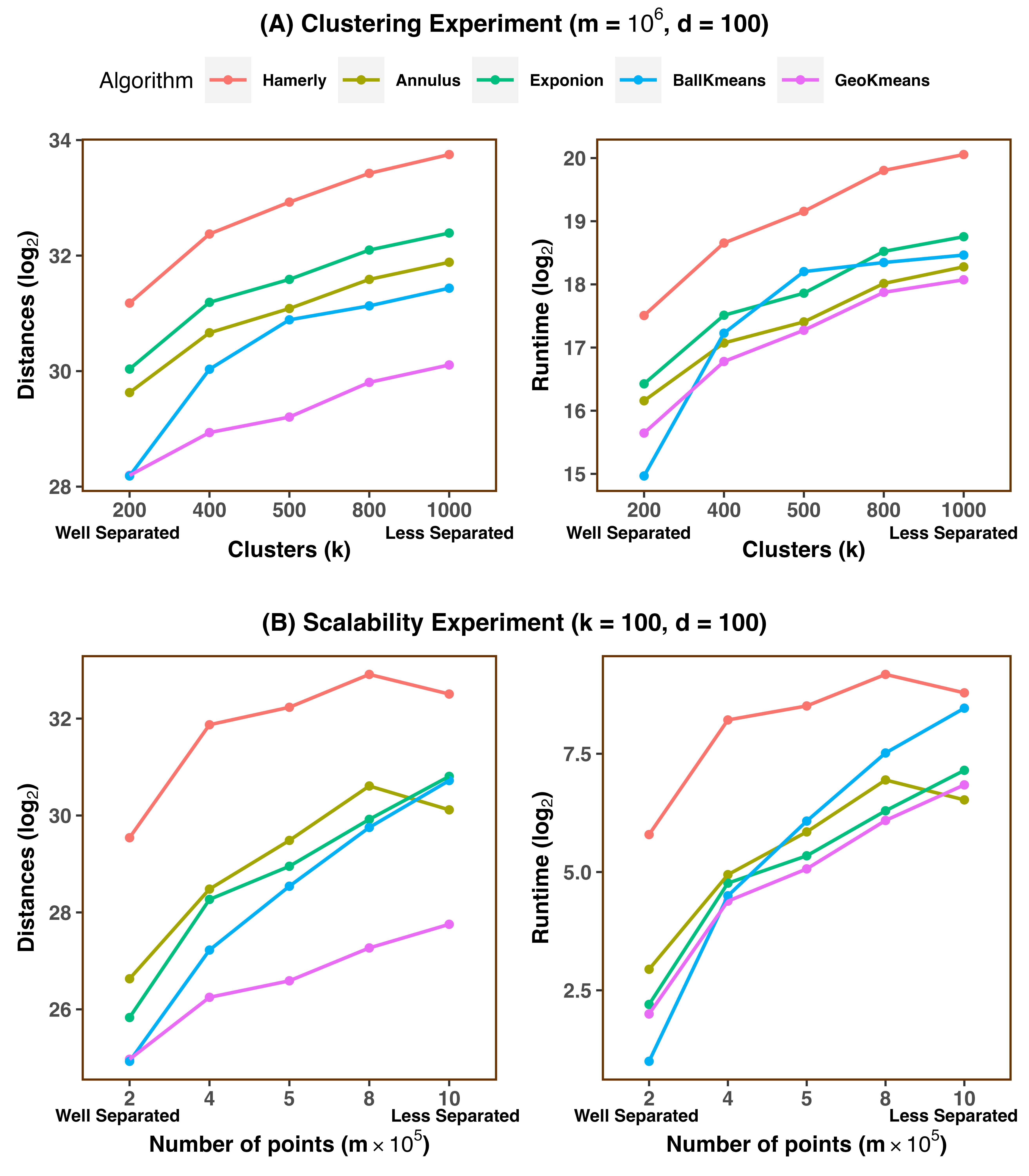}
\caption{\textbf{Evaluation on synthetic data:} On majority of datasets, cluster number and size-\kg{} has consistent speed-up. Performance of \kb{} is largely restricted to the more separated data. For both experiments, clusters at the left side of the plots (`More Separated') are generated with greater inter-cluster distance and lower covariance parameters, resulting in more distinguishable clusters. Moving rightward toward `Less Separated', we gradually decrease inter-cluster distances while increasing covariance parameters, creating datasets with higher cluster overlap.}
\label{fig:syn_data_results} 
\end{figure}

Fig.\,\ref{fig:syn_data_results} (A) illustrates the results of clustering experiment. On majority of datasets, \kg{} maintains a consistent speedup, compute less distances and take less time. In terms of distance computations, \kg{} performs significantly less computations than its counterparts, an order of magnitude less on $k= \{400, 500, 800\}$. For $k=200$, \kb{} is slightly better but its performance begin to suffer as $k$ and cluster overlap increase. However, as $k$ is further increased, number of neighbor clusters generally remains small compared to the total number of clusters, thus enabling \kb{} to eventually save distance computations \citep{xia2020fast}. This is observed as runtime of \kb{} first increase and then begin to drop at $k > 500$, helping it to catch-up. Amongst bounded algorithms, \ann{} shows relatively good performance, likely benefiting from the well-behaved distribution of synthetic data. Scalability experiments (Fig.\,\ref{fig:syn_data_results} (B)) show that, as $m$ increase-\ann{}, \exp{}, \kb{} and \kg{} has better scalability over \ham{}. This is expected as \ann{}, \exp{} are an improvement of \ham{}, while \kb{} and \kg{} benefit from unbounded design. However, \kg{} is noticeably better, as seen in the results. Similar to clustering experiments, performance of \kb{} is restricted to more separated data. In terms of runtime, since $k$ is fixed, \kb{} could no longer benefit from higher number of clusters. 

In scalability experiments, \exp{} did better than its bounded counterparts, though \ann{} does marginally better on data with high overlap ($m=10^6$). Despite minor differences, empirical results demonstrate that most faster \km{} variants enjoy good scalability on synthetic data. Interestingly, on most datasets, as $k$ and $m$ increase, \kg{} sustains better performance; aided by the knowledge of datum's orientation in the native space, \kg{} actively reduce DC which leads to lower runtime as well. As part of extreme testing, we further push this by clustering on 2-10 million data points. Results are given in Fig.\,\ref{fig:fig_scal_10_million} (appendix), and show that \kg{} is able to sustain speed-up on even larger data. The results demonstrate that \kg{} is well-behaved algorithm and it can efficiently process data across disparate cluster number ($k$), and data size ($m$).

\subsection{Experiments on High Dimensional scRNASeq Data}\label{sec:exp_highdim_data}
Introduction of single-cell RNA sequencing (scRNA-seq) has enabled the analysis of a cell’s transcriptome at an unprecedented granularity and processing speed. The experimental outcome of applying this technology is an $d \times M$ matrix containing aggregated mRNA expression counts of $d$ genes and $M$ cell samples. When it comes to analysing scRNA-seq datasets, \km{} clustering is at the heart of several downstream tasks, for example: profiling individual cells and quantization \citep{andrews2021tutorial}, discovering cell types \citep{kiselev2019challenges}, understanding developmental stages, and identifying new cell lineages \citep{stegle2015computational}. Here, we evaluate the efficacy of \kg{} on high dimensional scRNASeq data. Before clustering, standard processing was applied i.e. low expressed genes are removed followed by removal of cells with low overall gene counts. Normalization is done to adjust for differences in gene counts between cells followed by $log$ transformation and imputation \citep{van2018recovering}. Clustering is done by varying number of clusters while average DC and runtime is recorded.


\begin{figure}[t]
\centering
\includegraphics[scale=0.65]{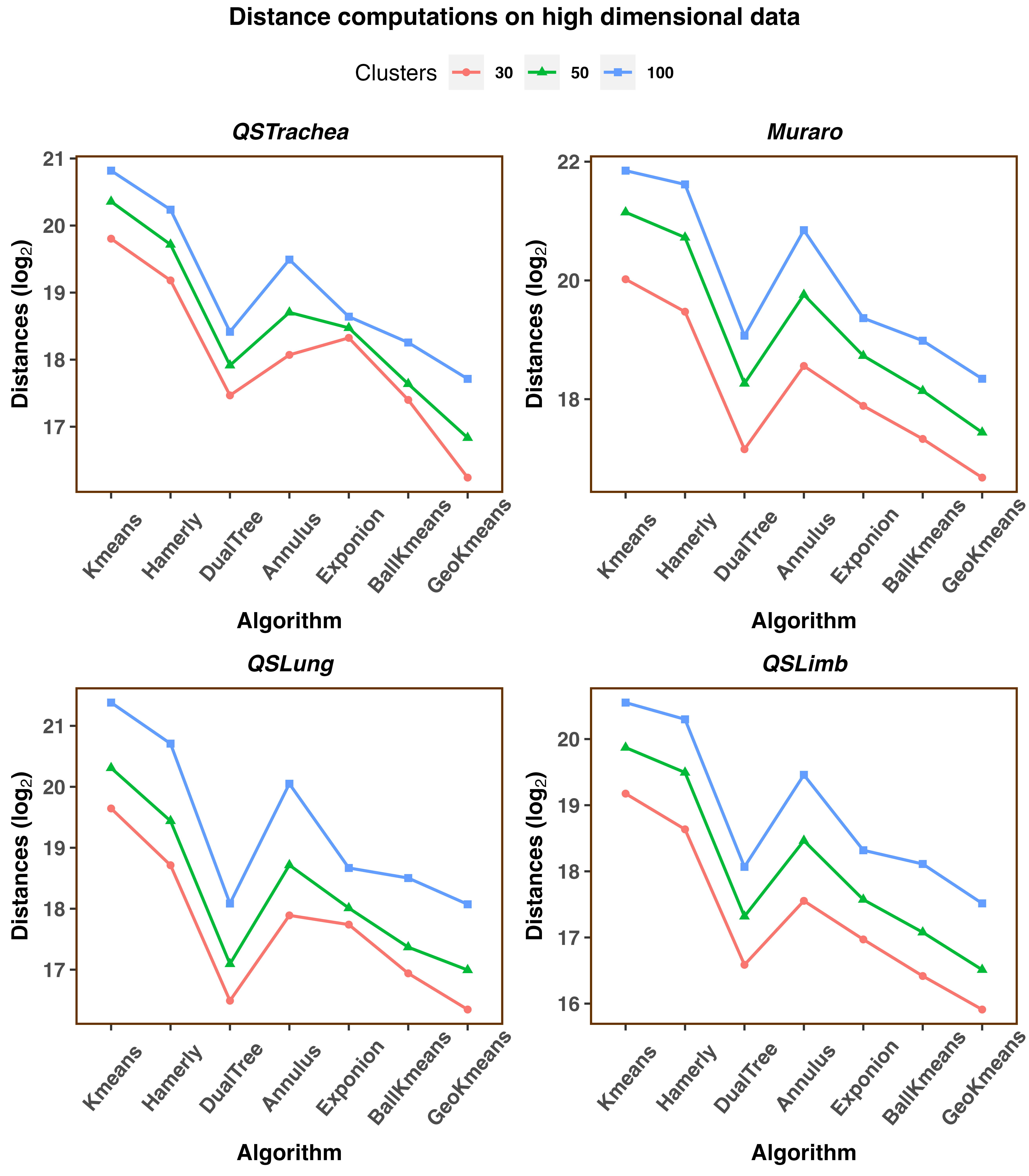}
\caption{\textbf{DC on high dimensional data:} \kg{} and \kb{} fare better in terms of DC. Across dataset and $k$, \kg{} consumes least DC. Reported DC is average of 10 trials.}
\label{fig:scRNA_distane_plots} 
\end{figure}

\begin{figure}[t]
\centering
\includegraphics[scale=0.65]{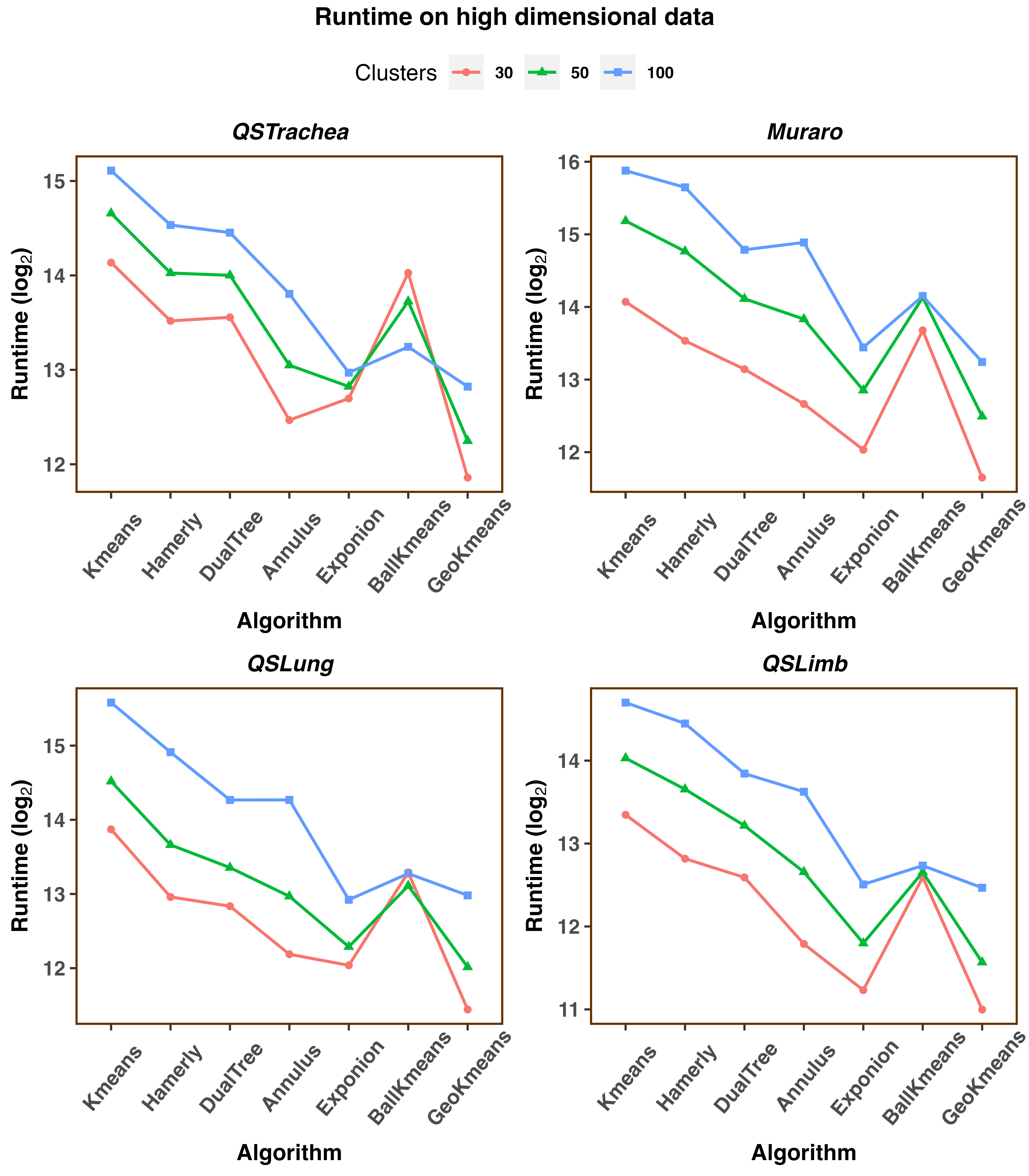}
\caption{\textbf{Runtime on high dimensional data:} As $k$ is increased, \kg{} exhibits good scalability. Reported runtime is average of 10 trials.}
\label{fig:scRNA_runtime_plots} 
\end{figure}


\subsubsection{Results}\label{sec:results_scRNA}
Fig.\,\ref{fig:scRNA_distane_plots} present the average distance computations and Fig.\,\ref{fig:scRNA_runtime_plots} demonstrate the average runtime for different algorithms. In terms of DC, both \kg{} and \kb{} do better than other algorithms, but \kg{} computes the fewest distances thus surpassing its counterparts. In higher dimensions, curse of dimensionality affects the scalability of all exact \km{} variants to different extent \citep{aggarwal2001surprising, hinneburg2000nearest}. While the performance of \ham{} degrades on high dimensional data \citep{hamerly1}, \kd{} becomes less effective as it spends more time on validating the bounds, due to bounds becoming loose in higher dimensions \citep{curtin2017dual}. Consequently, even though \kd{} does less DC than \ham{} and \ann{}-it does not translates into significant savings of runtime. In terms of runtime, both \kg{} and \exp{} are similar, though \kg{} is slightly faster. \kg{} is able to effectively use the knowledge about datums' orientation, leading to savings of DC and faster convergence. On $k = \langle30, 50\rangle$, \kb{} is slower than its bounded counterparts, but its performance improves marginally on $k=100$. The cause is found in the structures for DC reduction used by \kb{}: on small clusters, the computation of neighbors, stable and active areas is relatively costly. As $k$ increase, number of neighbors becomes small thus neighborhood computations in specific active area reduce increasingly larger DC.



\subsection{Sustainable Clustering: Profiling Energy Usgae}\label{sec:results_energy_usage}
\begin{figure}[!htbp]
\centering
\includegraphics[width=\linewidth]{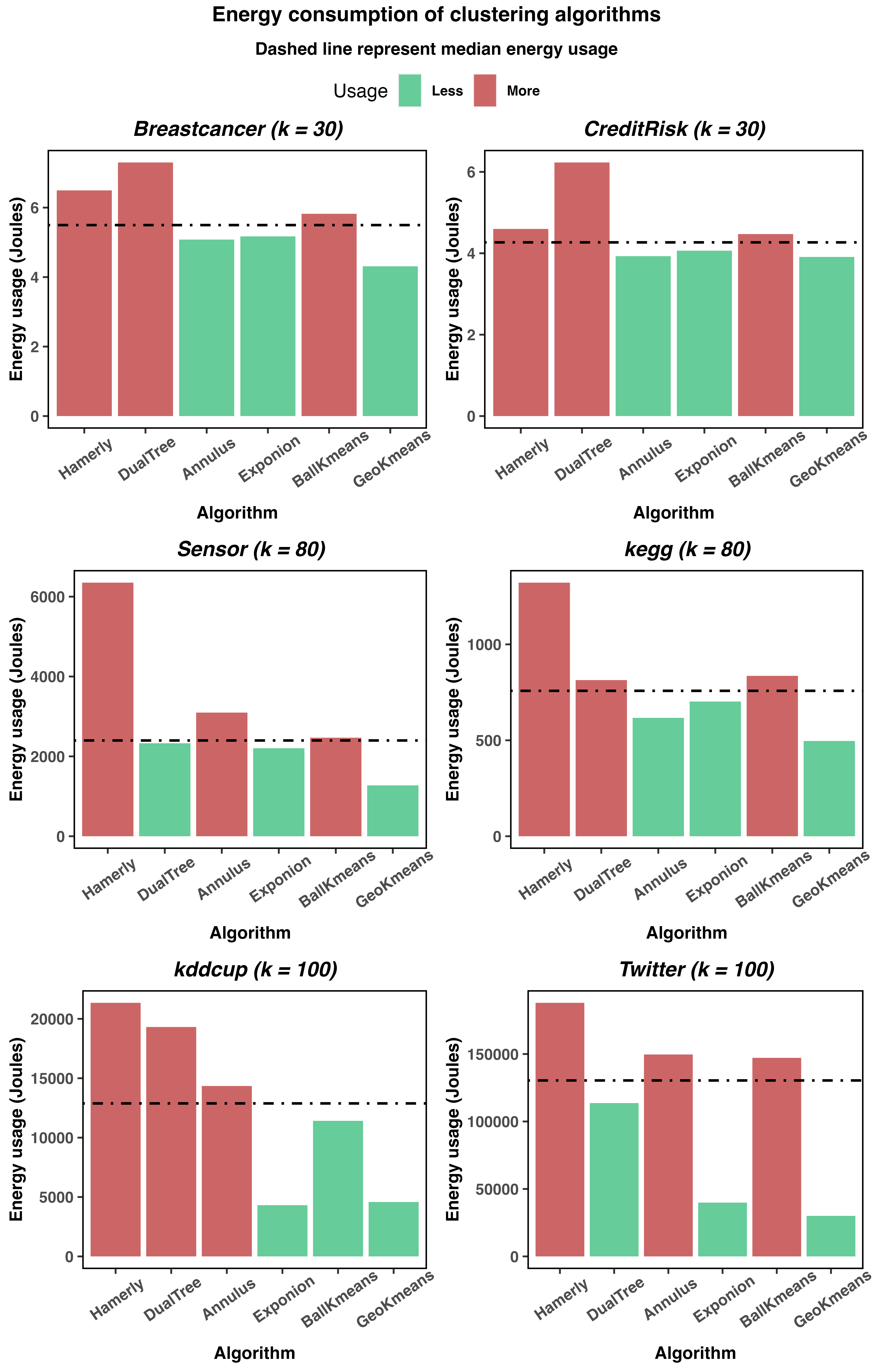}
\caption{\textbf{Energy cost of clustering algorithms:} Reported energy is the combined usage of CPU and RAM (memory), calculated as the average of 10 independent trials of the algorithms. \textbf{\textcolor{myred}{Red}} means usage is higher than median energy consumption, \textbf{\textcolor{mygreen}{Green}} indicate usage is less. Energy cost of \kg{} is consistently lower than median usage, and it consumes less energy than other algorithms.}
\label{fig:energyUsage} 
\end{figure}

The importance of considering the energy consumption of Machine Learning (ML) algorithms, is becoming increasingly critical for several reasons, for example: environmental impact \citep{strubell2019energy}, energy cost associated with inference on large complex models \citep{schwartz2020green}, impact of distributed computing workloads \citep{wu2023performance}. As the deployment and utilization of ML algorithms expand rapidly, the carbon footprint associated with these computational processes has gained attention \citep{temim2023analysis}. Therefore, in addition to optimizing computations and time, assessing the energy consumption of ML algorithms is gaining importance. Despite the widespread adoption of \km{} as a fundamental machine learning technique, analysis of its energy requirements has not yet been explored. As a novel exploration, we compare the energy consumption of existing fast \km{} algorithms, and outline the benefits offered by \kg{} as a more sustainable alternative. Statistics are collected via Intel Running Average Power Limit (RAPL) library \citep{khan2018rapl}. We refer to energy consumed by an algorithm as its \textit{energy cost} (EC), calculated as the sum of energy used by CPU and RAM (memory) during algorithm's execution. Fig.\,\ref{fig:energyUsage} summarize total EC and allocation of EC by computation (CPU) and memory is given in Table\ \ref{tab:table_energy_cost_cpu} and \ref{tab:table_energy_cost_memory}.

On small datasets-most algorithms exhibit similar energy cost, but \kg{} consumes the lowest amount of energy. On moderate sized sensor and kegg data, there is noticeable difference in the energy consumption, and \kg{} stands out as the most frugal. When comparing on kddcup data, \kg{} and \exp{} perform similar, but \exp{} does slightly less DC (as reported in Sec. \ref{sec:exp_real_data_distances}), we think that it helps \exp{} to consume relatively less energy. On Twitter data, there is a sharp increase in EC (driven by the larger data size and DC), but EC of \kg{} is significantly better than other algorithms. 

\input{Table6}
\input{Table7}

Overall, \ham{}, \kd{}, \ann{} and \kb{} tend to have higher energy demands, while \exp{} and \kg{} exhibit comparatively lower energy consumption, though \kg{} is noticeably economical, across most datasets. Interestingly, as discussed in Sec. \ref{sec:exp_real_data_distances}, and \ref{sec:exp_real_data_runtime}, in term of DC and runtime, \kb{} does better than most bounded variants and compares well with \exp{}. However in term of EC, on most datasets, \kb{} is expensive as its energy intake is higher than median usage, its only cheaper than \ham{}, while performing similar to \ann{}. Though, \kb{} reduces more DC than bounded variants, it also has extra burden, for example: determining whether to perform neighborhood calculation or not in each iteration and establishing annulus areas etc. This additional structure adds to the total energy cost for \kb{}. The empirical analysis underscores that independent of dataset and clusters, \kg{} incurs lower energy cost across CPU and memory, and demonstrates well-behaved energy demands. Being and unbounded algorithm, compared to \kb{}, \kg{} has significantly less EC and emerges as promising alternative for energy efficient clustering.

\subsubsection{Statistical analysis: relationship between runtime, DC and energy}\label{sec:results_pearson_correlation}
We statistically assess the impact of runtime and DC on energy cost of different algorithms. To empirically quantify their relationship, we compute the Pearson correlation coefficient ($r$). Pearson correlation is widely used to measure the strength and direction of relationship between continuous variables \citep{cohen2009pearson}. its value ranges from $-1$ to $1$, and can be interpreted as, $r = -1:$ negative correlation, $r = 0:$ no correlation, $r = 1:$ high correlation; magnitude ($|r|$) denote strength of correlation. Generally, $0.7 < r < 1$ is considered strong correlation \citep{rosenthal1986meta}. Intuitively speaking, one would also expect EC to rise if there is an increase in the runtime and distance computations. To validate this intuition, we aggregate the energy consumption, DC and runtime of each algorithm across datasets, and calculate the Pearson correlation between them. The results are given in Table\,\ref{tab:table_pearson_corr}. Indeed, higher EC is accompanied by higher runtime and DC, as evident from strong correlation between them. The pattern is consistent across algorithms, and illustrate the affect of DC and runtime on energy cost, besides outlining the impact of clustering algorithms beyond canonical savings of runtime.

\input{Table8}

%% file: Table2.tex
\begin{table}[h]

    \caption{Data used in experiments with size, number of features, and cluster number. (A) gives real-world data varying cluster number. (B) gives scRNA-Seq data.
    (C) gives synthetic data varying cluster and size.}
    
    \label{tab:data_details}

        \begin{tabular}{rrrrr} 
        \multicolumn{5}{c}{\textbf{(A) Real-world Data}} \\
        \toprule 
        
        Data & $m$ & $d$ & $k$ & source \\
        \cmidrule(lr){1-1} \cmidrule(lr){2-5} 
        BreastCancer &   569  & 30 & $\mathsf{K}_1$  &  \multirow{6}{*}{\href{https://archive.ics.uci.edu}{\makecell[c]{UCI machine\\learning\\repository}}}    \\ 
        CreditRisk   &   1000 & 7 &  $\mathsf{K}_1$  &      \\
        Sensor       &    58509   & 48 & $\mathsf{K}_2$  &   \\
        Kegg        &     65554   & 28 & $\mathsf{K}_2$  &    \\
        Kddcup      &   494020  & 38 & $\mathsf{K}_3$  &        \\
        Twitter     &   583250  & 78 & $\mathsf{K}_3$  &        \\
        \midrule \\[0.05cm]
        
        \multicolumn{5}{c}{\makecell[c]{$\mathsf{K}_1 = \{20, 30, 50\}$\\$\mathsf{K}_2 = \{30, 50, 80\}$\\$\mathsf{K}_3 = \{100, 300, 500\}$}}
        \\[0.7cm]


        \multicolumn{5}{c}{\textbf{(B) High Dimensional scRNA-seq Data}} \\
        \midrule
        Data & $m$ & $d$ & $k$ & source\\
        \midrule

        Trachea & 1350 & 18347 & $\mathsf{K}$  & \multirow{4}{*}{\citep{schaum2018single}}  \\
        
        Lung  & 1676  & 17243 & $\mathsf{K}$  & \\
        
        Limb & 1090  & 16624 & $\mathsf{K}$ & \\
        
        Muraro & 2122 & 15323 & $\mathsf{K}$  & \\
        
        \midrule \\[0.05cm]

        \multicolumn{5}{c}{\makecell[c]{$\mathsf{K} = \{30, 50, 100\}$}}\\[0.4cm]

        
        \multicolumn{5}{c}{\textbf{(C) Synthetic Data}} \\
        \midrule

        Data & $m$ & $d$ & \multicolumn{2}{r}{$k$} \\
        \cmidrule(lr){1-1} \cmidrule(lr){2-5} 
        
        Clustering & $10^6$ & $100$ & \multicolumn{2}{r}{$\mathsf{K}$} \\
        
        Scalability & $\mathsf{M}$ & $50$ & \multicolumn{2}{r}{$100$} \\

        \midrule \\[0.05cm]

        \multicolumn{5}{c}{\makecell[c]{$\mathsf{K} = \{2, 4, 5, 8, 10\} \times 10^2$\\$\mathsf{M} = \{2, 4, 5, 8, 10\} \times 10^5$}}
        \\
\end{tabular}
\end{table}

%% file: Table3.tex
\begin{sidewaystable}
{\footnotesize\setlength{\tabcolsep}{1pt}


    \centering
    \caption{Comparison of average distance computations of different algorithms}
    
    \label{tab:table_random_distances}
    
    \begin{tabular}{lcccccccccccccc}     
        
        \toprule
        Dataset & Clusters & \km{} & \multicolumn{2}{c}{\ham{}} & \multicolumn{2}{c}{\ann{}} &
        \multicolumn{2}{c}{\kd{}} & \multicolumn{2}{c}{\exp{}} & \multicolumn{2}{c}{\kb{}} &
        \multicolumn{2}{c}{\kg{}}\\
        \midrule
        & $k$ & $DC$ \ \ \ \ & $DC$ & $DC(S)$ & $DC$ & $DC(S)$ & $DC$ & $DC(S)$ & $DC$ & $DC(S)$ & $DC$ & $DC(S)$ & $DC$ & $DC(S)$\\
        \cmidrule(lr){4-5} \cmidrule(lr){6-7} \cmidrule(lr){8-9} \cmidrule(lr){10-11} \cmidrule(lr){12-13} \cmidrule(lr){14-15}
        \\

            \multirow{3}{*}{BreastCancer} & $20$ & $2.7\mathrm{e}5$ \ \ \ \ & 
            $1.2\mathrm{e}5 \pm 3.5\mathrm{e}4$ & $54.12$ & 
            $6.4\mathrm{e}4 \pm 1.6\mathrm{e}4$ & $76.69$ & 
            $4.0\mathrm{e}4 \pm 1.2\mathrm{e}4$ & $85.51$ & 
            $4.7\mathrm{e}4 \pm 1.1\mathrm{e}4$ & $82.87$ & 
            $3.9\mathrm{e}4 \pm 8.8\mathrm{e}3$ & $85.71$ & $\mathbf{2.9\mathrm{e}4 \pm 5.4\mathrm{e}3}$ & $\mathbf{89.36}$ \\ 
            
            & $30$ & $3.02\mathrm{e}5$ \ \ \ \ & $1.5\mathrm{e}5 \pm 4.0\mathrm{e}4$ & $47.21$ & 
            $8.8\mathrm{e}4 \pm 1.7\mathrm{e}4$ & $70.60$ & 
            $4.3\mathrm{e}4 \pm 1.2\mathrm{e}4$ & $85.55$ & 
            $5.4\mathrm{e}4 \pm 1.0\mathrm{e}4$ & $81.98$ & 
            $4.6\mathrm{e}4 \pm 8.7\mathrm{e}3$ & $84.60$ & $\mathbf{3.4\mathrm{e}4 \pm 5.4\mathrm{e}3}$ & $\mathbf{88.70}$ \\  
            
            & $50$ & $4.55\mathrm{e}5$ \ \ \ \ & $2.7\mathrm{e}5 \pm 1.0\mathrm{e}5$ & $40.01$ & 
            $1.6\mathrm{e}5 \pm 5.9\mathrm{e}4$ & $64.49$ & 
            $6.5\mathrm{e}4 \pm 3.2\mathrm{e}4$ & $85.70$ & 
            $8.3\mathrm{e}4 \pm 2.4\mathrm{e}4$ & $81.67$ & 
            $7.7\mathrm{e}4 \pm 2.1\mathrm{e}4$ & $82.90$ & $\mathbf{5.5\mathrm{e}4 \pm 1.2\mathrm{e}4}$ & $\mathbf{87.78}$ \\
            \\
            \midrule
            \\

            \multirow{3}{*}{CreditRisk} & $20$ & $9.1\mathrm{e}5$ \ \ \ \ & 
            $1.3\mathrm{e}5 \pm 4.9\mathrm{e}4$ & $85.56$ & 
            $8.4\mathrm{e}4 \pm 2.2\mathrm{e}4$ & $90.77$ & $\mathbf{4.7\mathrm{e}4 \pm 1.9\mathrm{e}4}$ & $\mathbf{94.83}$ & $5.5\mathrm{e}4 \pm 1.6\mathrm{e}4$ & $93.91$ & 
            $7.0\mathrm{e}4 \pm 2.38\mathrm{e}4$ & $92.30$ & 
            $7.4\mathrm{e}4 \pm 2.37\mathrm{e}4$ & $91.86$ \\  
            
            & $30$ & $9.57\mathrm{e}5$ \ \ \ \ & 
            $1.6\mathrm{e}5 \pm 2.1\mathrm{e}4$ & $82.88$ & 
            $1.1\mathrm{e}5 \pm 1.4\mathrm{e}4$ & $87.67$ & $\mathbf{4.8\mathrm{e}4 \pm 1.0\mathrm{e}4}$ & $\mathbf{94.97}$ & $6.4\mathrm{e}4 \pm 6.4\mathrm{e}3$ & $93.21$ & 
            $7.8\mathrm{e}4 \pm 9.8\mathrm{e}3$ & $91.84$ & 
            $7.5\mathrm{e}4 \pm 1.1\mathrm{e}4$ & $92.15$ \\  
            
            & $50$ & $1.20\mathrm{e}6$ \ \ \ \ & 
            $3.0\mathrm{e}5 \pm 7.0\mathrm{e}4$ & $74.36$ & 
            $2.2\mathrm{e}5 \pm 5.1\mathrm{e}4$ & $81.43$ & $\mathbf{6.0\mathrm{e}4 \pm 1.7\mathrm{e}4}$ & $\mathbf{94.99}$ & $1.0\mathrm{e}5 \pm 1.5\mathrm{e}4$ & $91.12$ & 
            $1.2\mathrm{e}5 \pm 1.8\mathrm{e}4$ & $89.83$ & 
            $1.0\mathrm{e}5 \pm 1.4\mathrm{e}4$ & $91.53$ \\
            \\
            \midrule
            \\

            \multirow{3}{*}{Sensor} & $50$ & 
            $5.89\mathrm{e}8$ \ \ \ \ & 
            $5.2\mathrm{e}8 \pm 1.5\mathrm{e}8$ & $74.36$ & 
            $2.2\mathrm{e}8 \pm 3.3\mathrm{e}7$ & $61.73$ & 
            $4.7\mathrm{e}7 \pm 7.5\mathrm{e}6$ & $91.89$ & 
            $1.7\mathrm{e}8 \pm 4.3\mathrm{e}7$ & $69.87$ & 
            $8.1\mathrm{e}7 \pm 1.5\mathrm{e}7$ & $86.15$ & $\mathbf{1.5\mathrm{e}7 \pm 3.6\mathrm{e}6}$ & $\mathbf{97.38}$ \\ 
            
            & $80$ & $1.02\mathrm{e}9$ \ \ \ \ & $8.8\mathrm{e}8 \pm 9.0\mathrm{e}7$ & $11.57$ & 
            $3.9\mathrm{e}8 \pm 3.3\mathrm{e}7$ & $61.73$ & 
            $7.9\mathrm{e}7 \pm 8.5\mathrm{e}6$ & $92.21$ & 
            $2.6\mathrm{e}8 \pm 2.7\mathrm{e}7$ & $73.94$ & 
            $1.1\mathrm{e}8 \pm 1.8\mathrm{e}7$ & $88.82$ & $\mathbf{1.7\mathrm{e}7 \pm 1.3\mathrm{e}6}$ & $\mathbf{98.24}$ \\ 
            
            & $100$ & $1.73\mathrm{e}9$ \ \ \ \ & $1.6\mathrm{e}9 \pm 4.4\mathrm{e}8$ & $13.55$ & 
            $6.6\mathrm{e}8 \pm 1.5\mathrm{e}68$ & $61.87$ & 
            $1.1\mathrm{e}8 \pm 1.9\mathrm{e}7$ & $93.34$ & 
            $4.4\mathrm{e}8 \pm 1.1\mathrm{e}8$ & $74.24$ & 
            $1.5\mathrm{e}8 \pm 3.5\mathrm{e}7$ & $91.02$ & $\mathbf{2.5\mathrm{e}7 \pm 5.2\mathrm{e}6}$ & $\mathbf{98.50}$ \\ 
            \\
            \midrule
            \\

            \multirow{3}{*}{Kegg} & $50$ & 
            $2.73\mathrm{e}8$ \ \ \ \ & 
            $1.7\mathrm{e}8 \pm 5.7\mathrm{e}7$ & $4.68$ & 
            $6.4\mathrm{e}7 \pm 1.6\mathrm{e}67$ & $76.34$ & 
            $1.7\mathrm{e}7 \pm 3.3\mathrm{e}6$ & $93.63$ & 
            $9.2\mathrm{e}7 \pm 2.7\mathrm{e}7$ & $66.33$ & 
            $6.4\mathrm{e}7 \pm 2.2\mathrm{e}7$ & $76.34$ & $\mathbf{8.8\mathrm{e}6 \pm 1.8\mathrm{e}6}$ & $\mathbf{96.74}$ \\ 
            
            & $80$ & $5.73\mathrm{e}8$ \ \ \ \ & $3.4\mathrm{e}8 \pm 9.8\mathrm{e}7$ & $34.86$ & 
            $1.4\mathrm{e}8 \pm 3.2\mathrm{e}7$ & $74.90$ & 
            $2.4\mathrm{e}7 \pm 6.0\mathrm{e}6$ & $95.68$ & 
            $1.5\mathrm{e}8 \pm 4.5\mathrm{e}7$ & $72.12$ & 
            $7.0\mathrm{e}7 \pm 1.6\mathrm{e}7$ & $87.75$ & $\mathbf{1.2\mathrm{e}7 \pm 2.1\mathrm{e}6}$ & $\mathbf{97.77}$ \\  
            
            & $100$ & $8.00\mathrm{e}8$ \ \ \ \ & $4.7\mathrm{e}8 \pm 1.7\mathrm{e}8$ & $39.21$ & 
            $2.1\mathrm{e}8 \pm 4.9\mathrm{e}7$ & $73.58$ & 
            $3.1\mathrm{e}7 \pm 7.5\mathrm{e}6$ & $96.05$ & 
            $2.0\mathrm{e}8 \pm 7.1\mathrm{e}7$ & $74.40$ & 
            $7.8\mathrm{e}7 \pm 2.6\mathrm{e}7$ & $90.23$ & $\mathbf{1.5\mathrm{e}7 \pm 3.0\mathrm{e}6}$ & $\mathbf{98.11}$ \\ 
            \\
            \midrule
            \\

            \multirow{3}{*}{Kddcup} & $100$ & $2.4\mathrm{e}10$ \ \ \ \ & 
            $3.8\mathrm{e}9 \pm 3.8\mathrm{e}8$ & $40.82$ & 
            $2.3\mathrm{e}9 \pm 1.4\mathrm{e}8$ & $90.37$ & 
            $1.9\mathrm{e}9 \pm 3.6\mathrm{e}8$ & $91.92$ & 
            $4.5\mathrm{e}8 \pm 4.3\mathrm{e}7$ & $98.14$ & 
            $6.4\mathrm{e}8 \pm 6.2\mathrm{e}7$ & $97.40$ & $\mathbf{3.0\mathrm{e}8 \pm 6.9\mathrm{e}6}$ & $\mathbf{98.75}$ \\ 
            
            & $300$ & $7.42\mathrm{e}10$ \ \ \ \ & $1.1\mathrm{e}10 \pm 3.3\mathrm{e}8$ & $84.57$ & 
            $4.3\mathrm{e}9 \pm 2.0\mathrm{e}8$ & $94.18$ & 
            $1.1\mathrm{e}10 \pm 4.9\mathrm{e}8$ & $84.10$ & 
            $8.2\mathrm{e}8 \pm 2.1\mathrm{e}7$ & $98.88$ & 
            $9.1\mathrm{e}8 \pm 7.0\mathrm{e}7$ & $98.77$ & $\mathbf{4.2\mathrm{e}8 \pm 1.9\mathrm{e}6}$ & $\mathbf{99.42}$ \\  
            
            & $500$ & $1.2\mathrm{e}11$ \ \ \ \ & $2.1\mathrm{e}10 \pm 4.5\mathrm{e}8$ & $82.65$ & 
            $6.6\mathrm{e}9 \pm 4.3\mathrm{e}8$ & $94.61$ & 
            $2.0\mathrm{e}10 \pm 8.0\mathrm{e}8$ & $83.75$ & 
            $1.1\mathrm{e}9 \pm 3.2\mathrm{e}7$ & $99.04$ & 
            $1.0\mathrm{e}9 \pm 1.0\mathrm{e}8$ & $99.17$ & $\mathbf{5.6\mathrm{e}8 \pm 2.2\mathrm{e}6}$ & $\mathbf{99.54}$ \\ 
            \\
            \midrule
            \\

            \multirow{3}{*}{Twitter} & $100$ & $2.9\mathrm{e}10$ \ \ \ \ & 
            $1.9\mathrm{e}10 \pm 2.7\mathrm{e}8$ & $33.36$ & 
            $1.5\mathrm{e}10 \pm 6.4\mathrm{e}8$ & $47.09$ & 
            $3.6\mathrm{e}9 \pm 1.0\mathrm{e}8$ & $87.56$ & 
            $3.7\mathrm{e}9 \pm 7.4\mathrm{e}7$ & $87.20$ & 
            $2.2\mathrm{e}9 \pm 4.0\mathrm{e}7$ & $92.19$ & $\mathbf{3.58\mathrm{e}8 \pm 3.2\mathrm{e}5}$ & $\mathbf{98.77}$ \\ 
            
            & $300$ & $8.76\mathrm{e}10$ \ \ \ \ & $7.5\mathrm{e}10 \pm 3.7\mathrm{e}8$ & $14.16$ & 
            $6.1\mathrm{e}10 \pm 5.9\mathrm{e}8$ & $29.66$ & 
            $1.2\mathrm{e}10 \pm 2.1\mathrm{e}8$ & $85.61$ & 
            $1.1\mathrm{e}10 \pm 1.6\mathrm{e}8$ & $87.17$ & 
            $5.6\mathrm{e}9 \pm 7.0\mathrm{e}7$ & $93.51$ & $\mathbf{4.9\mathrm{e}8 \pm 1.8\mathrm{e}5}$ & $\mathbf{99.43}$ \\ 
            
            & $500$ & $1.46\mathrm{e}11$ \ \ \ \ & $1.3\mathrm{e}11 \pm 5.9\mathrm{e}8$ & $6.83$ & 
            $1.1\mathrm{e}11 \pm 5.7\mathrm{e}8$ & $24.46$ & 
            $2.2\mathrm{e}10 \pm 3.2\mathrm{e}8$ & $84.61$ & 
            $1.8\mathrm{e}10 \pm 1.0\mathrm{e}8$ & $87.25$ & 
            $8.6\mathrm{e}9 \pm 3.2\mathrm{e}7$ & $94.05$ & $\mathbf{6.5\mathrm{e}8 \pm 9.3\mathrm{e}4}$ & $\mathbf{99.55}$ \\
            \\
            
        \bottomrule
        \\
        \multicolumn{15}{l}{\makecell[l]{$DC$ is the average (10 trials) count of distance computations; $DC(S)$ represent percentage DC savings over baseline \km{}. Entries in \textbf{bold} indicate the best\\ average DC. The values are reported as mean $\pm$ one standard deviation.}}
        
    \end{tabular}}
\end{sidewaystable}

%% file: Table4.tex
\begin{sidewaystable}
{\footnotesize\setlength{\tabcolsep}{1pt}

    \centering
    
    \caption{Comparison of average runtime and speed-up of different algorithms}
    
    \label{tab:table_random_runtime}

    \begin{tabular}{lcccccccccccccc}     
        
        \toprule
        Dataset & Clusters & \km{} & \multicolumn{2}{c}{\ham{}} & \multicolumn{2}{c}{\ann{}} &
        \multicolumn{2}{c}{\kd{}} & \multicolumn{2}{c}{\exp{}} & \multicolumn{2}{c}{\kb{}} &
        \multicolumn{2}{c}{\kg{}}\\
        \midrule
         & $k$ & $RT$ \ \ \ \ & $RT$ & $RT(S)$ & $RT$ & $RT(S)$ & $RT$ & $RT(S)$ & $RT$ & $RT(S)$ & $RT$ & $RT(S)$ & $RT$ & $RT(S)$\\
         \cmidrule(lr){4-5} \cmidrule(lr){6-7} \cmidrule(lr){8-9} \cmidrule(lr){10-11} \cmidrule(lr){12-13} \cmidrule(lr){14-15}
         \\

            \multirow{3}{*}{BreastCancer} & $20$ & $7.0\mathrm{e}5 \pm 3.2$ \ \ \ \ & $3.0\mathrm{e}5 \pm 1.04$ & $57.7$ & $1.7\mathrm{e}5 \pm 0.43$ & $74.7$ & $5.8\mathrm{e}5  \pm 1.67$ & $16.3$ & $1.5\mathrm{e}5 \pm 0.36$ & $78.3$ & $1.9\mathrm{e}5 \pm 0.55$ & $72.4$ & $\mathbf{1.3\mathrm{e}5 \pm 0.32}$ & $\mathbf{80.6}$ \\ 
            
            & $30$ & $6.9\mathrm{e}5 \pm 2.1$ \ \ \ \ & $3.6\mathrm{e}5 \pm 0.85$ & $47.8$ & $2.2\mathrm{e}5 \pm 0.41$ & $67.6$ & $6.2\mathrm{e}5 \pm1.55$ & $8.90$ & $1.6\mathrm{e}5 \pm 0.34$ & $75.8$ & $1.7\mathrm{e}5 \pm 0.39$ & $75.3$ & $\mathbf{1.5\mathrm{e}5 \pm 0.31}$ & $\mathbf{77.8}$ \\

            & $50$ & $9.4\mathrm{e}5 \pm 4.03$ \ \ \ \ & $5.6\mathrm{e}5 \pm 2.23$ & $39.8$ & $3.6\mathrm{e}5 \pm 1.31$ & $61.3$ & $8.5\mathrm{e}5 \pm 3.88$ & $9.14$ & $2.8\mathrm{e}5 \pm 
            1.0$ & $70.0$ & $\mathbf{1.9\mathrm{e}5 \pm 0.64}$ & $\mathbf{79.5}$ & $2.3\mathrm{e}5 \pm 0.74$ & $74.7$ \\ 
            \\
            \midrule
            \\

            \multirow{3}{*}{CreditRisk} & $20$ & $13.7\mathrm{e}5 \pm 6.01$ \ \ \ \ & $2.3\mathrm{e}5 \pm 0.92$ & $82.8$ & $1.8\mathrm{e}5 \pm 0.59$ & $86.6$ & $5.8\mathrm{e}5 \pm 2.28$ & $57.1$ & $\mathbf{1.6\mathrm{e}5 \pm 0.58}$ & $\mathbf{88.3}$ & $2.3\mathrm{e}5 \pm1.07$ & $82.6$ & $1.7\mathrm{e}5 \pm 0.61$ & $87.5$ \\ 
            
            & $30$ & $14.1\mathrm{e}5 \pm 5.25$ \ \ \ \ & $2.6\mathrm{e}5 \pm 0.34$ & $81.4$ & $2.1\mathrm{e}5 \pm 0.27$ & $84.7$ & $6.2\mathrm{e}5 \pm 1.25$ & $55.6$ & $1.7\mathrm{e}5 \pm 0.23$ & $87.7$ & $1.8\mathrm{e}5 \pm 0.29$ & $86.9$ & $\mathbf{1.6\mathrm{e}5 \pm 0.28}$ & $\mathbf{88.0}$ \\ 
            
            & $50$ & $16.2\mathrm{e}5 \pm 4.53$ \ \ \ \ & $4.4\mathrm{e}5 \pm 0.99$ & $72.5$ & $3.5\mathrm{e}5 \pm 0.84$ & $78.1$ & $8.0\mathrm{e}5 \pm 2.27$ & $50.4$ & $3.1\mathrm{e}5 \pm 0.64$ & $80.9$ & $\mathbf{1.9\mathrm{e}5 \pm 0.33}$ & $\mathbf{87.9}$ & $2.2\mathrm{e}5 \pm0.40$ & $85.9$ \\
            \\
            \midrule
            \\

            \multirow{3}{*}{Sensor} & $50$ & $2.1\mathrm{e}5 \pm 6\mathrm{e}3$ \ \ \ \ & $1.8\mathrm{e}5 \pm 5.5\mathrm{e}3$ & $11.4$ & $8.5\mathrm{e}4 \pm 1.3\mathrm{e}3$ & $59.6$ & $6.5\mathrm{e}4 \pm 1.2\mathrm{e}3$ & $69.4$ & $7.1\mathrm{e}4 \pm 1.8\mathrm{e}3$ & $66.4$ & $9.7\mathrm{e}4 \pm 2.1\mathrm{e}3$ & $54.3$ & $\mathbf{3.5\mathrm{e}4 \pm 9.4\mathrm{e}2}$ & $\mathbf{83.2}$ \\  
            
            & $80$ & $3.6\mathrm{e} \pm 4.7\mathrm{e}3$ \ \ \ \ & $3.1\mathrm{e}4 \pm 3.1\mathrm{e}3$ & $13.3$ & $1.4\mathrm{e}4 \pm 1.2\mathrm{e}3$ & $59.1$ & $1.0\mathrm{e}4 \pm 1.5\mathrm{e}3$ & $70.8$ & $1.0\mathrm{e}4 \pm 1.0\mathrm{e}3$ & $71.1$ & $1.3\mathrm{e}4 \pm 2.1\mathrm{e}3$ & $64.1$ & $\mathbf{5.0\mathrm{e}3 \pm 5.2\mathrm{e}2}$ & $\mathbf{86.2}$ \\

            & $100$ & $6.2\mathrm{e}4 \pm 1.8\mathrm{e}4$ \ \ \ \ & $5.9\mathrm{e}4 \pm 1.5\mathrm{e}4$ & $4.1$ & $2.4\mathrm{e}4 \pm 5.7\mathrm{e}3$ & $60.0$ & $1.5\mathrm{e}4 \pm 2.7\mathrm{e}3$ & $74.6$ & $1.7\mathrm{e}4 \pm 4.2\mathrm{3}3$ & $71.5$ & $1.8\mathrm{e}4 \pm 5.0\mathrm{e}3$ & $70.3$ & $\mathbf{8.4\mathrm{e}3 \pm 2.0\mathrm{e}3}$ & $\mathbf{86.4}$ \\ 
            \\
            \midrule
            \\

            \multirow{3}{*}{Kegg} & $50$ & 
            $5.1\mathrm{e}3 \pm 1.6\mathrm{e}3$ \ \ \ \ & 
            $3.4\mathrm{e}3 \pm 1.0\mathrm{e}3$ & $33.0$ & 
            $1.3\mathrm{e}3 \pm 3.3\mathrm{e}2$ & $73.9$ & 
            $2.61\mathrm{e}3 \pm 5.1\mathrm{e}2$ & $49.6$ & 
            $2.0\mathrm{e}3 \pm 6.0\mathrm{e}2$ & $61.3$ & 
            $3.9\mathrm{e}3 \pm 1.2\mathrm{e}3$ & $23.4$ & $\mathbf{1.0\mathrm{e}3 \pm 3.1\mathrm{e}2}$ & 
            $\mathbf{79.1}$ \\ 
            
            & $80$ & 
            $1.0\mathrm{e}4 \pm 3.0\mathrm{e}3$ \ \ \ \ & 
            $6.6\mathrm{e}3 \pm 1.8\mathrm{e}3$ & $39.4$ & 
            $2.8\mathrm{e}3 \pm 6.3\mathrm{e}2$ & $73.8$ & 
            $3.6\mathrm{e}3 \pm 9.3\mathrm{e}2$ & $67.1$ & 
            $3.3\mathrm{e}3 \pm 9.4\mathrm{e}2$ & $69.2$ & 
            $4.0\mathrm{e}3 \pm 9.2\mathrm{e}2$ & $62.6$ & $\mathbf{1.7\mathrm{e}3 \pm 4.9\mathrm{e}2}$ & $\mathbf{83.6}$ \\ 
            
            & $100$ & 
            $1.5\mathrm{e}4 \pm 5.2\mathrm{e}3$ \ \ \ \ & 
            $9.0\mathrm{e}3 \pm 3.2\mathrm{e}3$ & $40.8$ & 
            $4.1\mathrm{e}3 \pm 9.6\mathrm{e}2$ & $72.6$ & 
            $4.5\mathrm{e}3 \pm 1.1\mathrm{e}3$ & $69.8$ & 
            $4.2\mathrm{e}3 \pm 1.4\mathrm{e}3$ & $72.0$ & 
            $4.5\mathrm{e}3 \pm 1.3\mathrm{e}3$ & $70.2$ & $\mathbf{2.3\mathrm{e}3 \pm 7.7\mathrm{e}2}$ & $\mathbf{84.5}$ \\
            \\
            \midrule
            \\

            \multirow{3}{*}{Kddcup} & $100$ & $7.0\mathrm{e}5 \pm 2.8\mathrm{e}3$ \ \ \ \ & 
            $1.1\mathrm{e}5 \pm 1.1\mathrm{e}5$ & $83.2$ & 
            $7.4\mathrm{e}4 \pm 4.0\mathrm{e}3$ & $89.3$ & 
            $9.9\mathrm{e}4 \pm 1.4\mathrm{e}4$ & $85.9$ & $\mathbf{2.1\mathrm{e}4 \pm 1.1\mathrm{e}3}$ & $\textcolor{blue}{\mathbf{96.9}}$ & 
            $6.0\mathrm{e}4 \pm 5.7\mathrm{e}3$ & $91.4$ & 
            $2.2\mathrm{e}4 \pm 5.3\mathrm{e}2$ & $\textcolor{blue}{\mathbf{96.8}}$ \\
            
            &  $300$ & 
            $2.0\mathrm{e}6 \pm 2.5\mathrm{e}4$ \ \ \ \ & 
            $3.2\mathrm{e}5 \pm 1.1\mathrm{e}4$ & $82.5$ & 
            $1.2\mathrm{e}5 \pm 6.2\mathrm{e}3$ & $93.6$ & 
            $4.7\mathrm{e}5 \pm 1.7\mathrm{e}4$ & $76.9$ & 
            $3.2\mathrm{e}4 \pm 7.5\mathrm{e}2$ & $62.3$ & 
            $6.4\mathrm{e}4 \pm 5.5\mathrm{e}3$ & $96.8$ & $\mathbf{3.2\mathrm{e}4 \pm 7.0\mathrm{e}2}$ & 
            $\mathbf{98.4}$ \\ 
            
            & $500$ & 
            $3.4\mathrm{e}6 \pm 5.2\mathrm{e}4$ \ \ \ \ & $6.0\mathrm{e}5 \pm 1.3\mathrm{e}4$ & $32.5$ & 
            $2.0\mathrm{e}5 \pm 1.2\mathrm{e}4$ & $94.1$ & 
            $7.9\mathrm{e}5 \pm 2.9\mathrm{e}4$ & $76.8$ & $\mathbf{4.6\mathrm{e}4 \pm 1.0\mathrm{e}3}$ & $\textcolor{blue}{\mathbf{98.6}}$ & 
            $6.1\mathrm{e}4 \pm 8.5\mathrm{e}3$ & $98.2$ & 
            $4.8\mathrm{e}4 \pm 1.8\mathrm{e}3$ & $\textcolor{blue}{\mathbf{98.6}}$ \\
            \\
            \midrule
            \\

            \multirow{3}{*}{Twitter} & $100$ & $1.7\mathrm{e}6 \pm 6.1\mathrm{e}4$ \ \ \ \ & 
            $1.1\mathrm{e}6 \pm 4.6\mathrm{e}4$ & $32.5$ & 
            $9.4\mathrm{e}5 \pm 4.7\mathrm{e}4$ & $45.7$ & 
            $6.7\mathrm{e}5 \pm 2.6\mathrm{e}4$ & $61.0$ & 
            $2.4\mathrm{e}5 \pm 1.0\mathrm{e}4$ & $85.7$ & 
            $9.4\mathrm{e}5 \pm 1.1\mathrm{e}5$ & $45.3$ & $\mathbf{1.6\mathrm{e}5 \pm 1.1\mathrm{e}4}$ & 
            $\mathbf{90.2}$ \\ 
            
            & $300$ & $5.01\mathrm{e}6 \pm 6.7\mathrm{e}4$ \ \ \ \ & $4.3\mathrm{e}6 \pm 7.0\mathrm{e}4$ & $14.1$ & 
            $3.6\mathrm{e}6 \pm 6.0\mathrm{e}4$ & $27.3$ & 
            $2.1\mathrm{e}6 \pm 4.3\mathrm{e}4$ & $57.8$ & 
            $6.8\mathrm{e}5 \pm 1.3\mathrm{e}4$ & $86.3$ & 
            $1.4\mathrm{e}6 \pm 7.9\mathrm{e}4$ & $71.5$ & $\mathbf{5.5\mathrm{e}5 \pm 8.3\mathrm{e}4}$ & 
            $\mathbf{89.0}$ \\ 
            
            & $500$ & $8.30\mathrm{e}6 \pm 1.2\mathrm{e}5$ \ \ \ \ & $7.7\mathrm{e}6 \pm 1.1\mathrm{e}5$ & $6.7$ & 
            $6.5\mathrm{e}6 \pm 1.4\mathrm{e}5$ & $20.7$ & 
            $3.7\mathrm{e}6 \pm 5.5\mathrm{e}4$ & $55.1$ & $\mathbf{1.13\mathrm{e}6 \pm 2.0\mathrm{e}4}$ & $\textcolor{blue}{\mathbf{86.3}}$ & 
            $1.8\mathrm{e}6 \pm 6.2\mathrm{e}4$ & $78.0$ & 
            $1.15\mathrm{e}6 \pm 1.1\mathrm{e}5$ & $\textcolor{blue}{\mathbf{86.1}}$ \\
            \\
            
        \bottomrule
        \\
        \multicolumn{15}{l}{\makecell[l]{$RT$ is the average (10 trials) runtime in milliseconds. $RT(S)$ indicate percentage runtime speed-up over baseline \km{}. Entries in \textbf{bold} indicate the best average\\ runtime. Entries in \textcolor{blue}{Blue} indicate similar runtime speedup. The values are reported as mean $\pm$ one standard deviation.}}
        
    \end{tabular}}
\end{sidewaystable}


%% file: Table6.tex
\begin{table}[htb]
    \centering
    \caption{\textbf{Energy Cost (Joules) on CPU}}
    \label{tab:table_energy_cost_cpu}

    \begin{tabular}{lccccccc}
    \toprule
    \textsf{Data} & \textsf{Clusters ($k$)} & \textsf{\ham{}} & \textsf{\ann{}} & \textsf{\exp{}} & \textsf{\kd{}} & \textsf{Ball-$k$m} & \textsf{Geo-$k$m}\\
    \midrule

    Breastcancer & $30$ & \cellcolor{myred}$4.57$ & \cellcolor{mygreen}$3.58$ & \cellcolor{mygreen}$3.64$ & \cellcolor{myred}$5.15$ & \cellcolor{myred}$4.1$ & \cellcolor{mygreen}$\mathbf{3.04}$ \\ 

     CreditRisk & $30$ & \cellcolor{myred}$3.23$ & \cellcolor{mygreen}$2.77$ & \cellcolor{mygreen}$2.85$ & \cellcolor{myred}$4.38$ & \cellcolor{myred}$3.15$ & \cellcolor{mygreen}$\mathbf{2.75}$ \\ 
    
     Sensor & $80$ & \cellcolor{myred}$4697.5$ & \cellcolor{myred}$2291.76$ & \cellcolor{mygreen}$1634.89$ & \cellcolor{mygreen}$1738.28$ & \cellcolor{myred}$1842.83$ & \cellcolor{mygreen}$\mathbf{944.35}$ \\ 
    
     kegg & $80$ & \cellcolor{myred}$979.18$ & \cellcolor{mygreen}$457.04$ & \cellcolor{mygreen}$519.59$ & \cellcolor{myred}$603.57$ & \cellcolor{myred}$620.67$ & \cellcolor{mygreen}$\mathbf{368.4}$ \\ 
    
     kddcup & $100$ & \cellcolor{myred}$1.5\mathrm{e}4$ & \cellcolor{myred}$1.05\mathrm{e}4$ & \cellcolor{mygreen}$\mathbf{3155.97}$ & \cellcolor{myred}$1.4\mathrm{e}4$ & \cellcolor{mygreen}$8327.2$ & \cellcolor{mygreen}$3343.25$ \\ 
    
     Twitter & $100$ & \cellcolor{myred}$1.3\mathrm{e}5$ & \cellcolor{myred}$1.04\mathrm{e}5$ & 
     \cellcolor{mygreen}$2.8\mathrm{e}4$ & \cellcolor{mygreen}$8.01\mathrm{e}4$ & \cellcolor{myred}$1.03\mathrm{e}5$ & \cellcolor{mygreen}$\mathbf{2.1\mathrm{e}4}$ \\ 

    \bottomrule
    
    \multicolumn{8}{l}{\makecell[l]{Energy usage is average of 10 trials. \textbf{\textcolor{myred}{Red}} indicate usage is higher than median,\textbf{\textcolor{mygreen}{Green}} indicate\\ usage is lower. Entries in \textbf{Bold} denote lowest consumption.}}\\
    \end{tabular}

\end{table}

%% file: Table7.tex
\begin{table}[t]
    \centering
    \caption{\textbf{Energy Cost (Joules) on Memory}}
    \label{tab:table_energy_cost_memory}

    \begin{tabular}{lccccccc}
    \toprule
    \textsf{Data} & \textsf{Clusters ($k$)} & \textsf{\ham{}} & \textsf{\ann{}} & \textsf{\exp{}} & \textsf{\kd{}} & \textsf{Ball-$k$m} & \textsf{Geo-$k$m}\\
    \midrule

    Breastcancer & $30$ & \cellcolor{myred}$1.93$ & \cellcolor{mygreen}$1.5$ & \cellcolor{mygreen}$1.53$ & \cellcolor{myred}$2.15$ & \cellcolor{myred}$1.72$ & \cellcolor{mygreen}$\mathbf{1.27}$ \\ 

     CreditRisk & $30$ & \cellcolor{myred}$1.36$ & \cellcolor{mygreen}$1.18$ & \cellcolor{mygreen}$1.21$ & \cellcolor{myred}$1.85$ & \cellcolor{myred}$1.32$ & \cellcolor{mygreen}$\mathbf{1.16}$ \\ 
    
     Sensor & $80$ & \cellcolor{myred}$1652.8$ & \cellcolor{myred}$804.65$ & \cellcolor{mygreen}$569.1$ & \cellcolor{mygreen}$590.19$ & \cellcolor{myred}$624.94$ & \cellcolor{mygreen}$\mathbf{329.36}$ \\ 
    
     kegg & $80$ & \cellcolor{myred}$342.39$ & \cellcolor{mygreen}$159.98$ & \cellcolor{mygreen}$182.52$ & \cellcolor{myred}$210.19$ & \cellcolor{myred}$215.17$ & \cellcolor{mygreen}$\mathbf{128.1}$ \\ 
    
     kddcup & $100$ & \cellcolor{myred}$5785.15$ & \cellcolor{myred}$3844.5$ & \cellcolor{mygreen}$\mathbf{1156.63}$ & \cellcolor{myred}$5203.68$ & \cellcolor{mygreen}$3087.42$ & \cellcolor{mygreen}$1237.72$ \\ 
    
     Twitter & $100$ & \cellcolor{myred}$5.5\mathrm{e}4$ & \cellcolor{myred}$4.4\mathrm{e}4$ & \cellcolor{mygreen}$1.1\mathrm{e}4$ & \cellcolor{mygreen}$3.3\mathrm{e}4$ & \cellcolor{myred}$4.3\mathrm{e}4$ & \cellcolor{mygreen}$\mathbf{8774.85}$ \\
    
    \bottomrule
    
    \multicolumn{8}{l}{\makecell[l]{Energy usage is average of 10 trials. \textbf{\textcolor{myred}{Red}} indicate usage is higher than median, \textbf{\textcolor{mygreen}{Green}} indicate\\ usage is lower. Entries in \textbf{Bold} denote lowest consumption.}}
    \end{tabular}
\end{table}

%% file: Table8.tex
\begin{table}[!htbp]
\caption{\textbf{Correlation between runtime, distance and energy.}}
\label{tab:table_pearson_corr}
        \begin{tabular}{lcc}
                
            \textsf{Algorithm} & $r(RT, EC)$ & $r(DC, EC)$ \\
            
            \midrule
            
            \ham{} & 0.99 & 0.98 \\ 
    
            \ann{} & 0.99 & 0.99 \\ 
            
            \exp{} & 0.99 & 0.99 \\
            
            \kd{} & 0.99 & 0.91 \\ 
            
            \kb{} & 0.99 & 0.97 \\ 
            
            \kg{} & 0.99 & 0.79 \\

            \bottomrule
             \multicolumn{3}{l}{\makecell[l]{$r(RT, EC)$ denote correlation is computed between\\ runtime and energy cost, similarly for $r(DC, EC)$.}}
        \end{tabular}
\end{table}

%% file: ConclusionFutureWork.tex
\section{Summary \& Future Work}\label{sec:summary}
Herein, we unveil an innovative, geometrically inspired optimization of the \km{} algorithm, which we call as \kg{}. By elucidating the relationship between high expressive (HE), distance computations, and scalar projection, \kg{} can achieve notable reductions in both runtime and DC, and also upholds solution quality. Through extensive experimentation spanning both real-world and synthetically generated data, \kg{} emerges as a significant catalyst for expediting convergence, positioning itself as a promising alternative to existing bounded and unbounded methodologies. We note that designing faster versions of the \km{} is also directly useful to single-cell RNA sequencing, due to the increasing size and complexity of scRNA-seq datasets. \citep{kiselev2019challenges, svensson2018exponential} emphasize the need of scalable clustering algorithms for efficiently processing these datasets and extracting meaningful biological insights. By addressing the challenges of data size and scalability, \kg{} can significantly contribute to real world use cases for advancing scRNA-seq data analysis.

A pivotal facet of this work lies in the initiation of a discourse on the energy costs associated with state-of-the-art clustering algorithms. As expounded in Section \ref{sec:results_energy_usage}, our analysis unveils hitherto unexplored dimensions of existing clustering techniques. Notably, while an algorithm may exhibit superior speed on certain datasets, this does not invariably translate to enhanced downstream energy efficiency. Apart from clustering, this insight also underscores the potential for designing algorithms with enhanced energy efficiency, thereby mitigating the environmental impact from intensive computational workloads. Furthermore, \kg{} manifests notably lower energy requisites, rendering it amenable for deployment in resource-constrained environments. 

Looking ahead, our aspirations encompass the formulation of a comprehensive framework tailored for iterative AI algorithms, leveraging insights from \kg{}. A possible direction would be to explore if the principles of geometry as used in \kg{}, can they be used to expedite convergence for soft-clustering such as EM, and if its possible to enhance results by combining affine product with bounded techniques, possibly leading to clustering which adopt best of both viewpoints. Additionally, we endeavor to explore the bounds on expressiveness and distance computations, inherent in this technique, while striving towards a formal, non-procedural characterization, if one exists.


%% file: Appendix.tex
\section{Supplementary Proofs}
\input{supplementary_proofs}

\section{Experiment Infrastructure}\label{secA1}

\subsubsection{Computing system}\label{sec:app_experimental_system}
Experiments on real and synthetic data were done on a 64-bit Ubuntu Linux system with 1.5TB of memory and 24 dual socket Intel Xeon cores. The C++ code used in real-world and synthetic data experiments was compiled with g++ version 9.4.0 with -O3 flag enabled.

\subsubsection{Synthetic data}\label{sec:app_data_generation_process}
Synthetic datasets are generated by sampling from multivariate Gaussian distribution. Dataset are well balanced by ensuring equal number of samples in each cluster. For well separated data, on each axis, cluster mean are separated by 3 units and covariance parameters is selected from the range $\langle1, 5\rangle$ to simulate low overlap. To increase cluster overlap (to resemble more difficult clustering tasks), separation between cluster means is decreased periodically by 1.5 units (thereafter by 0.5 units), and covariance is gradually increased in the range $\langle8, 15\rangle$.

\section{Additional Results}

\subsection{Distance computation and scalar projection}\label{sec:app_cost_dist_comp_vs_scal_proj}
We perform supplementary experiments to demonstrate the computational efficiency of scalar projection. To obtain robust estimate, we use the larger datasets in this experiment i.e. large synthetic data ($\{ 2, 4, 5, 8, 10\} \times 10^5$) and real world Twitter data. Clusters are fixed at $k = \langle 1000, 5000, 10000 \rangle$. We record the time to compute distances and affine vector products between all points in dataset and centroids. For each $k$, 10 trials are done, and average time is reported. Table \ref{tab:app_scal_proj_dist_comp} present the results. In comparison to euclidean distances, time of scalar projection is negligible. While computation time for scalar projection increase at minuscule rate; time to compute distances increase sharply with increase in data size ($m$) and clusters ($k$). Evidently, computing distance is slower due to approximations involved in square root operation [1, 2]. Consequently, \kg{} becomes faster by skipping full distance computations guided by the affine product.

\input{Table9}

\subsection{Clustering Assignments in \kg{}}\label{app:exp_real_data_accuracy}
Table \ref{tab:accuracy_table} show that \km{} and \kg{} converge to the same solution. Solution quality is assessed via: a) sum of squared errors by using the centroids obtained at convergence, and b) adjusted rand index (ARI) \citep{hubert1985comparing}. ARI measure the extant of similarity between different clustering assignments and its value range between $0-1$, where $0$ indicates random labeling and $1$ denotes perfect match. \kg{} has the same $SSE$ as \km{} (showing both algorithms produced same centroids). Further, the ARI score is $1$ when comparing the clustering labels of \kg{} with \km{}, indicating that both algorithms have identical clustering assignments.

\input{Table5}

\subsection{Experiments with $k$++ initialization}\label{app:kpp_comparison}
Table~\ref{tab:table_kpp_distances} and \ref{tab:table_kpp_runtime} present the results for $k$++ initialization. In terms of distance computations, \kg{} consistently performs the fewest computations across all datasets, with distance computation speed-ups of $83-99\%$ over baseline \km{}. This performance advantage is particularly pronounced on larger datasets such as Twitter ($98-99\%$) and Kddcup ($97-99\%$). For runtime efficiency, \kg{} performs similar or better than other methods, with runtime speed-ups ranging from $64-99\%$ compared to \km{}.

\input{table_realdata_kpp_distances}

\input{table_realdata_kpp_runtime}

\subsection{Additional scalability experiments}\label{sec:app_scal_10_million}

\begin{figure}[t]
\centering
\includegraphics[width=\textwidth]{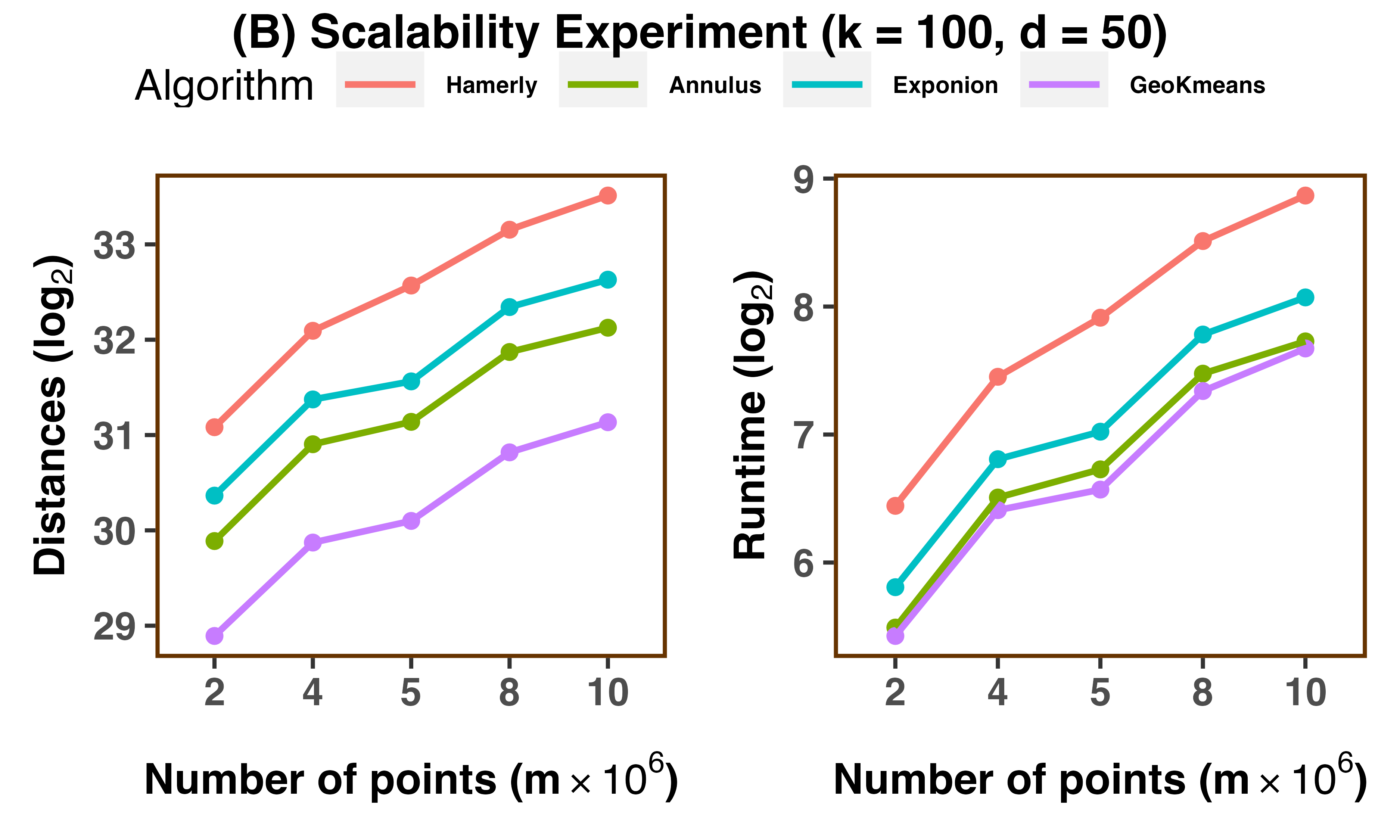}
\caption{\textbf{Clustering large synthetic data (upto 10 million points):} In DC, \kg{} is significantly better than other algorithms. In runtime, \kg{} and \ann{} are similar, though \kg{} is marginally better.}
\label{fig:fig_scal_10_million} 
\end{figure}

Fig.\,\ref{fig:fig_scal_10_million} compares the performance, specifically runtime and DC on larger dataset ($2-10$ million points), generated as outlined in \ref{sec:app_data_generation_process}. The comparison keeps the number of clusters ($k = 100$) and the dimensionality of the data ($d = 50$) constant, while varying the size of the dataset ($m$) through the values $2, 4, 5, 8, 10$ million points. \kb{} was excluded from the comparison as it encountered a segmentation fault and could not complete the first iteration (as seen previously, \kd{} is removed due to early termination). When $m$ is increased from 2 to 10 million points, we see that all algorithms do better than \ham{}. \exp{} is outperformed by \ann{} in these experiments. In runtime, \ann{} and \kg{} are similar on $m = \langle 2, 10 \rangle$ million points, whereas \kg{} does better on $m = \langle 4,5,8\rangle$ million points. Given that \kg{} is more resource efficient (consumes less energy), when comparing runtime, even though \kg{} and \ann{} are similar on some instances, \kg{} is relatively faster and has the advantage of significantly less energy consumption on larger datasets.

\subsection{Computation time per iteration}\label{sec:app_runtime_per_iter}

Following table show the runtime per iteration on Breastcancer, CreditRisk, kddcup and Twitter datasets.

\input{Table10}

\subsection{Comparative analysis with available implementation}\label{app:comparison_available_implementations}

We compare \kg{} with the implementations of \ham{} and \ann{} available at \href{https://github.com/ghamerly/fast-kmeans/blob/master/src/hamerly_kmeans.cpp}{Hamerly}. The results are shown in Table~\ref{tab:comparison_with_ham}. These new experiments yield results consistent with our original findings, confirming the relative performance characteristics we observed. The performance patterns across different datasets remain consistent with our original implementation results. Due to unavailability of DC in \href{https://github.com/ghamerly/fast-kmeans/blob/master/src/hamerly_kmeans.cpp}{Hamerly}, only runtime comparison is done.

\input{Table11}

\subsection{Relative Contributions of LE, LHE, and Neighborhood Optimizations in Efficiency}\label{app:profiling_optimization}

To understand the impact of specific design choices in \kg{}, we conducted additional experiments and recorded the DC savings contributed by specific components, sampled periodically. The results are given in Table~\ref{tab:savings1} and \ref{tab:savings2}, these include the following,

    \begin{enumerate}

        \item \textbf{Column 1:} Iteration at which the data is recorded.
        
        \item \textbf{Column 2:} Actual number of neighbor centroids out of total possible, and within the brackets, fraction of total neighborhood size. For example, if $k=80$, then theoretically neighborhood size can reach $k^2 = 6400$, where data point assigned to a given centroid will be compared with every other centroid. In practice, however, this is rarely observed (Column 1 in Table~\ref{tab:savings1} and \ref{tab:savings2}). Thus, neighborhood design saves a significant portion of DC.
        
        \item \textbf{Column 3:} Number of LE data points, and inside the brackets-LE data points as a fraction of the entire datasets. LE data points entirely skips the DC. 
        
        \item \textbf{Column 4:} Number of possible DC after accounting for LE data points i.e. $m \: - \: |LE| \: \times k \;, m=$ number of total data points in the dataset, and $k=$ number of clusters, and within the brackets-actual number of LHE data points, and percentage DC saved by only checking the LHE data points.

        \item \textbf{Column 5:} Number of possible DC due to LHE data points i.e. $|LHE| \times k$, and within the brackets-actual number of HE data points, and DC saved by only performing the computations for HE points.
    \end{enumerate}

As mentioned in Sec. 2 (Related work) in the manuscript, we expect the LE points to increase, and conversely LHE and HE points to decrease, as algorithm proceeds, due to centroids becoming stable, thus savings substantial number of DC. The results provides insights into the algorithmic design and the quantum of savings in \kg{}. For example, on the smaller CreditRisk data: 
    
    \begin{enumerate}
        \item Actual number of neighbor centroids is between $174-205$ (roughly $7-8\%$ of the possible $k^2 = 2500$ neighbor centroids).
        
        \item Number of LE data points represents a significant $81-85\%$ proportion of the total data. This corresponds to a direct reduction in DC, since LE points skip the computations.
        
        \item Number of LHE data points varies between $203-257$, significantly less than the maximum possible DC ($7250-9400$), corresponding to $\approx 97\%$ savings in DC.
        
        \item Number of data points that actually qualify as HE (out of LHE) is much less than the total possible DC, contributing to almost $99\%$ DC savings in a given iteration.
    \end{enumerate}
    
In general, as the number of LHE and HE points decrease, number of LE points increase. The savings contributed by each design choice in \kg{} becomes more substantial as data size increase.

\input{Savings_Table1}

\input{Savings_Table2}

\section{Profiling Memory Usage}\label{sec:app_memory_usage}

Table~\ref{tab:table_space_usage} show the memory allocation for \kg{} and other cluster distance based algorithms - \ham{}, \ann{} and \exp{}. Our findings show that, memory usage of \kg{} remains competitive with other methods. Our empirical evaluations in the manuscript combined with findings in Table 2, show that this more modest memory footprint does not impede \kg{} performance advantages in terms of reduced distance computations and runtime.

\input{Table12}

\section{References}

[1] Soderquist, Peter, and Miriam Leeser. ``Area and performance tradeoffs in floating-point divide and square-root implementations." ACM Computing Surveys (CSUR) 28.3 (1996): 518-564.

[2] Soderquist, Peter, and Miriam Leeser. ``Division and square root: choosing the right implementation." IEEE Micro 17.4 (1997): 56-66.

[3] Hastie, Trevor, et al. ``Multi-class adaboost." Statistics and its Interface 2.3 (2009): 349-360.

%% file: supplementary_proofs.tex
\subsection{Low Expressive Data}\label{sec:app_le_proof}

\begin{lemma}[distance and \textsf{LE}]
A data point $\mathbf{x}$ assigned to $\mathbf{c}_i$ is defined as $\mathsf{LE}$, $\exists \mathbf{c}_j \in n(\mathbf{c}_i)$, if $d(\mathbf{x}, \mathbf{c}_i) < \frac{1}{2} d(\mathbf{c}_i, \mathbf{c}_j)$. Such a $\mathsf{LE}$ data point $\mathbf{x}$ cannot change its membership and will remain assigned to $\mathbf{c}_i$; consequently, a distance computation is not required. Notice that among all neighbors of $\mathbf{c}_i$, the closest one is the natural candidate to check whether there exists a $\mathbf{c}_j$ confirming our definition of \textsf{LE}. Therefore, it is enough to check the LE condition on the closest neighboring centroid $\mathbf{c}_j$ of $\mathbf{c}_i$.
\BlankLine

\noindent \textbf{\textsc{Proof}} 
We prove that, if $d(\mathbf{x}, \mathbf{c}_i) < \frac{1}{2} d(\mathbf{c}_i, \mathbf{c}_j)$, then $d(\mathbf{x}, \mathbf{c}_i) < d(\mathbf{x}, \mathbf{c}_j)$, which implies that $\mathbf{x}$ is nearer to $\mathbf{c}_i$ and will remain assigned to $\mathbf{c}_i$.

Assume: $d(\mathbf{x}, \mathbf{c}_i) < \frac{1}{2} d(\mathbf{c}_i, \mathbf{c}_j)$. For a positive constant $\epsilon > 0$, we can rewrite $d(\mathbf{x}, \mathbf{c}_i)$ as:
\begin{eqnarray}
d(\mathbf{x}, \mathbf{c}_i) &=& \frac{1}{2} d(\mathbf{c}_i, \mathbf{c}_j) - \epsilon \; (\epsilon > 0) \label{eq:init_assume}\\
d(\mathbf{x}, \mathbf{c}_i) + d(\mathbf{x}, \mathbf{c}_j) &\geq& d(\mathbf{c}_i, \mathbf{c}_j) \; (\triangle\: inequality) \label{eq:tri_ineq}\\
\frac{1}{2} d(\mathbf{c}_i, \mathbf{c}_j) - \epsilon + d(\mathbf{x}, \mathbf{c}_j) &\geq& d(\mathbf{c}_i, \mathbf{c}_j) \; (using \: Eqs.\ \: \ref{eq:init_assume} \: \& \: \ref{eq:tri_ineq})\\
d(\mathbf{x}, \mathbf{c}_j) &\geq& \frac{1}{2} d(\mathbf{c}_i, \mathbf{c}_j) + \epsilon \label{eq:init_conclusion}\\
d(\mathbf{x}, \mathbf{c}_j) &>& d(\mathbf{x}, \mathbf{c}_i) \; (using\: Eqs.\ \ref{eq:init_assume} \: \& \: \ref{eq:init_conclusion})
\end{eqnarray}

Therefore, a data point $\mathbf{x}$ assigned to $\mathbf{c}_i, \exists \mathbf{c}_j \in n(\mathbf{c}_i)$, if distance between $\mathbf{x}$ and $\mathbf{c}_i$ is less than $\frac{1}{2} d(\mathbf{c}_i, \mathbf{c}_j)$, then $\mathbf{x}$ will remain assigned to $\mathbf{c}_i$, and distance computation can be skipped. An illustration of \textsf{LE} points is shown in Fig.\,\ref{fig:low_expressive}. We note that when $d(\mathbf{x}, \mathbf{c}_i) = \frac{1}{2} d(\mathbf{c}_i, \mathbf{c}_j)$ then assignment of data by \kg{} is identical to \km{}, given ties are broken evenly, this is discussed in \ref{sec:proof_le_when_aequalc}.
\hfill $\blacksquare$\\
\end{lemma}

\subsection{Distance and $\mathsf{LE}$}\label{sec:proof_le_when_aequalc}

For a data point $\mathbf{x}$ assigned to $\mathbf{c}_i, \: \exists \mathbf{c}_j \in n(\mathbf{c}_i)$, if $d(\mathbf{x}, \mathbf{c}_i) = \frac{1}{2} d(\mathbf{c}_i, \mathbf{c}_j)$; whether membership of $\mathbf{x}$ changes or not depends on how one choose to break the ties.

\noindent \textbf{\textsc{Discussion}} \\Assume: $d(\mathbf{x}, \mathbf{c}_i) = \frac{1}{2} d(\mathbf{c}_i, \mathbf{c}_j)$. Then:

\begin{eqnarray}
2d(\mathbf{x}, \mathbf{c}_i) &=& d(\mathbf{c}_i, \mathbf{c}_j) \label{eq:assume1}
\end{eqnarray}

From triangle inequality: $d(\mathbf{c}_i, \mathbf{c}_j) \leq d(\mathbf{x}, \mathbf{c}_i) + d(\mathbf{x}, \mathbf{c}_j)$. We can write $d(\mathbf{c}_i, \mathbf{c}_j)$ as:

\begin{equation}
d(\mathbf{c}_i, \mathbf{c}_j) = d(\mathbf{x}, \mathbf{c}_i) + d(\mathbf{x}, \mathbf{c}_j) - \epsilon \; (\epsilon \geq 0)\label{eq:tri_conseq}
\end{equation}

Using Eqs.\ \ref{eq:assume1} and \ref{eq:tri_conseq}
\begin{eqnarray}
2d(\mathbf{x}, \mathbf{c}_i) &=& d(\mathbf{x}, \mathbf{c}_i) + d(\mathbf{x}, \mathbf{c}_j) - \epsilon\label{eq:epsilon_and_tri_ineq}\\
d(\mathbf{x}, \mathbf{c}_i) &=& d(\mathbf{x}, \mathbf{c}_j) - \epsilon\\
d(\mathbf{x}, \mathbf{c}_i) + \epsilon &=& d(\mathbf{x}, \mathbf{c}_j)\label{eq:epsilon_cases}
\end{eqnarray}

\textsc{\textbf{Case 1}}: If $\epsilon = 0$:

\begin{equation}
    d(\mathbf{x}, \mathbf{c}_i) = d(\mathbf{x}, \mathbf{c}_j)\label{eq:epsilon_case_1}
\end{equation}

From Eq.\ \ref{eq:epsilon_case_1}, $\mathbf{x}$ is equidistant from $\mathbf{c}_i$ and $\mathbf{c}_j$, hence we have a tie for membership; therefore one can choose to either keep $\mathbf{x}$ assigned to $\mathbf{c}_i$, or re-assign $\mathbf{x}$ to $\mathbf{c}_j$. It depends on how one choose to implement the algorithm. However, \textbf{as long as $\mathbf{x}$ is assigned the same way in \km{} and \kg{}} - both algorithms will converge to the same solution. In our implementation, we keep $\mathbf{x}$ assigned to same cluster as \km{}. This is similar to considering $\mathbf{x}$ as $LE$, since $LE$ does not change its membership. 

\textsc{\textbf{Case 2}}: If $\epsilon > 0$:

\begin{equation}
    d(\mathbf{x}, \mathbf{c}_j) = d(\mathbf{x}, \mathbf{c}_i) + \epsilon \; (from \: Eq.\ \: \ref{eq:epsilon_cases}) \label{eq:epsilon_case_2}
\end{equation}

If $\epsilon>0$ then Eq.\ \ref{eq:epsilon_case_2} is true when: $d(\mathbf{x}, \mathbf{c}_j) > d(\mathbf{x}, \mathbf{c}_i)$. Hence, $\mathbf{x}$ is nearer to $\mathbf{c}_i$ and will not change its membership. Therefore, distance does not need to be computed.

From \textbf{case 1}, we observe that $\mathbf{x}$ can be treated as $LE$ by keeping it assigned to the current cluster. Further, if ties are broken in same way in both \km{} and \kg{} then both algorithms will converge to the same solution. From \textbf{case 2}, we see that $\mathbf{x}$ is closer to $\mathbf{c}_i$ than $\mathbf{c}_j$; hence distance computation is not required. Therefore, if $d(\mathbf{x}, \mathbf{c}_i) = \frac{1}{2} d(\mathbf{c}_i, \mathbf{c}_j)$ then $\mathbf{x}$ can be safely treated as $\mathsf{LE}$.

\subsection{Distance and $\mathsf{LHE}$}\label{sec:proof_he_case}
For a data point $\mathbf{x}$ assigned to $\mathbf{c}_i, \: \forall \mathbf{c}_j \in n(\mathbf{c}_i)$, $\mathbf{x}$ is $\mathsf{LHE}$ if $d(\mathbf{x}, \mathbf{c}_i) > \frac{1}{2} d(\mathbf{c}_i, \mathbf{c}_j)$; As a result, $\mathbf{x}$ may or may not change membership, therefore requires a distance computation.

\noindent \textbf{\textsc{Discussion}} \\Assume: $d(\mathbf{x}, \mathbf{c}_i) > \frac{1}{2} d(\mathbf{c}_i, \mathbf{c}_j)$. Then:

\begin{equation}
2d(\mathbf{x}, \mathbf{c}_i) > d(\mathbf{c}_i, \mathbf{c}_j) \label{eq:assume}
\end{equation}

From triangle inequality:
\vspace{-0.5cm}

\begin{eqnarray}
d(\mathbf{c}_i, \mathbf{c}_j) &\leq& d(\mathbf{x}, \mathbf{c}_i) + d(\mathbf{x}, \mathbf{c}_j)\\
d(\mathbf{c}_i, \mathbf{c}_j) &=& d(\mathbf{x}, \mathbf{c}_i) + d(\mathbf{x}, \mathbf{c}_j) - \epsilon \; (\epsilon \geq 0) \label{eq:assume_0}\\
2d(\mathbf{x}, \mathbf{c}_i) &>& d(\mathbf{x}, \mathbf{c}_i) + d(\mathbf{x}, \mathbf{c}_j) - \epsilon \;\; (using \: Eqs.\ \ref{eq:assume} \:and\: \ref{eq:assume_0}) \label{eq:epsilon_and_tri_ineq}
\end{eqnarray}

\textsc{\textbf{Case 1}}: $\epsilon = 0$
\begin{eqnarray}
    2d(\mathbf{x}, \mathbf{c}_i) &>& d(\mathbf{x}, \mathbf{c}_i) + d(\mathbf{x}, \mathbf{c}_j)\\
    d(\mathbf{x}, \mathbf{c}_i) &>& d(\mathbf{x}, \mathbf{c}_j)
\end{eqnarray}

For $\epsilon = 0$: $\mathbf{x}$ moves nearer to $\mathbf{c}_j$; therefore it can not remain assigned to $\mathbf{c}_i$ and distance computation is required.

\textsc{\textbf{Case 2}}: $\epsilon > 0$

\vspace{-0.5cm}

\begin{align}
    d(\mathbf{x}, \mathbf{c}_i) > \frac{1}{2}d(\mathbf{x}, \mathbf{c}_i) + \frac{1}{2}d(\mathbf{x}, \mathbf{c}_j) - \frac{1}{2}\epsilon \;\; (using \: Eq.\: \ref{eq:epsilon_and_tri_ineq})\\
    \frac{1}{2}d(\mathbf{x}, \mathbf{c}_i) > \frac{1}{2}d(\mathbf{x}, \mathbf{c}_j) - \frac{1}{2}\epsilon\\
    d(\mathbf{x}, \mathbf{c}_i) > d(\mathbf{x}, \mathbf{c}_j) - \epsilon \;\; \label{eq:lhe_eq_1}
\end{align}

For $\epsilon > 0$: whether $\mathbf{x}$ has moved nearer to $\mathbf{c}_j$, depends on the exact values of $\epsilon$, $d(\mathbf{x}, \mathbf{c}_i)$ and $d(\mathbf{x}, \mathbf{c}_j)$. Hence, until we re-calculate the distance, we can not be sure that $\mathbf{x}$ will change its membership. For example: assume $d(\mathbf{x}, \mathbf{c}_i) = 20$, $d(\mathbf{x}, \mathbf{c}_j) = 30$, and $\epsilon = 11$. Even though Eq.\ \ref{eq:lhe_eq_1} is true i.e. 20 $>$ 30 - 11 $\implies$ 20 $>$ 19, but 20 $\ngtr$ 30 $\implies$ $d(\mathbf{x}, \mathbf{c}_i) \ngtr d(\mathbf{x}, \mathbf{c}_j)$. Hence, $\mathbf{x}$ is still nearer to $\mathbf{c}_i$ and will not change membership. On the other hand, assume $d(\mathbf{x}, \mathbf{c}_i) = 5$, $d(\mathbf{x}, \mathbf{c}_j) = 4$, and $\epsilon = 1$, in this case: 5 $>$ 4 - 1 $\implies$ 5 $>$ 3, also 5 $>$ 4 $\implies$ $d(\mathbf{x}, \mathbf{c}_i) > d(\mathbf{x}, \mathbf{c}_j)$. Hence, $\mathbf{x}$ has moved nearer to $\mathbf{c}_j$ and will change membership. Consequently, resolving these cases requires a distance computation.

From \textsc{\textbf{Case 1}}, $\mathbf{x}$ will change its membership and distance computation is necessary, but in \textsc{\textbf{Case 2}}, $\mathbf{x}$ may or may not change its membership and we can not be sure unless the distances are re-computed. Therefore, for a data point $\mathbf{x}$ assigned to $\mathbf{c}_i, \: \forall \mathbf{c}_j \in n(\mathbf{c}_i)$, $\mathbf{x}$ is $\mathsf{LHE}$ if $d(\mathbf{x}, \mathbf{c}_i) > \frac{1}{2} d(\mathbf{c}_i, \mathbf{c}_j)$; As a result, $\mathbf{x}$ may or may not change membership, and mandates a distance computation.

More importantly, for ascertaining if a data point will change its membership or not, one does not need to compute distances, rather, both DC and compute time can be saved by using vector affine product. As a result, there is no need to perform DC for \textsf{LHE} points which won't change their membership. By Lemma \ref{lemma1}, we know that $\textsf{LE}$ data does not change membership, from \ref{sec:proof_he_case}, we know that \textsf{LHE} points can impose redundant computations, therefore significant runtime savings can be achieved by selectively performing the distance calculations only for $\textsf{HE}$. Empirical evaluations provided in the paper supports this analysis.

\subsection{Proof of Convergence}\label{app:convergence_proof}

\input{convergence_proof_appendix_1}

In order to partition $m$ points in $k$ clusters, each point can be assigned to one of $k$ clusters. There are $k^m$ ways to partition the data, since each clustering is unique (no clustering is visited twice), there are a finite number of possible clustering, hence \kg{} must terminate. From the pseudo code in algorithm\, 2, we know that \kg{} will naturally terminate when $\mathbf{C}^{t+1} = \mathbf{C}^{t}$ i.e. centroids does not change between successive iterations. It suffice to show that $SSE$ is non-increasing between successive iterations i.e. $SSE(D, \mathbf{C}^{(t+1)})_{\mathbf{A}^{(t+1)}} < SSE(D, \mathbf{C}^{(t)})_{\mathbf{A}^{(t)}}$. We proof this in two steps.\\

\input{convergence_proof_appendix_2}

%% file: convergence_proof_appendix_1.tex
\begin{lemma}\label{lemma:convergence_proof_part1}
Given a set of data points $D = \{ \mathbf{x}_{1}, \ldots, \mathbf{x}_m \}$ and $m > 1$. The actual centroid of points in $D$ can be found as: $\mathbf{c} = \frac{1}{m} \sum_{i=1}^m \mathbf{x}_i$, let $\mathbf{x}_p$ be any arbitrary point in $D$. We show that $SSE(D, \mathbf{c}) \leq SSE(D, \mathbf{x}_p)$.

\noindent \textbf{\textsc{Proof}}

\begin{equation}\label{eq:proof_conv_1}
    SSE(D, \mathbf{c}) = \; \sum_{i=1}^m || \mathbf{x}_i - \mathbf{c}||^2
\end{equation}
$\because (\mathbf{x}_p - \mathbf{x}_p) = 0$, we can put this term inside eq. \ref{eq:proof_conv_1} without affecting the $SSE$.
\begin{align}
    SSE(D, \mathbf{c}) = \sum_{i=1}^m \left[ || \mathbf{x}_i + (\mathbf{x}_p - \mathbf{x}_p) - \mathbf{c}||^2 \right]\\
    = \sum_{i=1}^m \; \left[ || (\mathbf{x}_i - \mathbf{x}_p) + (\mathbf{x}_p - \mathbf{c}) ||^2 \right]\label{eq:conv_eq_ref_1}
\end{align}
Eq. \ref{eq:conv_eq_ref_1} can be expanded as follows:
\begin{align}
    = \sum_{i=1}^m \; \left[ \; || \mathbf{x}_i - \mathbf{x}_p ||^2 + || \mathbf{x}_p - \mathbf{c} ||^2 + 2 \; (\mathbf{x}_i - \mathbf{x}_p) \cdot (\mathbf{x}_p - \mathbf{c}) \; \right]\\
    = \sum_{i=1}^m \; || \mathbf{x}_i - \mathbf{x}_p ||^2 + \sum_{i=1}^m \; || \mathbf{x}_p - \mathbf{c} ||^2 + 2 \; \sum_{i=1}^m  \left\{ (\mathbf{x}_i - \mathbf{x}_p) \cdot (\mathbf{x}_p - \mathbf{c}) \right\}\label{eq:intermediate_eq1}
\end{align}
Note that second term in Eq. \ref{eq:intermediate_eq1} does not contain subscript over $i$, it can be written as $m \cdot ||\mathbf{x}_p - \mathbf{c}||^2$.
\begin{align}
    = \sum_{i=1}^m \; || \mathbf{x}_i - \mathbf{x}_p ||^2 + m \cdot ||\mathbf{x}_p - \mathbf{c}||^2 + 2 \; \sum_{i=1}^m  \left\{ (\mathbf{x}_i - \mathbf{x}_p) \cdot (\mathbf{x}_p - \mathbf{c}) \right\}
\end{align}
In the third term, $(\mathbf{x}_p - \mathbf{c})$ does not have subscript over $i$, we can move it out of the summation.
\begin{align}
    = \sum_{i=1}^m \; || \mathbf{x}_i - \mathbf{x}_p ||^2 + m \cdot ||\mathbf{x}_p - \mathbf{c}||^2 + 2 \cdot \; (\mathbf{x}_p - \mathbf{c}) \; \cdot \left\{ \sum_{i=1}^m \; (\mathbf{x}_i - \mathbf{x}_p) \right\}\\
    = \sum_{i=1}^m \; || \mathbf{x}_i - \mathbf{x}_p ||^2 + m \cdot ||\mathbf{x}_p - \mathbf{c}||^2 + 2 \cdot \; (\mathbf{x}_p - \mathbf{c}) \; \cdot \left\{ (\sum_{i=1}^m \; \mathbf{x}_i  - \sum_{i=1}^m \mathbf{x}_p) \right\}
\end{align}
$\because \mathbf{c} = \frac{1}{m}\sum_{i=1}^m \mathbf{x}_i \; \Rightarrow m \mathbf{c} = \sum_{i=1}^m \mathbf{x}_i$.
\begin{align}
    = \sum_{i=1}^m \; || \mathbf{x}_i - \mathbf{x}_p ||^2 + m \cdot ||\mathbf{x}_p - \mathbf{c}||^2 + 2 \cdot \; (\mathbf{x}_p - \mathbf{c}) \; \cdot (m \mathbf{c} - m \mathbf{x}_p)\\
    = \sum_{i=1}^m \; || \mathbf{x}_i - \mathbf{x}_p ||^2 + m \cdot ||\mathbf{x}_p - \mathbf{c}||^2 + 2 \cdot \; (\mathbf{x}_p - \mathbf{c}) \; \cdot (-m \cdot (\mathbf{x}_p - \mathbf{c}) )\\
    = \sum_{i=1}^m \; || \mathbf{x}_i - \mathbf{x}_p ||^2 + m \cdot ||\mathbf{x}_p - \mathbf{c}||^2 - 2 \cdot m \cdot \; (\mathbf{x}_p - \mathbf{c})^2
\end{align}
$\because (\mathbf{x}_p - \mathbf{c})^2 \; = \; ||\mathbf{x}_p - \mathbf{c}||^2$
\begin{align}
    = \sum_{i=1}^m \; || \mathbf{x}_i - \mathbf{x}_p ||^2 + m \cdot ||\mathbf{x}_p - \mathbf{c}||^2 - 2 \cdot m \cdot \; ||\mathbf{x}_p - \mathbf{c}||^2\\
    = \sum_{i=1}^m \; || \mathbf{x}_i - \mathbf{x}_p ||^2 - m \cdot ||\mathbf{x}_p - \mathbf{c}||^2\\
    SSE(D, \mathbf{c}) \; = \; SSE(D, \mathbf{x}_p) - m \cdot ||\mathbf{x}_p - \mathbf{c}||^2\label{eq:intermediate_eq2}
\end{align}

From Eq. \ref{eq:intermediate_eq2}, it follows that

\begin{equation}\label{eq:conclude_conv_proof_part_1}
SSE(D, \mathbf{c}) \; \leq \; SSE(D, \mathbf{x}_p)
\end{equation}
\hfill $\blacksquare$
\end{lemma} 

%% file: convergence_proof_appendix_2.tex
\noindent \textbf{Case 1 (Centroid update):} $SSE$ is non-increasing after centroid update i.e. $SSE(D, \mathbf{C}^{(t+1)})_{\mathbf{A}^{(t+1)}} < SSE(D, \mathbf{C}^{(t)})_{\mathbf{A}^{(t+1)}}$

\vspace{-0.5cm}

\begin{align}
   SSE(D, \mathbf{C}^{(t+1)})_{\mathbf{A}^{(t+1)}} = \sum_{i=1}^{m} || \mathbf{x}_i - \mathbf{C}^{(t+1)}||_{\mathbf{A}^{(t+1)}}^2\\
   SSE(D, \mathbf{C}^{(t+1)})_{\mathbf{A}^{(t+1)}} = \sum_{i=1}^{m} \sum_{j=1}^k || \mathbf{x}_i - \mathbf{c}^{(t+1)}_{j\in\mathcal{I}(\mathbf{x}_i)}||_{\mathbf{A}^{(t+1)}}^2\label{eq:proof2_eq1}
\end{align}

Using lemma \ref{lemma:convergence_proof_part1},

\begin{equation}\label{eq:proof2_eq2}
\sum_{i=1}^{m} \sum_{j=1}^k || \mathbf{x}_i - \mathbf{c}^{(t+1)}_{j\in\mathcal{I}(\mathbf{x}_i)}||_{\mathbf{A}^{(t+1)}}^2 \; \leq \; \sum_{i=1}^{m} \sum_{j=1}^k  || \mathbf{x}_i - \mathbf{c}_{j\in\mathcal{I}(\mathbf{x}_i)}^{(t)}||_{\mathbf{A}^{(t+1)}}^2
\end{equation}

From Eqs. \ref{eq:proof2_eq1} and \ref{eq:proof2_eq2}

\begin{equation}\label{eq:proof_conv_3}
SSE(D, \mathbf{C}^{(t+1)})_{\mathbf{A}^{(t+1)}} \leq SSE(D, \mathbf{C}^{(t)})_{\mathbf{A}^{(t+1)}}
\end{equation}
\hfill
$\blacksquare$

\textbf{Case 2 (Reassignment):} $SSE$ is non-increasing after data is re-assigned i.e. $SSE(D, \mathbf{C}^{(t)})_{\mathbf{A}^{(t+1)}} < SSE(D, \mathbf{C}^{(t)})_{\mathbf{A}^{(t)}}$. Consider a data point $\mathbf{x}$ assigned to $\mathbf{c}_i$ in $\mathbf{A}^{(t)}$. From the construction of the algorithm, we know that a new clustering is only possible if $\mathbf{x}$ has become closer to centroid $\mathbf{c}_j$ ($\mathbf{c}_i \neq \mathbf{c}_j$) as a result, $\mathbf{x}$ is reassigned to $\mathbf{c}_j$ in $\mathbf{A}^{(t+1)}$. Therefore, 

\begin{equation}\label{eq:proof_conv_4}
SSE(D, \mathbf{C}^{(t)})_{\mathbf{A}^{(t+1)}} < SSE(D, \mathbf{C}^{(t)})_{\mathbf{A}^{(t)}}
\end{equation}

From Eqs. \ref{eq:proof_conv_3} and \ref{eq:proof_conv_4},

\begin{equation}
SSE(D, \mathbf{C}^{(t+1)})_{\mathbf{A}^{(t+1)}} < SSE(D, \mathbf{C}^{(t)})_{\mathbf{A}^{(t)}}
\end{equation}
\hfill
$\blacksquare$

Since $SSE$ is non-increasing, hence \kg{} must terminate in a finite number of steps.

%% file: Table9.tex
\begin{table}[!htbp]
    \centering
    \caption{\textbf{Average execution time for scalar projection and distance computation}}
    \label{tab:app_scal_proj_dist_comp}

    \begin{tabular}{cccc}
    \toprule
    \textsf{$|D| = m\mathrm{e}5$} & \textsf{Clusters} & \textsf{Distance Computation} & \textsf{Scalar Projection}\\
    \midrule

    $m$ & $k$ & RT & RT \\
    
    \cmidrule{3-3} \cmidrule{4-4} \\

    \multirow{3}{*}{$2$} & $1000$ & $1.5\mathrm{e}4$ & $\mathbf{0.000034}$ \\
    & $5000$ & $1.1\mathrm{e}5$ & $\mathbf{0.000077}$ \\
    & $10000$ & $3.3\mathrm{e}5$ & $\mathbf{0.00011}$ \\

    \midrule

    \multirow{3}{*}{$5$} & $1000$ & $3.7\mathrm{e}5$ & $\mathbf{0.00015}$ \\
    & $5000$ & $6.4\mathrm{e}5$ & $\mathbf{0.00018}$ \\
    & $10000$ & $1.2\mathrm{e}6$ & $\mathbf{0.00023}$ \\

    \midrule

    \multirow{3}{*}{$8$} & $1000$ & $1.2\mathrm{e}6$ & $\mathbf{0.00027}$ \\
    & $5000$ & $1.7\mathrm{e}6$ & $\mathbf{0.00031}$ \\
    & $10000$ & $2.4\mathrm{e}6$ & $\mathbf{0.00035}$ \\

    \midrule

    \multirow{3}{*}{$10$} & $1000$ & $2.5\mathrm{e}6$ & $\mathbf{0.00038}$ \\
    & $5000$ & $2.9\mathrm{e}6$ & $\mathbf{0.00041}$ \\
    & $10000$ & $3.6\mathrm{e}6$ & $\mathbf{0.00044}$ \\

    \midrule

    \multirow{3}{*}{\makecell[c]{Twitter\\($m = 5.8\mathrm{e}5$)}} & $1000$ & $3.6\mathrm{e}6$ & $\mathbf{0.00046}$ \\
    & $5000$ & $3.7\mathrm{e}6$ & $\mathbf{0.00049}$ \\
    & $10000$ & $4.0\mathrm{e}6$ & $\mathbf{0.00053}$ \\
    
    \bottomrule\\
    \multicolumn{4}{l}{\makecell[l]{\textsf{RT} is the runtime in milliseconds (average of 10 trials). Entries in \textbf{Bold}\\ indicate lower runtime.}}
    \end{tabular}
\end{table}

%% file: Table5.tex
\begin{table}[!htb]
    \centering
    \caption{\textbf{Accuracy comparison:} $SSE$ obtained by \km{} and \kg{}}
    \label{tab:accuracy_table}

    \begin{tabular}{lccccc}
    \toprule
    \textsf{Data} & \textsf{Clusters ($k$)} & \textsf{$SSE$ (\km{})} & \textsf{$SSE$ (\kg{})} & \textsf{$SSE$ Diff.} & \textsf{ARI}\\
    \midrule

    \multirow{3}{*}{BreastCancer} & $20$ & $6.16\mathrm{e}5$ & $6.16\mathrm{e}5$ & $0$ & $1$ \\
    & $30$ & $6.16\mathrm{e}5$ & $6.16\mathrm{e}5$ & 
    $0$ & $1$  \\
    & $50$ & $7.30\mathrm{e}5$ & $7.30\mathrm{e}5$ & $0$ & $1$ \\

    \midrule

    \multirow{3}{*}{CreditRisk} & $20$ & $2.68\mathrm{e}6$ & $2.68\mathrm{e}6$ & $0$ & $1$  \\
    & $30$ & $2.18\mathrm{e}6$ & $2.18\mathrm{e}6$ & $0$ & $1$  \\
    & $50$ & $2.07\mathrm{e}6$ & $2.07\mathrm{e}6$ & $0$ & $1$  \\

    \midrule
    
    \multirow{3}{*}{Sensor} & $30$ & $1.56\mathrm{e}7$ & $1.56\mathrm{e}7$ & $0$ & $1$ \\
    & $50$ & $1.57\mathrm{e}7$ & $1.57\mathrm{e}7$ & $0$ & $1$ \\
    & $80$ & $1.57\mathrm{e}7$ & $1.57\mathrm{e}7$ & $0$ & $1$  \\

    \midrule

    \multirow{3}{*}{Kegg} & $30$ & $1.95\mathrm{e}3$ & $1.95\mathrm{e}3$ & $0$ & $1$  \\ 
    & $50$ & $7.70\mathrm{e}3$ & $7.70\mathrm{e}3$ & $0$ & $1$ \\
    & $80$ & $8.16\mathrm{e}3$ & $8.16\mathrm{e}3$ & $0$ & $1$ \\

    \midrule
    
    \multirow{3}{*}{Kddcup} & $100$ & $4.81\mathrm{e}17$ & $4.81\mathrm{e}17$ & $0$ & $1$ \\
    & $300$ & $4.81\mathrm{e}17$ & $4.81\mathrm{e}17$ & $0$ & $1$ \\
    & $500$ & $4.81\mathrm{e}17$ & $4.81\mathrm{e}17$ & $0$ & $1$ \\

    \midrule

    \multirow{3}{*}{Twitter} & $100$ & $5.59\mathrm{e}10$ & $5.59\mathrm{e}10$ & $0$ & $1$  \\
    & $300$ & $7.62\mathrm{e}10$ & $7.62\mathrm{e}10$ & $0$ & $1$  \\
    & $500$ & $1.16\mathrm{e}10$ & $1.16\mathrm{e}10$ & $0$ & $1$  \\
    
    \bottomrule\\
    \multicolumn{6}{l}{\makecell[l]{\textsf{$SSE$ Diff.} is the difference between the $SSE$ of \km{} and \kg{}. $0$ means\\ both algorithms converged to same solution. \textsf{ARI} is adjusted rand index.}}
    \end{tabular}
\end{table}

%% file: table_realdata_kpp_distances.tex

\begin{sidewaystable}

    \centering
    \caption{\textbf{$k$++ seeding:} Comparison of average distance computations of different algorithms}
    
    \label{tab:table_kpp_distances}

    \begin{tabular}{lcccccccccccc}     
        
        \toprule
        Dataset & Clusters & \km{} & \multicolumn{2}{c}{\ham{}} & \multicolumn{2}{c}{\ann{}} & \multicolumn{2}{c}{\exp{}} & \multicolumn{2}{c}{\kb{}} &
        \multicolumn{2}{c}{\kg{}}\\
        \midrule
         & $k$ & $DC$ \ \ \ \ & $DC$ & $DC(S)$ & $DC$ & $DC(S)$ & $DC$ & $DC(S)$ & $DC$ & $DC(S)$ & $DC$ & $DC(S)$\\
         \cmidrule(lr){4-5} \cmidrule(lr){6-7} \cmidrule(lr){8-9} \cmidrule(lr){10-11} \cmidrule(lr){12-13}
         \\

            \multirow{3}{*}{BreastCancer} & $20$ & $1.39\mathrm{e}5$ \ \ \ \ & 
            $4.9\mathrm{e}4$ & $64.37$ & 
            $3.1\mathrm{e}4$ & $77.83$ & 
            $2.3\mathrm{e}4$ & $83.48$ & 
            $2.3\mathrm{e}4$ & $83.53$ & 
            $\mathbf{2.0\mathrm{e}4}$ & $\mathbf{85.68}$ \\ 
            
            & $30$ & $1.5\mathrm{e}5$ \ \ \ \ & 
            $6.7\mathrm{e}4$ & $55.31$ & 
            $4.6\mathrm{e}4$ & $68.74$ & 
            $2.9\mathrm{e}4$ & $80.04$ & 
            $2.9\mathrm{e}4$ & $80.43$ & 
            $\mathbf{2.4\mathrm{e}4}$ & $\mathbf{83.37}$ \\

            & $50$ & 
            $2.6\mathrm{e}5$ \ \ \ \ & 
            $1.3\mathrm{e}5$ & $47.19$ & 
            $9.2\mathrm{e}4$ & $64.73$ & 
            $5.2\mathrm{e}4$ & $80.04$ & 
            $5.4\mathrm{e}4$ & $79.16$ & 
            $\mathbf{4.3\mathrm{e}4}$ & $\mathbf{83.50}$ \\
            \\
            \midrule
            \\

            \multirow{3}{*}{CreditRisk} & $20$ & 
            $3.6\mathrm{e}5$ \ \ \ \ & 
            $5.2\mathrm{e}4$ & $85.62$ & 
            $3.8\mathrm{e}4$ & $89.33$ & 
            $\mathbf{3.0\mathrm{e}4}$ & 
            $\mathbf{91.62}$ & 
            $3.8\mathrm{e}4$ & $89.45$ 
            &$4.0\mathrm{e}4$ & $88.83$ \\ 
            
            & $30$ & 
            $4.3\mathrm{e}5$ \ \ \ \ & 
            $6.6\mathrm{e}4$ & $84.50$ & 
            $5.5\mathrm{e}4$ & $87.25$ & 
            $\mathbf{4.2\mathrm{e}4}$ & $\mathbf{90.23}$ & $4.9\mathrm{e}4$ & $88.56$ & 
            $4.9\mathrm{e}4$ & $88.57$ \\ 
            
            & $50$ & 
            $5.6\mathrm{e}5$ \ \ \ \ & 
            $1.0\mathrm{e}5$ & $80.61$ & 
            $9.0\mathrm{e}4$ & $83.98$ & 
            $6.9\mathrm{e}4$ & $87.72$ & 
            $8.1\mathrm{e}4$ & $85.65$ & 
            $7.3\mathrm{e}4$ & $87.07$ \\
            \\
            \midrule
            \\

            \multirow{3}{*}{Sensor} & $50$ & 
            $3.0\mathrm{e}8$ \ \ \ \ & 
            $2.5\mathrm{e}8$ & $17.32$ & 
            $7.6\mathrm{e}7$ & $74.92$ & 
            $6.5\mathrm{e}7$ & $78.55$ & 
            $3.5\mathrm{e}7$ & $88.52$ & 
            $\mathbf{9.1\mathrm{e}6}$ & $\mathbf{97.0}$ \\ 
            
            & $80$ & 
            $5.5\mathrm{e}8$ \ \ \ \ & 
            $4.4\mathrm{e}8$ & $11.95$ & 
            $1.6\mathrm{e}8$ & $70.11$ & 
            $1.1\mathrm{e}8$ & $79.77$ & 
            $5.3\mathrm{e}7$ & $90.25$ & 
            $\mathbf{1.2\mathrm{e}7}$ & $\mathbf{97.80}$ \\ 
            
            & $100$ & $7.1\mathrm{e}8$ \ \ \ \ & 
            $6.2\mathrm{e}8$ & $12.16$ & 
            $2.03\mathrm{e}8$ & $71.55$ & 
            $1.3\mathrm{e}8$ & $81.44$ & 
            $5.4\mathrm{e}7$ & $92.43$ & 
            $\mathbf{1.3\mathrm{e}7}$ & 
            $\mathbf{98.09}$ \\
            \\
            \midrule
            \\

            \multirow{3}{*}{Kegg} & $30$ & 
            $1.6\mathrm{e}8$ \ \ \ \  & 
            $1.05\mathrm{e}8$ & $35.25$ & 
            $3.5\mathrm{e}7$ & $78.22$ & 
            $4.5\mathrm{e}7$ & $71.76$ & 
            $2.1\mathrm{e}7$ & $86.78$ & 
            $\mathbf{6.5\mathrm{e}6}$ & 
            $\mathbf{95.96}$ \\ 
            
            & $50$ & $2.5\mathrm{e}8$ \ \ \ \ 
            &$1.5\mathrm{e}8$ & $38.59$ & 
            $5.5\mathrm{e}7$ & $78.31$ & 
            $5.7\mathrm{e}7$ & $77.48$ & 
            $2.4\mathrm{e}7$ & $90.48$ & 
            $\mathbf{8.5\mathrm{e}6}$ & 
            $\mathbf{96.64}$ \\
            
            & $80$ & $2.5\mathrm{e}8$ \ \ \ \ & 
            $1.5\mathrm{e}8$ & $40.77$ & 
            $5.5\mathrm{e}7$ & $78.97$ & 
            $5.7\mathrm{e}7$ & $79.97$ & 
            $2.41\mathrm{e}7$ & $91.94$ & 
            $\mathbf{8.5\mathrm{e}6}$ & $\mathbf{97.07}$ \\ 
            \\
            \midrule
            \\

            \multirow{3}{*}{Kddcup} & $100$ & 
            $2.6\mathrm{e}9$ \ \ \ \ & 
            $2.5\mathrm{e}8$ & $90.55$ & 
            $1.1\mathrm{e}8$ & $95.84$ & 
            $\mathbf{5.9\mathrm{e}7}$ & \textcolor{blue}{$97.77$} & $6.2\mathrm{e}7$ & \textcolor{blue}{$97.65$} & 
            $7.5\mathrm{e}7$ 
            & \textcolor{blue}{$97.15$} \\
            
            & $300$ & $1.2\mathrm{e}10$ \ \ \ \ & 
            $1.1\mathrm{e}9$ & $90.65$ & 
            $4.08\mathrm{e}8$ & $96.80$ & 
            $1.75\mathrm{e}8$ & \textcolor{blue}{$98.63$} & 
            $\mathbf{1.70\mathrm{e}8}$ & \textcolor{blue}{$98.66$} & $1.9\mathrm{e}8$ & \textcolor{blue}{$98.48$} \\
            
            & $500$ & 
            $2.003\mathrm{e}10$ \ \ \ \ & 
            $2.7\mathrm{e}9$ & $86.29$ & 
            $1.0\mathrm{e}9$ & $94.57$ & 
            $2.9\mathrm{e}8$ & \textcolor{blue}{$98.54$} & 
            $\mathbf{2.8\mathrm{e}8}$ & \textcolor{blue}{$98.59$} & $2.9\mathrm{e}8$ & \textcolor{blue}{$98.51$} \\
            \\
            \midrule
            \\

            \multirow{3}{*}{Twitter} & $100$ & 
            $1.21\mathrm{e}10$ \ \ \ \ & 
            $5.03\mathrm{e}9$ & $58.54$ & 
            $1.44\mathrm{e}9$ & $88.06$ & 
            $6.17\mathrm{e}8$ & $94.91$ & 
            $4.70\mathrm{e}8$ & $96.12$ & 
            $\mathbf{1.76\mathrm{e}8}$ & $\mathbf{98.54}$ \\ 
            
            & $300$ & 
            $5.3\mathrm{e}10$ \ \ \ \ & 
            $3.1\mathrm{e}10$ & $42.50$ & 
            $1.1\mathrm{e}10$ & $78.40$ & 
            $2.2\mathrm{e}9$ & $95.87$ & 
            $1.3\mathrm{e}9$ & $97.53$ & 
            $\mathbf{3.6\mathrm{e}8}$ & $\mathbf{99.31}$ \\ 
            
            & $500$ & 
            $7.9\mathrm{e}10$ \ \ \ \ & 
            $4.8\mathrm{e}10$ & $38.66$ & 
            $2.1\mathrm{e}10$ & $73.19$ & 
            $2.9\mathrm{e}9$ & $96.28$ & 
            $1.5\mathrm{e}9$ & $97.98$ & 
            $\mathbf{4.8\mathrm{e}8}$ & $\mathbf{99.39}$ \\
            \\
            
        \bottomrule
        \\
        \multicolumn{13}{l}{\makecell[l]{$DC$ is the average (10 trials) count of distance computations. $DC(S)$ represent percentage DC savings over baseline \km{}. Entries\\ in \textbf{bold} indicate the best average DC. Entries in \textcolor{blue}{blue} indicate similar performance.}}
    
    \end{tabular}

\end{sidewaystable}


%% file: table_realdata_kpp_runtime.tex
\begin{sidewaystable}


    \centering
    \caption{\textbf{$k$++ seeding:} Comparison of average runtime and speed-up of different algorithms}
    
    \label{tab:table_kpp_runtime}
    
    \begin{tabular}{lcccccccccccc}     
        
        \toprule
        Dataset & Clusters & \km{} & \multicolumn{2}{c}{\ham{}} & \multicolumn{2}{c}{\ann{}} & \multicolumn{2}{c}{\exp{}} & \multicolumn{2}{c}{\kb{}} &
        \multicolumn{2}{c}{\kg{}}\\
        \midrule
        
        & $k$ & $RT$ \ \ \ \ & $RT$ & $RT(S)$ & $RT$ & $RT(S)$ & $RT$ & $RT(S)$ & $RT$ & $RT(S)$ & $RT$ & $RT(S)$\\
        \cmidrule(lr){4-5} \cmidrule(lr){6-7} \cmidrule(lr){8-9} \cmidrule(lr){10-11} \cmidrule(lr){12-13}
        \\

        \multirow{3}{*}{BreastCancer} & $20$ & 
        $3.3$ \ \ \ \ & 
        $1.5$ & $54.57$ & 
        $1.1$ & $64.56$ & 
        \textcolor{blue}{$0.9$} & $71.36$ & 
        \textcolor{blue}{$1.0$} & $69.30$ & 
        \textcolor{blue}{$0.9$} & $\mathbf{72.41}$ \\ 
        
        & $30$ & $3.41$\ \ \ \  & 
        $1.8$ & $45.94$ & 
        $1.5$ & $54.90$ & 
        \textcolor{blue}{$1.28$} & $62.49$ & 
        \textcolor{blue}{$1.0$} & $\mathbf{69.44}$ & 
        \textcolor{blue}{$1.20$} & $64.64$ \\ 
        
        & $50$ & $5.72\mathrm{e}0$ \ \ \ \ & 
        $3.4$ & $40.60$ & 
        $2.5$ & $55.17$ & 
        $2.2$ & $61.41$ & 
        $\mathbf{1.4}$ & $\mathbf{73.98}$ & 
        $1.9$ & $65.32$ \\
        \\
        \midrule
        \\

        \multirow{3}{*}{CreditRisk} & $20$ & 
        $5.4$ \ \ \ \ & 
        $1.2$ & $77.28$ & 
        \textcolor{blue}{$1.0$} & $79.70$ & 
        $\mathbf{0.9}$ & $\mathbf{82.98}$ & 
        \textcolor{blue}{$1.1$} & $79.49$ & 
        \textcolor{blue}{$1.01$} & $81.22$ \\ 
        
        & $30$ & 
        $6.16$ \ \ \ \ & 
        $1.4$ & $77.26$ &
        \textcolor{blue}{$1.3$} & $78.65$ & 
        $\mathbf{1.1}$ & $\mathbf{80.70}$ & 
        \textcolor{blue}{$1.0$} & $83.54$ & 
        \textcolor{blue}{$1.2$} & $79.90$ \\ 
        
        & $50$ & 
        $7.8$ \ \ \ \ & 
        $2.0$ & $74.43$ & 
        $1.8$ & $76.89$ & 
        $1.9$ & $75.08$ & 
        $\mathbf{1.2}$ & $\mathbf{84.14}$ & 
        $1.8$ & $76.80$ \\
        \\
        \midrule
        \\

        \multirow{3}{*}{Sensor} & $50$ & 
        $1.1\mathrm{e}4$ \ \ \ \ & 
        $9.1\mathrm{e}3$ & $17.32$ & 
        $2.9\mathrm{e}3$ & $74.92$ & 
        $2.6\mathrm{e}3$ & $78.55$ & 
        $4.3\mathrm{e}3$ & $88.52$ & 
        $\mathbf{1.5\mathrm{e}3}$ & $\mathbf{97.00}$ \\ 
        
        & $80$ & 
        $1.9\mathrm{e}4$ \ \ \ \ & 
        $1.7\mathrm{e}4$ & $11.95$ & 
        $6.1\mathrm{e}3$ & $70.11$ & 
        $4.4\mathrm{e}3$ & $79.77$ & 
        $5.8\mathrm{e}3$ & $90.25$ & 
        $\mathbf{2.2\mathrm{e}3}$ & $\mathbf{97.80}$ \\

        & $100$ & 
        $2.5\mathrm{e}4$ \ \ \ \ & 
        $2.2\mathrm{e}4$ & $12.16$ & 
        $7.6\mathrm{e}3$ & $71.55$ & 
        $5.3\mathrm{e}3$ & $81.44$ & 
        $6.4\mathrm{e}3$ & $92.43$ & 
        $\mathbf{2.7\mathrm{e}3}$ & $\mathbf{98.09}$ \\
        \\
        \midrule
        \\

        \multirow{3}{*}{Kegg} & $50$ & 
        $3.0\mathrm{e}3$ \ \ \ \ & 
        $2.0\mathrm{e}3$ & $35.25$ & 
        $7.5\mathrm{e}2$ & $78.22$ & 
        $1.0\mathrm{e}3$ & $71.76$ & 
        $1.3\mathrm{e}3$ & $86.78$ & 
        $\mathbf{6.0\mathrm{e}2}$ & $\mathbf{95.96}$ \\ 
        
        & $80$ & 
        $4.8\mathrm{e}3$ \ \ \ \ & 
        $2.9\mathrm{e}3$ & $38.59$ & 
        $1.1\mathrm{e}3$ & $78.31$ & 
        $1.2\mathrm{e}3$ & $77.48$ & 
        $1.3\mathrm{e}3$ & $90.48$ & 
        $\mathbf{7.0\mathrm{e}2}$ & $\mathbf{96.64}$ \\

        & $100$ & 
        $6.6\mathrm{e}3$ \ \ \ \ & 
        $3.9\mathrm{e}3$ & $40.77$ & 
        $1.4\mathrm{e}3$ & $78.97$ & 
        $1.50\mathrm{e}3$ & $79.97$ & 
        $1.54\mathrm{e}3$ & $91.94$ & 
        $\mathbf{8.7\mathrm{e}2}$ & $\mathbf{97.07}$ \\ 
        \\
        \midrule
        \\

        \multirow{3}{*}{kddcup} & $100$ & 
        $7.7\mathrm{e}4$ \ \ \ \ & 
        $8.1\mathrm{e}3$ & $90.55$ & 
        $4.0\mathrm{e}3$ & $95.84$ & 
        $2.5\mathrm{e}3$ & \textcolor{blue}{$97.77$} & 
        $3.7\mathrm{e}3$ & \textcolor{blue}{$97.65$} & 
        $\mathbf{3.3\mathrm{e}3}$ & $\mathbf{97.15}$ \\

        & $300$ & 
        $3.6\mathrm{e}5$ \ \ \ \ & 
        $3.4\mathrm{e}4$ & $90.65$ & 
        $1.3\mathrm{e}4$ & $96.80$ & 
        $6.4\mathrm{e}3$ & \textcolor{blue}{$98.63$} & 
        $\mathbf{4.3\mathrm{e}3}$ & \textcolor{blue}{$98.66$} & 
        $7.7\mathrm{e}3$ & \textcolor{blue}{$98.48$} \\

        & $500$ & 
        $5.5\mathrm{e}5$ \ \ \ \ & 
        $7.7\mathrm{e}4$ & $86.29$ & 
        $3.2\mathrm{e}4$ & $94.57$ & 
        $9.9\mathrm{e}3$ & \textcolor{blue}{$98.54$} & 
        $\mathbf{5.6\mathrm{e}3}$ & \textcolor{blue}{$98.59$} & 
        $1.0\mathrm{e}4$ & \textcolor{blue}{$98.51$} \\ 
        \\
        \midrule
        \\

        \multirow{3}{*}{Twitter} & $100$ & 
        $6.9\mathrm{e}5$ \ \ \ \ & 
        $2.9\mathrm{e}5$ & $58.54$ & 
        $8.9\mathrm{e}4$ & $88.06$ & 
        $4.3\mathrm{e}4$ & $94.91$ & 
        $1.6\mathrm{e}5$ & $96.12$ & 
        $\mathbf{3.6\mathrm{e}4}$ & $\mathbf{98.54}$ \\ 
        
        & $300$ & 
        $3.1\mathrm{e}6$ \ \ \ \ & 
        $1.7\mathrm{e}6$ & $42.50$ & 
        $6.9\mathrm{e}5$ & $78.40$ & 
        $1.4\mathrm{e}5$ & $95.87$ & 
        $3.8\mathrm{e}5$ & $97.53$ & 
        $\mathbf{1.1\mathrm{e}5}$ & $\mathbf{99.31}$ \\ 
        
        & $500$ & 
        $4.4\mathrm{e}6$ \ \ \ \ & 
        $2.7\mathrm{e}6$ & $38.66$ & 
        $1.2\mathrm{e}6$ & $73.19$ & 
        $1.8\mathrm{e}5$ & $96.28$ & 
        $3.6\mathrm{e}5$ & $97.98$ & 
        $\mathbf{1.6\mathrm{e}5}$ & $\mathbf{99.39}$ \\ 
        \\
   
        \bottomrule
        \\
        \multicolumn{13}{l}{\makecell[l]{$RT$ is the average (10 trials) runtime in milliseconds. $RT(S)$ indicate percentage runtime speed-up over baseline \km{}. Entries\\ in \textbf{bold} indicate the best average runtime. Entries in \textcolor{blue}{blue} indicate similar performance.}}
    \end{tabular}

\end{sidewaystable}


%% file: Table10.tex
\begin{longtable}{llcc}
\label{tab:table_runtime_per_iter}
\makecell[l]{\textbf{Algorithm}} & \makecell[l]{\textbf{Data}} & {\textbf{Clusters}} & {\textbf{RT/Iter}} \\ 
  \hline
  Kmeans & Breastcancer &   20 & 0.292 \\ 
  Hamerly & Breastcancer &   20 & 0.126 \\ 
  DualTree & Breastcancer &   20 & 0.244 \\ 
  Annulus & Breastcancer &   20 & 0.074 \\ 
  Exponion & Breastcancer &   20 & 0.063 \\ 
  GeoKmeans & Breastcancer &   20 & 0.057 \\ 
  BallKmeans & Breastcancer &   20 & 0.080 \\ 
  Kmeans & Breastcancer &   30 & 0.407 \\ 
  Hamerly & Breastcancer &   30 & 0.212 \\ 
  DualTree & Breastcancer &   30 & 0.371 \\ 
  Annulus & Breastcancer &   30 & 0.132 \\ 
  Exponion & Breastcancer &   30 & 0.098 \\ 
  GeoKmeans & Breastcancer &   30 & 0.090 \\ 
  BallKmeans & Breastcancer &   30 & 0.100 \\ 
  Kmeans & Breastcancer &   50 & 0.588 \\ 
  Hamerly & Breastcancer &   50 & 0.354 \\ 
  DualTree & Breastcancer &   50 & 0.534 \\ 
  Annulus & Breastcancer &   50 & 0.227 \\ 
  Exponion & Breastcancer &   50 & 0.176 \\ 
  GeoKmeans & Breastcancer &   50 & 0.148 \\ 
  BallKmeans & Breastcancer &   50 & 0.120 \\ 
  Kmeans & CreditRisk &   20 & 0.306 \\ 
  Hamerly & CreditRisk &   20 & 0.052 \\ 
  DualTree & CreditRisk &   20 & 0.131 \\ 
  Annulus & CreditRisk &   20 & 0.041 \\ 
  Exponion & CreditRisk &   20 & 0.036 \\ 
  GeoKmeans & CreditRisk &   20 & 0.038 \\ 
  BallKmeans & CreditRisk &   20 & 0.053 \\ 
  Kmeans & CreditRisk &   30 & 0.455 \\ 
  Hamerly & CreditRisk &   30 & 0.079 \\ 
  DualTree & CreditRisk &   30 & 0.190 \\ 
  Annulus & CreditRisk &   30 & 0.065 \\ 
  Exponion & CreditRisk &   30 & 0.052 \\ 
  GeoKmeans & CreditRisk &   30 & 0.054 \\ 
  BallKmeans & CreditRisk &   30 & 0.059 \\ 
  Kmeans & CreditRisk &   50 & 0.679 \\ 
  Hamerly & CreditRisk &   50 & 0.165 \\ 
  DualTree & CreditRisk &   50 & 0.299 \\ 
  Annulus & CreditRisk &   50 & 0.132 \\ 
  Exponion & CreditRisk &   50 & 0.115 \\ 
  GeoKmeans & CreditRisk &   50 & 0.096 \\ 
  BallKmeans & CreditRisk &   50 & 0.082 \\ 
  Kmeans & kddcup &  100 & 1410.508 \\ 
  Hamerly & kddcup &  100 & 236.023 \\ 
  DualTree & kddcup &  100 & 214.744 \\ 
  Annulus & kddcup &  100 & 149.793 \\ 
  Exponion & kddcup &  100 & 43.639 \\ 
  GeoKmeans & kddcup &  100 & 44.791 \\ 
  BallKmeans & kddcup &  100 & 120.287 \\ 
  Kmeans & kddcup &  500 & 6884.794 \\ 
  Hamerly & kddcup &  500 & 1200.342 \\ 
  DualTree & kddcup &  500 & 1611.385 \\ 
  Annulus & kddcup &  500 & 403.393 \\ 
  Exponion & kddcup &  500 & 92.402 \\ 
  GeoKmeans & kddcup &  500 & 96.202 \\ 
  BallKmeans & kddcup &  500 & 124.432 \\ 
  Kmeans & Twitter &  500 & 16617.898 \\ 
  Hamerly & Twitter &  500 & 15497.373 \\ 
  DualTree & Twitter &  500 & 7448.485 \\ 
  Annulus & Twitter &  500 & 13173.470 \\ 
  Exponion & Twitter &  500 & 2272.200 \\ 
  GeoKmeans & Twitter &  500 & 2307.934 \\ 
  BallKmeans & Twitter &  500 & 3652.406 \\ 
   \hline
   \caption{\textbf{RT/Iter} is time (milliseconds) per iteration. Average of 10 trials is shown. On BreastCancer and CreditRisk, most methods have similar RT/Iter due to relatively small data size.} 
\end{longtable}

%% file: Table11.tex
\begin{sidewaystable}
    
    \caption{Comparison of runtime of \kg{} with available implementation at \href{https://github.com/ghamerly/fast-kmeans/blob/master/src/hamerly_kmeans.cpp}{Hamerly}}
    
    \centering
    \begin{tabular}{ccccccc}
        \toprule 
        Dataset  & Clusters & \ham{} & \ann{} & \kg{} & \makecell[c]{$RT(S)$ of \kg{}\\ over \ham{}} & 
        \makecell[c]{$RT(S)$ of \kg{}\\ over \ann{}}\\
        
        \midrule\\
    
        & $k$ & $RT$  & $RT$ & $RT$ & $RT(S)$ & $RT(S)$ \\
        \cmidrule{2-5}\cmidrule{6-7}\\

        BreastCancer & 50 & 11.5 & 4.8 & \textbf{2.96} & $74.26\%$ & $38.33\%$ \\ 
        \\
        \midrule
        \\

        CreditRisk & 50 & 9.34 & 5 & \textbf{2.65} & $71.62\%$ & $47\%$ \\ 
        \\
        \midrule   
        \\

        Sensor & 80 & $5.03\mathrm{e}4$ & $1.19\mathrm{e}4$ & $\mathbf{5.79\mathrm{e}3}$ & $88.20\%$ & $51.54\%$ \\ 
        \\
        \midrule
        \\

        Kegg & 80 & $7.74\mathrm{e}3$ & $4.29\mathrm{e}4$ & $\mathbf{1.64\mathrm{e}3}$ & $78.81\%$ &  $61.13\%$\\ 
        \\
        \midrule
        \\

        Kddcup & 100 & $2.93\mathrm{e}5$ & $3.59\mathrm{e}4$ & $\mathbf{2.15\mathrm{e}4}$ & $92.64\%$ & $39.93\%$\\
        
        \\
        \midrule
        \\

        Twitter & 100 & $3.09\mathrm{e}6$ & $3.13\mathrm{e}5$ & $\mathbf{1.74\mathrm{e}5}$ & $94.36\%$ & $44.28\%$\\ 
        \\   
        \bottomrule\\
        
        \multicolumn{7}{l}{\makecell[l]{$RT$ is the average (5 trials) runtime in milliseconds. $RT(S)$ indicate percentage runtime speed-up over\\ other algorithms. Entries in \textbf{bold} indicate better performance. Entries in \textcolor{blue}{Blue} indicate similar runtime\\ speedup}} 
    
    \end{tabular}
    \label{tab:comparison_with_ham}
    
\end{sidewaystable}

%% file: Savings_Table1.tex
\begin{table}[!htb]
\footnotesize
\centering
    
   \caption{Distance computations savings due to LE and LHE conditions in Algorithm 2}
    \begin{tabular}{ccccc}
         \toprule
         \multicolumn{5}{c}{(A) BreastCancer, $m = 569$, $k = 50$, $m \times k = 28,450$}\\
         \midrule
         Iteration & \makecell[c]{Neighborhood\\Size\\($\text{neighborhood size}/k^2$)} & \makecell[c]{Number of \textsf{LE}\\(\textsf{LE}/$m$)} & \makecell[c]{PDC1\\(\textsf{LHE}, ((PDC1-\textsf{LHE})/PDC1)*100)} &  \makecell[c]{PDC2\\(\textsf{HE}, ((PDC2-\textsf{HE})/PDC2)*100)} \\
         \midrule
         5 & 435 (17.4) & 295 (0.518453)  & 13700 (596, 95.6496)  & 29800 (12, 99.9597)  \\
        10 & 428 (17.12) & 299 (0.525483)  & 13500 (574, 95.7481)  & 28700 (2, 99.993)  \\
        \midrule
        \midrule\\
         
        \multicolumn{5}{c}{(B) CreditRisk, $m = 1000$, $k = 50$, $m \times k = 50,000$}\\
        \midrule

        5 & 205 (8.2) & 812 (0.812)  & 9400 (257, 97.266)  & 12850 (44, 99.6576) \\
        10 & 184 (7.36) & 842 (0.842)  & 7900 (217, 97.2532)  & 10850 (22, 99.7972) \\
        15 & 176 (7.04) & 851 (0.851)  & 7450 (203, 97.2752)  & 10150 (12, 99.8818) \\
        20 & 174 (6.96) & 855 (0.855)  & 7250 (210, 97.1034)  & 10500 (6, 99.9429) \\

        \midrule
        \midrule\\

        \multicolumn{5}{c}{(C) Sensor, $m = 58,509$, $k = 80$, $m \times k = 4,680,720$}\\
        \midrule
        
        10 & 4926 (76.9688) & 5117 (0.0874566)  & 4.27e+06 (539269, 87.3748)  & 4.31e+07 (3614, 99.9916) \\
        20 & 4917 (76.8281) & 5817 (0.0994206)  & 4.21e+06 (478889, 88.6394)  & 3.83e+07 (2463, 99.9936) \\
        30 & 4886 (76.3438) & 6029 (0.103044)  & 4.19e+06 (439903, 89.5221)  & 3.51e+07 (1635, 99.9954) \\
        40 & 4839 (75.6094) & 6448 (0.110205)  & 4.16e+06 (413687, 90.0673)  & 3.30e+07 (1125, 99.9966) \\
        50 & 4906 (76.6562) & 6485 (0.110838)  & 4.16e+06 (393063, 90.5557)  & 3.14e+07 (839, 99.9973) \\
        60 & 4961 (77.5156) & 6601 (0.11282)  & 4.15e+06 (379645, 90.8577)  & 3.03e+07 (639, 99.9979) \\
        70 & 4984 (77.875) & 6629 (0.113299)  & 4.15e+06 (367975, 91.134)  & 2.94e+07 (588, 99.998) \\
        80 & 4970 (77.6562) & 6627 (0.113265)  & 4.15e+06 (359410, 91.3407)  & 2.87e+07 (349, 99.9988) \\
        90 & 4998 (78.0938) & 6678 (0.114136)  & 4.14e+06 (354969, 91.4393)  & 2.83e+07 (283, 99.999) \\
        100 & 4993 (78.0156) & 6737 (0.115145)  & 4.14e+06 (349973, 91.5501)  & 2.79e+07 (263, 99.9991) \\
        110 & 4999 (78.1094) & 6741 (0.115213)  & 4.14e+06 (346972, 91.6219)  & 2.77e+07 (202, 99.9993) \\
        120 & 4995 (78.0469) & 6670 (0.114)  & 4.14e+06 (346072, 91.6551)  & 2.76e+07 (114, 99.9996) \\
        130 & 5014 (78.3438) & 6729 (0.115008)  & 4.14e+06 (345566, 91.6578)  & 2.76e+07 (128, 99.9995) \\
        140 & 5002 (78.1562) & 6735 (0.11511)  & 4.14e+06 (345591, 91.6563)  & 2.76e+07 (137, 99.9995) \\
        150 & 4949 (77.3281) & 6816 (0.116495)  & 4.13e+06 (345446, 91.6467)  & 2.76e+07 (117, 99.9996) \\
        160 & 4951 (77.3594) & 6878 (0.117555)  & 4.13e+06 (345136, 91.6442)  & 2.76e+07 (82, 99.9997) \\
        170 & 4940 (77.1875) & 6926 (0.118375)  & 4.12e+06 (344721, 91.6464)  & 2.75e+07 (38, 99.9999) \\
        180 & 4941 (77.2031) & 6973 (0.119178)  & 4.12e+06 (344474, 91.6448)  & 2.75e+07 (22, 99.9999) \\
        190 & 4938 (77.1562) & 6975 (0.119212)  & 4.12e+06 (344474, 91.6445)  & 2.75e+07 (14, 99.9999) \\
        200 & 4938 (77.1562) & 6997 (0.119588)  & 4.12e+06 (344377, 91.6433)  & 2.75e+07 (27, 99.9999) \\
        210 & 4938 (77.1562) & 7001 (0.119657)  & 4.12e+06 (344314, 91.6442)  & 2.75e+07 (9, 100) \\
        220 & 4937 (77.1406) & 6999 (0.119623)  & 4.12e+06 (344384, 91.6428)  & 2.75e+07 (0, 100) \\

        \midrule
        \midrule\\

        \multicolumn{5}{c}{(D) Kegg, $m = 65,554$, $k = 80$, $m \times k = 5,244,320$}\\
         \midrule

        10 & 4650 (72.6562) & 26159 (0.399045)  & 3.1516e+06 (486727, 84.5562)  & 3.89382e+07 (2265, 99.9942) \\
        20 & 4633 (72.3906) & 27343 (0.417107)  & 3.05688e+06 (393977, 87.1118)  & 3.15182e+07 (872, 99.9972) \\
        30 & 4613 (72.0781) & 27737 (0.423117)  & 3.02536e+06 (385587, 87.2548)  & 3.0847e+07 (192, 99.9994) \\
        40 & 4633 (72.3906) & 28179 (0.429859)  & 2.99e+06 (383156, 87.1854)  & 3.06525e+07 (135, 99.9996) \\
        50 & 4632 (72.375) & 28214 (0.430393)  & 2.9872e+06 (380755, 87.2538)  & 3.04604e+07 (91, 99.9997) \\
        60 & 4633 (72.3906) & 28303 (0.431751)  & 2.98008e+06 (380063, 87.2466)  & 3.0405e+07 (91, 99.9997) \\
        70 & 4646 (72.5938) & 28289 (0.431537)  & 2.9812e+06 (379630, 87.2659)  & 3.03704e+07 (26, 99.9999) \\
        80 & 4643 (72.5469) & 28274 (0.431309)  & 2.9824e+06 (379131, 87.2877)  & 3.03305e+07 (0, 100) \\

        \bottomrule
        \bottomrule
        \multicolumn{5}{c}{\makecell[c]{LE and LHE indicate the number of LE and LHE data points in a given iteration\\
        PDC1 (possible distance computations 1): total DC possible left after accounting for LE points. PDC1 = $(m -|\textsf{LE}|) \times k$\\
        PDC2 (possible distance computations 2): total DC possible with LHE points. PDC2 = $(|\textsf{LHE}|) \times k$}}
    \end{tabular}
    \label{tab:savings1}
\end{table}

%% file: Savings_Table2.tex
\begin{table}[!htb]
    \centering
    \caption{Distance computations savings due to LE and LHE conditions in Algorithm 2}
    \begin{tabular}{ccccc}
        \toprule
         \multicolumn{5}{c}{(A) KddCup, $m = 494,020$, $k = 100$, $m \times k = 49,402,000$}\\
         \midrule
         Iteration & \makecell[c]{Number\\Neighbor} & \makecell[c]{Number of \textsf{LE}\\(\textsf{LE}/$m$)} & \makecell[c]{PDC1\\(\textsf{LHE}, ((PDC1-\textsf{LHE})/PDC1)*100)} &  \makecell[c]{PDC2\\(\textsf{HE}, ((PDC2-\textsf{HE})/PDC2)*100)} \\
         \midrule
            20 & 6082 (60.82) & 211805 (0.428738)  & 2.82e+07 (844174, 97.0088)  & 8.44e+07 (34054, 99.9597) \\
            40 & 7209 (72.09) & 213285 (0.431734)  & 2.80e+07 (532775, 98.1022)  & 5.32e+07 (18084, 99.9661) \\
            60 & 5140 (51.4) & 234119 (0.473906)  & 2.59e+07 (425843, 98.3615)  & 4.25e+07 (5341, 99.9875) \\
            80 & 4609 (46.09) & 429455 (0.869307)  & 6.45e+06 (213042, 96.7003)  & 2.13e+07 (4763, 99.9776) \\
            100 & 5025 (50.25) & 427005 (0.864348)  & 6.70e+06 (183579, 97.2606)  & 1.83e+07 (4061, 99.9779) \\
            120 & 4375 (43.75) & 426498 (0.863321)  & 6.75e+06 (180435, 97.3278)  & 1.80e+07 (5238, 99.971) \\
            140 & 4383 (43.83) & 426593 (0.863514)  & 6.74e+06 (171405, 97.4579)  & 1.71e+07 (3182, 99.9814) \\
            160 & 4320 (43.2) & 425792 (0.861892)  & 6.82e+06 (168925, 97.5241)  & 1.68925e+07 (2826, 99.9833) \\
            180 & 4096 (40.96) & 430512 (0.871447)  & 6.35e+06 (161098, 97.4633)  & 1.61e+07 (1710, 99.9894) \\
            200 & 4082 (40.82) & 431417 (0.873278)  & 6.26e+06 (158477, 97.4685)  & 1.58e+07 (1988, 99.9875) \\
            220 & 4071 (40.71) & 431053 (0.872542)  & 6.29e+06 (161158, 97.4406)  & 1.61e+07 (1555, 99.9904) \\
            240 & 3978 (39.78) & 430613 (0.871651)  & 6.34e+06 (162069, 97.444)  & 1.62e+07 (1701, 99.9895) \\
            260 & 4007 (40.07) & 428224 (0.866815)  & 6.57e+06 (159269, 97.5794)  & 1.59e+07 (1796, 99.9887) \\
            280 & 3882 (38.82) & 431084 (0.872604)  & 6.2936e+06 (155799, 97.5245)  & 1.55e+07 (1225, 99.9921) \\
            300 & 3857 (38.57) & 430658 (0.871742)  & 6.3362e+06 (158968, 97.4911)  & 1.58e+07 (1172, 99.9926) \\
            320 & 3893 (38.93) & 431695 (0.873841)  & 6.2325e+06 (154046, 97.5283)  & 1.54e+07 (1041, 99.9932) \\
            340 & 3862 (38.62) & 431745 (0.873942)  & 6.22e+06 (154731, 97.5154)  & 1.54e+07 (745, 99.9952) \\
            360 & 3860 (38.6) & 431282 (0.873005)  & 6.27e+06 (152898, 97.5629)  & 1.52e+07 (596, 99.9961) \\
            380 & 3901 (39.01) & 431093 (0.872623)  & 6.29e+06 (153523, 97.5603)  & 1.53e+07 (977, 99.9936) \\
            400 & 3738 (37.38) & 430665 (0.871756)  & 6.33e+06 (154443, 97.5623)  & 1.54e+07 (899, 99.9942) \\
            420 & 3714 (37.14) & 431568 (0.873584)  & 6.24e+06 (152011, 97.566)  & 1.52e+07 (563, 99.9963) \\
            440 & 3766 (37.66) & 431239 (0.872918)  & 6.27e+06 (152955, 97.5637)  & 1.52e+07 (161, 99.9989) \\
            460 & 3747 (37.47) & 431176 (0.872791)  & 6.28e+06 (152938, 97.5664)  & 1.52e+07 (69, 99.9995) \\
            480 & 3774 (37.74) & 430922 (0.872276)  & 6.30e+06 (153003, 97.5752)  & 1.53e+07 (197, 99.9987) \\
            500 & 3706 (37.06) & 431284 (0.873009)  & 6.27e+06 (152114, 97.5753)  & 1.52e+07 (975, 99.9936) \\

            \midrule
            \midrule\\

            \multicolumn{5}{c}{(B) Twitter, $m = 583, 250$, $k = 100$, $m \times k = 58,325,000$}\\
            \midrule

            20 & 7269 (72.69) & 17415 (0.0298586)  & 5.65835e+07 (7.84955e+06, 86.1275)  & 7.84955e+08 (47932, 99.9939) \\
            40 & 7215 (72.15) & 23496 (0.0402846)  & 5.59754e+07 (6.48537e+06, 88.4139)  & 6.48537e+08 (36237, 99.9944) \\
            60 & 7145 (71.45) & 57722 (0.0989661)  & 5.25528e+07 (5.48781e+06, 89.5575)  & 5.48781e+08 (29175, 99.9947) \\
            80 & 7180 (71.8) & 90246 (0.15473)  & 4.93004e+07 (4.70312e+06, 90.4603)  & 4.70312e+08 (23897, 99.9949) \\
            100 & 7172 (71.72) & 121898 (0.208998)  & 4.61352e+07 (4.1235e+06, 91.0621)  & 4.1235e+08 (16040, 99.9961) \\
            120 & 7242 (72.42) & 150036 (0.257241)  & 4.33214e+07 (3.664e+06, 91.5423)  & 3.664e+08 (13187, 99.9964) \\
            140 & 7261 (72.61) & 172799 (0.296269)  & 4.10451e+07 (3.31647e+06, 91.9199)  & 3.31647e+08 (9195, 99.9972) \\
            160 & 7269 (72.69) & 188570 (0.323309)  & 3.9468e+07 (3.10025e+06, 92.1449)  & 3.10025e+08 (7037, 99.9977) \\
            180 & 7261 (72.61) & 202846 (0.347786)  & 3.80404e+07 (2.91154e+06, 92.3462)  & 2.91154e+08 (6019, 99.9979) \\
            200 & 7245 (72.45) & 214272 (0.367376)  & 3.68978e+07 (2.71546e+06, 92.6406)  & 2.71546e+08 (5694, 99.9979) \\
            220 & 7259 (72.59) & 231414 (0.396766)  & 3.51836e+07 (2.57578e+06, 92.679)  & 2.57578e+08 (4900, 99.9981) \\
            240 & 7261 (72.61) & 241341 (0.413787)  & 3.41909e+07 (2.47189e+06, 92.7703)  & 2.47189e+08 (2924, 99.9988) \\
            260 & 7235 (72.35) & 247266 (0.423945)  & 3.35984e+07 (2.42259e+06, 92.7896)  & 2.42259e+08 (1900, 99.9992) \\
            280 & 7222 (72.22) & 251525 (0.431247)  & 3.31725e+07 (2.37744e+06, 92.8331)  & 2.37744e+08 (1794, 99.9992) \\
            300 & 7200 (72) & 256209 (0.439278)  & 3.27041e+07 (2.32782e+06, 92.8822)  & 2.32782e+08 (1718, 99.9993) \\
            320 & 7125 (71.25) & 260117 (0.445979)  & 3.23133e+07 (2.28356e+06, 92.9331)  & 2.28356e+08 (1730, 99.9992) \\
            340 & 7130 (71.3) & 264356 (0.453246)  & 3.18894e+07 (2.2389e+06, 92.9792)  & 2.2389e+08 (2108, 99.9991) \\
            360 & 7123 (71.23) & 269373 (0.461848)  & 3.13877e+07 (2.20056e+06, 92.9891)  & 2.20056e+08 (1897, 99.9991) \\
            380 & 7109 (71.09) & 274107 (0.469965)  & 3.09143e+07 (2.16172e+06, 93.0074)  & 2.16172e+08 (1558, 99.9993) \\
            400 & 7136 (71.36) & 277528 (0.47583)  & 3.05722e+07 (2.13878e+06, 93.0042)  & 2.13878e+08 (1020, 99.9995) \\
            420 & 7152 (71.52) & 280449 (0.480838)  & 3.02801e+07 (2.12078e+06, 92.9961)  & 2.12078e+08 (867, 99.9996) \\
            440 & 7159 (71.59) & 282352 (0.484101)  & 3.00898e+07 (2.10947e+06, 92.9894)  & 2.10947e+08 (346, 99.9998) \\
            460 & 7158 (71.58) & 283070 (0.485332)  & 3.0018e+07 (2.10442e+06, 92.9895)  & 2.10442e+08 (260, 99.9999) \\
            480 & 7160 (71.6) & 283811 (0.486603)  & 2.99439e+07 (2.10028e+06, 92.9859)  & 2.10028e+08 (241, 99.9999) \\
            500 & 7164 (71.64) & 284564 (0.487894)  & 2.98686e+07 (2.09503e+06, 92.9858)  & 2.09503e+08 (326, 99.9998) \\

            \bottomrule
            \bottomrule
            \multicolumn{5}{c}{\makecell[c]{LE and LHE indicate the number of LE and LHE data points in a given iteration\\
            PDC1 (possible distance computations 1): total DC possible left after accounting for LE points. PDC1 = $(m -|\textsf{LE}|) \times k$\\
            PDC2 (possible distance computations 2): total DC possible with LHE points. PDC2 = $(|\textsf{LHE}|) \times k$}}
            
    \end{tabular}
    \label{tab:savings2}
\end{table}

%% file: Table12.tex
\begin{table}
    \centering

    \caption{Comparison of memory allocation of different algorithms}
    
    \label{tab:table_space_usage}
    
    \begin{tabular}{cccccc}     
        
        \toprule
        Dataset & Clusters \ \ \ \ &  \ham{} \ \ \ \ & \ann{} \ \ \ \ & \exp{} \ \ \ \ & \kg{} 
        \\
        \midrule
        &  &  &  &  &  \\

        BreastCancer & 20 \ \ \ \ & 179 KB \ \ \ \ & 243 KB \ \ \ \ & 185 KB \ \ \ \ 
        & 173 KB  
        \\ 
        \midrule
        \\
        
        CreditRisk & 20 \ \ \ \ & 194 KB \ \ \ \ & 305 KB \ \ \ \ & 200 KB \ \ \ \  & 275 KB 
        \\  
        \midrule
        \\
            
        Sensor & 50 \ \ \ \ & 13 MB \ \ \ \ & 27 MB \ \ \ \ & 13 MB \ \ \ \ & 14 MB \\  
        \midrule
        \\

        Kegg & 50 \ \ \ \ & 11 MB \ \ \ \ & 26 MB \ \ \ \ & 11 MB \ \ \ \ & 11 MB \\  
        \midrule
        \\

        Kddcup & 20 \ \ \ \ & 93 MB \ \ \ \ & 148 MB \ \ \ \  & 93 MB \ \ \ \ & 89 MB 
        \\  
        \midrule
        \\
             
        Twitter & 20 \ \ \ \ & 214 MB \ \ \ \ & 278 MB \ \ \ \ & 214 MB \ \ \ \ & 209 MB \\   
        \bottomrule
        \\
        \multicolumn{6}{l}{\makecell[c]{Memory allocated during the program's execution (obtained from \href{https://valgrind.org/docs/manual/ms-manual.html}{Valgrind's} memory profiler).\\ To ensure fairness, all algorithms are started with the same seed on a given dataset.}}
    \end{tabular}
\end{table}